\RequirePackage{fix-cm}
\pdfoutput=1
\PassOptionsToPackage{table}{xcolor}
\PassOptionsToPackage{capitalize,noabbrev}{cleveref}
\documentclass{xiaolab}
\usepackage[T1]{fontenc}
\usepackage{lmodern}
\usepackage{amsmath,amssymb,amsthm,booktabs,graphicx,tabularx,multirow,longtable,hyperref,url}
\usepackage{xcolor}
\usepackage{algorithm,algpseudocode}
\usepackage[most]{tcolorbox}
\usepackage{tikz}
\usetikzlibrary{positioning,arrows.meta}
\usepackage{subcaption}
\usepackage{needspace}
\usepackage[capitalize,noabbrev]{cleveref}
\newcommand{\panelref}[2]{\hyperref[#1]{\ref*{#1}(#2)}}
\newcommand{\subpanelref}[2]{\hyperref[#2]{\ref*{#1}(\subref*{#2})}}
\newtheorem{proposition}{Proposition}

\newtheorem{lemma}{Lemma}

\DeclareMathOperator{\KL}{KL}
\definecolor{bestcell}{RGB}{200,230,201}
\definecolor{mydarkblue}{rgb}{0,0.08,0.45}
\colorlet{septwash}{bestcell!30}
\title{A Model Can Help Itself: Reward-Free Self-Training for LLM Reasoning}
\authors{
Mengqi Li$^{1}$ \quad Lei Zhao$^{2}$ \quad Anthony Man-Cho So$^{3}$ \quad Ruoyu Sun$^{1,\dagger}$ \quad Xiao Li$^{1,\dagger}$\\ 
\texttt{mengqili1@link.cuhk.edu.cn, l.zhao@sjtu.edu.cn,} \\ 
\texttt{manchoso@se.cuhk.edu.hk, \{sunruoyu,lixiao\}@cuhk.edu.cn}
}

\affiliations{
  $^{1}$The Chinese University of Hong Kong, Shenzhen \quad
  $^{2}$Shanghai Jiao Tong University\\
  $^{3}$The Chinese University of Hong Kong\\
\quad $^{\dagger}$Corresponding authors
}
\hypersetup{colorlinks=true,pdfborder={0 0 0},linkcolor=mydarkblue,citecolor=mydarkblue,filecolor=mydarkblue,urlcolor=mydarkblue,pdfauthor={Mengqi Li, Lei Zhao, Anthony Man-Cho So, Ruoyu Sun, Xiao Li},pdftitle={A Model Can Help Itself},pdfsubject={},pdfkeywords={}}
\abstract{
Can a language model improve reasoning by learning from its own imperfect responses, without rewards or teacher-provided solutions?
We present Self-evolving Post-Training (SePT), a simple method that alternates temperature-controlled self-generation with next-token likelihood training.
Each round uses the updated model to generate new training responses, with one response per prompt by default and no correctness filtering.
Across six mathematical benchmarks, SePT improves a temperature-selected no-training baseline by 11.4 and 6.7 AVG points on Qwen2.5-Math-7B and Qwen2.5-7B, respectively, where AVG averages Pass@1, Pass@8 and Pass@32 across benchmarks.
We analyze how sampling temperature shapes the learning signal and investigate the value of the resulting responses.
Responses from a SePT-trained model improve a student initialized from the original weights, while their reasoning prefixes help an unchanged model complete solutions, even when matched in length to prefixes from a colder initial model.
Comparing next-token predictions at identical contexts also reveals changes in token rankings that decoding-temperature adjustment cannot reproduce.
Further evaluations across nine starting models, general reasoning and code generation examine broader applicability.
Together, these results show that reward-free self-training can improve both a model's predictions and the supervision it provides.
}
\projectpage{https://github.com/ElementQi/SePT}
\codelink{https://github.com/ElementQi/SePT}
\begin{document}
\maketitle
\begin{center}
\captionsetup{type=figure}
\centering
\begin{subfigure}[t]{0.35\linewidth}
\centering\vspace{0pt}
\includegraphics[width=\linewidth]{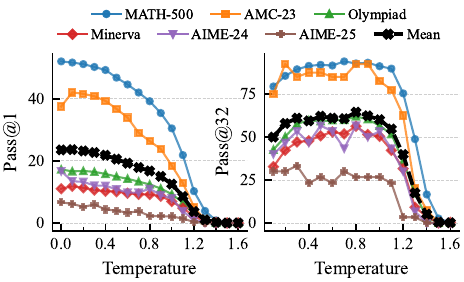}
\caption{Decoding-temperature sensitivity.}\label{fig:motivation-temperature}
\end{subfigure}\hfill
\begin{subfigure}[t]{0.35\linewidth}
\centering\vspace{0pt}
\includegraphics[width=\linewidth]{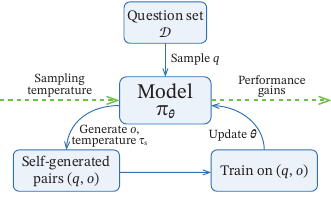}
\caption{SePT self-training loop.}\label{fig:motivation-loop}
\end{subfigure}\hfill
\begin{subfigure}[t]{0.28\linewidth}
\centering\vspace{0pt}
\includegraphics[width=\linewidth]{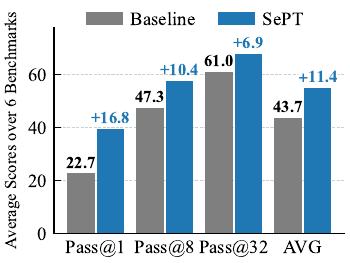}
\caption{Improvement after training.}\label{fig:motivation-gain}
\end{subfigure}
\caption{Motivation, workflow and performance of SePT on Qwen2.5-Math-7B.
\textbf{(a)} Lower decoding temperatures favor Pass@1, while moderate temperatures favor Pass@32.
\textbf{(b)} The model generates responses, learns from them, and generates the next batch with its updated parameters.
\textbf{(c)} SePT improves over the temperature-tuned no-training Baseline on the six-benchmark mean Pass@1, Pass@8, Pass@32 and AVG.}\label{fig:original-motivation}
\end{center}

\section{Introduction}\label{sec:intro}
Large language models (LLMs) have made substantial progress on complex reasoning tasks through chain-of-thought prompting and reasoning-oriented post-training \citep{wei2022chain,guo2025deepseek}.
Developing simple and effective training strategies for reasoning models has consequently attracted growing interest.

Reinforcement learning with verifiable rewards (RLVR) favors responses using rule-based feedback, as in GRPO \citep{shao2024deepseekmath} and subsequent reasoning-training systems \citep{yu2025dapo,zheng2025gspo}.
Supervised fine-tuning (SFT) instead learns from supplied reasoning demonstrations: s1 and LIMO study carefully chosen traces \citep{muennighoff2025s1,ye2025limo}, while OpenThoughts studies their composition and curation \citep{guha2025openthoughts}.

Self-training offers another route by reusing a model's own generations \citep{zelikman2022star,zuo2025ttrl,agarwal2025unreasonable}.
The model already produces candidate reasoning responses, but it is unclear whether these raw responses are useful training targets without a separate mechanism for judging their quality.

\paragraph{Motivation.}
Figure~\subpanelref{fig:original-motivation}{fig:motivation-temperature} shows that the same model achieves different reasoning accuracy at different decoding temperatures. This motivates a simple hypothesis: the starting distribution already contains useful response preferences that likelihood training can reinforce.
We therefore ask: \emph{Can LLMs improve their reasoning performance without rewards or teacher solutions, using only their own sampled responses for training?}

\paragraph{Main contributions.}
In this work, we provide an affirmative answer to this question. Specifically:
\begin{enumerate}
\item We propose Self-evolving Post-Training (SePT), a simple reward-free self-training method that alternates self-generation with likelihood training on every generated response. \emph{Temperature Dynamics} controls sampling; \emph{Online Data Refresh} uses the updated model to generate each batch, without correctness filtering or reward weighting.
\item We analyze how sampling temperature shapes the learning signal and show that SePT-generated responses provide useful supervision for a student initialized from the original model. Their reasoning prefixes also help an unchanged model complete solutions. By comparing next-token predictions at identical contexts, we further show that SePT changes token rankings in ways that scalar decoding-temperature adjustments cannot reproduce.
\item We evaluate SePT across nine starting models on six mathematical benchmarks, demonstrating gains over a strong no-training baseline selected by a decoding-temperature sweep. Additional experiments on general reasoning and code generation examine its applicability beyond mathematics.
\end{enumerate}

\section{The SePT Method}\label{sec:method}
\subsection{Self-generation and likelihood training}
Let $\pi_\theta(o\mid q;\tau)$ denote an autoregressive model generating response $o$ for prompt $q$ at temperature $\tau$.
In each round, SePT snapshots the current model, samples prompts from $\mathcal D$, and generates responses at sampling temperature $\tau_s$.
It then performs $\mu$ optimization steps on minibatches of these generated pairs at training temperature $\tau_t$, without regenerating responses between those steps.
After these updates, the learner becomes the source for the next round (\Cref{alg:sept}).
SePT fits the generated responses by negative log likelihood:
\begin{equation}\label{eq:reference}
L_b(\theta)
=-\mathbb E_{q\sim\mathcal D,\ o\sim\pi_{\rm old}(\cdot\mid q;\tau_s)}
\log\pi_\theta(o\mid q;\tau_t).
\end{equation}
For the analysis, we use the sequence-sum likelihood objective; the implementation averages negative log likelihood over generated response tokens (Appendix~\ref{app:implementation-objective}).

\begin{algorithm}[t]
\caption{Self-evolving Post-Training (SePT)}\label{alg:sept}
\begin{algorithmic}[1]
\Require Initial model $\pi_{\theta_0}$, prompts $\mathcal D$, sampling schedule $\tau_s(r)$, training temperature $\tau_t$, rollouts $G$, rounds $M$, optimization steps per round $\mu$
\State Initialize $\theta\leftarrow\theta_0$
\For{$r=1,\ldots,M$}
  \State Freeze the round source: $\pi_{\rm old}\leftarrow\pi_\theta$
  \State Sample a prompt batch $\mathcal D_b$ from $\mathcal D$; initialize $\mathcal D_{\rm SePT}\leftarrow\varnothing$
  \For{each $q\in\mathcal D_b$}
    \State Sample $G$ responses $o_i\sim\pi_{\rm old}(\cdot\mid q;\tau_s(r))$
    \State Add all pairs $\{(q,o_i)\}_{i=1}^G$ to $\mathcal D_{\rm SePT}$
  \EndFor
  \For{$j=1,\ldots,\mu$}
    \State Take the next minibatch from $\mathcal D_{\rm SePT}$; update $\theta$ by minimizing
    \Statex \hspace{\algorithmicindent}\hspace{\algorithmicindent}response negative log likelihood at $\tau_t$ (Eq.~\ref{eq:reference})
  \EndFor
\EndFor
\State \Return $\pi_\theta$
\end{algorithmic}
\end{algorithm}

\subsection{Temperature Dynamics}\label{sec:temperature-dynamics}
\paragraph{Learning signal and ideal fitting.}
Sampling temperature $\tau_s$ determines which responses supply supervision, while training temperature $\tau_t$ determines their likelihood under the student.
At prefix $\zeta=(q,o_{<t})$, let $b=\pi_{\rm old}(\cdot\mid\zeta;\tau_s)$ and $p_\theta=\pi_\theta(\cdot\mid\zeta;\tau_t)$.
The expected logit update direction is $(b-p_\theta)/\tau_t$. At the start of a round, $\tau_s<\tau_t$ makes likely tokens appear more often than their probabilities at the training temperature, supplying a learning signal without correctness labels.

\Needspace{6\baselineskip}
For positive temperatures under the finite-horizon, full-support sequence-sum reference, independently and exactly fitting each visited conditional to a frozen source gives $p_\theta^*=b$, or
\[
z^*(\zeta)=\frac{\tau_t}{\tau_s}z_{\rm old}(\zeta)+c(\zeta)\mathbf1,
\]
where $z$ denotes unscaled logits and $c(\zeta)$ is an all-token constant (Appendix~\ref{app:reference-proofs}). Thus cooling reinforces existing preferences, and repeated ideal fitting remains equivalent to adjusting the initial model's decoding temperature.

\paragraph{Why scaling alone is insufficient.}
A stronger preference helps only when it favors successful continuations. Cooling can reinforce confident errors as well as correct choices. Appendix~\ref{app:mechanism-proofs} relates the change in success probability to probability--correctness alignment across visited prefixes. The temperature-ratio target therefore explains the learning signal, but not the benefit of shared-model training.

\paragraph{Shared parameters enable useful transfer.}
A transformer jointly fits many prefixes with one parameter vector, so learning from one response can change predictions elsewhere. Let $g=-\nabla_\theta L_b(\theta)$ be the aggregate training direction and $J(\theta)$ the expected task accuracy. For a small gradient step,
\[
J(\theta+\eta g)-J(\theta)
=\eta\,\nabla J(\theta)^\top g+O(\eta^2).
\]
For sufficiently small steps, positive alignment $\nabla J^\top g>0$ gives a net gain even with incorrect targets. Appendix~\ref{app:mechanism-proofs} expands this alignment into gradient interactions between training and evaluation responses. $J$ is used only for analysis. Appendix~\ref{app:constructive-projection} illustrates this with a two-parameter model: fitting a colder source at one context changes the most likely action at another, improving success beyond every decoding temperature of the initial model. Pure temperature scaling also fits the training targets, but the optimization trajectory selects a different solution. Section~\ref{sec:components} tests these effects through fixed-prefix comparisons, generator controls and prefix interventions.

\subsection{Online Data Refresh}\label{sec:refresh-mechanism}
\paragraph{Updating the generator.}
For a given question, the response distribution depends on both the decoding settings and the model weights. Temperature Dynamics controls the sampling temperature, while Online Data Refresh updates the weights used for generation: after each round of likelihood training, the updated model generates the next batch of responses.

\paragraph{Changing contexts and token targets.}
Changing the generator's weights can change both which reasoning prefixes it visits and which tokens it generates at those prefixes. To see how both affect training, let $\phi$ denote the generator's weights, $d_\phi(\zeta)$ the expected number of visits to prefix $\zeta$, and $b_\phi(a\mid\zeta)$ its next-token distribution at the sampling temperature. For a student with weights $\theta$, the expected negative-loss gradient under the sequence-sum objective is
\[
g(\theta,\phi)=-\nabla_\theta L_{b_\phi}(\theta)
=\sum_\zeta d_\phi(\zeta)\,\mathbb E_{a\sim b_\phi(\cdot\mid\zeta)}
\bigl[\nabla_\theta\log p_\theta(a\mid\zeta)\bigr].
\]
Here, $d_\phi$ weights the contexts in which the student learns, while $b_\phi$ determines the token targets at each context. Appendix~\ref{app:mechanism-proofs} decomposes the gradient change when the generator is refreshed. Appendix~\ref{app:implementation-objective} gives the corresponding expression for token-averaged training.

\paragraph{Evaluating supervision quality.}
Whether updating the generator provides more useful supervision is an empirical question. Section~\ref{sec:components} tests this by comparing SePT with training on fresh responses from a fixed initial generator. It then examines whether a SePT-trained generator's responses can teach a student initialized from the original model and whether its reasoning prefixes help an unchanged model complete solutions.

\section{Experiments}\label{sec:experiments}
\subsection{Experimental settings}\label{sec:settings}
\textbf{Data and models.}
Our main mathematical training questions come from DeepScaleR (DSR); OpenThoughts-Math (OTM) provides an alternative training set \citep{deepscaler2025,guha2025openthoughts}.
The six evaluation benchmarks are MATH-500, AMC-23, AIME-24, AIME-25, OlympiadBench, and Minerva \citep{hendrycks2021measuringmathematicalproblemsolving,he2024olympiadbenchchallengingbenchmarkpromoting,lewkowycz2022solvingquantitativereasoningproblems}.
We evaluate the starting models in Table~\ref{tab:overall}, including both pretrained and instruction-tuned checkpoints.
Prompt templates and answer extraction remain matched within each comparison. Model-specific configurations are provided in Appendix~\ref{app:study-settings}.

\textbf{SePT training.}
The main mathematical recipe uses AdamW with learning rate $10^{-7}$ for three epochs and generates one response per prompt. Sampling temperature rises linearly to one over the first two epochs and remains there in the third; it starts at 0.6 for Qwen2.5-Math-7B and DeepSeek-Math-7B-Instruct, and at 0.9 for the other models. Training temperature stays at one. Each generated batch is split into two minibatches for two updates before refreshing the generator (Appendix~\ref{app:original-details}).

\textbf{Baseline and reporting.}
We estimate Pass@1, Pass@8 and Pass@32 from 32 responses per question using the formula in Appendix~\ref{app:passk}.
AVG is the arithmetic mean of Pass@1, Pass@8 and Pass@32, with each Pass@$k$ first averaged equally across the six benchmarks. Scores are percentages and differences are percentage points.
We denote evaluation decoding temperature by $\tau_{\mathrm{eval}}$. The \emph{Base Model} represents the starting checkpoint evaluated at $\tau_{\mathrm{eval}}=1$. The \emph{Baseline} is the no-training model at the single swept temperature maximizing AVG.
SePT is reported at its AVG-selected checkpoint and evaluated at $\tau_{\mathrm{eval}}=1$ (Appendix~\ref{app:selection-recovery}). Appendix~\ref{app:finite-budget-scope} defines the decoding reference.

\textbf{Mechanism studies.}
Generator controls use fixed training endpoints and eight evaluation responses per question on Qwen2.5-Math-7B, distinct from Table~\ref{tab:overall}'s selected-model, 32-response evaluation. The frozen SePT generator is obtained after 300 refreshes (600 updates), rather than by AVG selection. Prefix interventions use four continuations per question and completer (Table~\ref{tab:crossed-prefix}). A separate Llama-3.1-8B-Instruct study tests response budgets; see Appendix~\ref{app:study-settings} for details.

\textbf{Uncertainty and overlap.}
Where reported, 95\% confidence intervals are computed by resampling questions jointly across the compared configurations (paired-question bootstrap) and shown in square brackets. Data-order variability is summarized separately by mean $\pm$ sample standard deviation (Appendix~\ref{app:shuffle-seeds}). Appendix~\ref{app:statistics} gives the resampling procedure and inferential scope.
We also provide a DSR overlap audit in Appendix~\ref{app:overlap}.

\suppressfloats[t]
\subsection{SePT can improve over the Baseline}\label{sec:performance}
\begin{table}[t]
\centering\small
\caption{Performance across six mathematical benchmarks. AVG compares the Base Model, its temperature-selected Baseline and SePT. All $\Delta$ columns are SePT minus Baseline. Differences are computed before rounding. Budget-specific differences distinguish first-draw accuracy from sampling coverage. Detailed benchmark Pass@$k$ values appear in Appendix~\ref{app:recovery}.}
\label{tab:overall}
\setlength{\tabcolsep}{3.5pt}
\renewcommand{\arraystretch}{1.08}
\begin{tabular}{@{}lrrrrrrr@{}}\toprule
 & \multicolumn{4}{c}{AVG ($\uparrow$)} & \multicolumn{3}{c}{$\Delta$Pass@$k$ (points)}\\
\cmidrule(lr){2-5}\cmidrule(lr){6-8}
Model & Base & Baseline & \textbf{SePT} & $\Delta$ & $k=1$ & $k=8$ & $k=32$\\\midrule
Qwen2.5-7B & 34.6 & 44.1 & 50.7 & +6.7 & +11.0 & +5.9 & +3.2\\
Qwen2.5-7B-Instruct & 51.6 & 53.6 & 54.1 & +0.5 & -0.3 & +0.2 & +1.7\\
Qwen2.5-Math-7B & 38.2 & 43.7 & 55.0 & +11.4 & +16.8 & +10.4 & +6.9\\
Qwen3-8B-Base & 46.6 & 54.9 & 55.8 & +0.9 & +5.0 & +0.3 & -2.6\\
\addlinespace[3pt]
DeepSeek-Math-7B & 11.7 & 16.8 & 17.2 & +0.4 & +0.0 & -0.2 & +1.2\\
DeepSeek-Math-7B-Instruct & 31.2 & 32.7 & 33.9 & +1.2 & +0.5 & +0.8 & +2.3\\
\addlinespace[3pt]
Llama-3.1-8B & 12.8 & 18.0 & 19.5 & +1.5 & +1.0 & +1.8 & +1.7\\
Llama-3.1-8B-Instruct & 33.9 & 38.9 & 37.8 & -1.2 & -0.4 & -1.1 & -1.9\\
\addlinespace[3pt]
Phi-4-mini-Instruct & 43.4 & 46.3 & 47.0 & +0.8 & -1.5 & +1.2 & +2.6\\
\bottomrule\end{tabular}
\end{table}

Table~\ref{tab:overall} summarizes performance across nine starting models. Eight have higher AVG point estimates than their temperature-selected Baselines. The clearest AVG gains are on Qwen2.5-Math-7B and Qwen2.5-7B: 11.4 and 6.7 points, respectively, with both 95\% confidence intervals entirely above zero. For Qwen2.5-Math-7B, Pass@1 rises from 22.7\% to 39.5\% and Pass@32 from 61.0\% to 67.9\%.

\Needspace{7\baselineskip}
Although the AVG-difference intervals for the other seven models include zero, the budget-specific analysis identifies additional gains. Llama-3.1-8B has a positive AVG point estimate and gains in Pass@1 and Pass@8 whose 95\% intervals are entirely above zero; the negative AVG point estimate occurs for Llama-3.1-8B-Instruct. Qwen3-8B-Base's gain is concentrated in Pass@1, while Phi-4-mini-Instruct has higher Pass@8 and Pass@32 point estimates alongside lower Pass@1. Appendix~\ref{app:recovery} gives the full results, confidence intervals and sampling-budget profiles.

The analysis in Section~\ref{sec:temperature-dynamics} identifies a sufficient condition for local improvement: positive alignment between the self-training update and the gradient of expected task accuracy. Appendix~\ref{app:mechanism-proofs} further shows how heterogeneous changes across questions can produce different gains in Pass@1 and AVG.

For Qwen2.5-Math-7B, an additional evaluation compares SePT with four alternative decoders on a fixed set of questions. SePT outperforms all four (Appendix~\ref{app:decoder-stress}).

Across three data-order runs for Qwen2.5-Math-7B and six for DeepSeek-Math-7B-Instruct, mean AVG is $54.2\pm0.7$ and $33.5\pm0.3$, respectively (mean $\pm$ sample SD). Every listed run exceeds its corresponding Baseline in AVG. Appendix~\ref{app:shuffle-seeds} reports individual runs and their Pass@$k$ scores.

\subsection{Comparison with other training-based methods}\label{sec:training-comparisons}
We compare SePT with two other training paradigms: reward-free EM-FT and verifiable-reward GRPO.
\begin{table}[!htb]
\centering\small
\renewcommand{\arraystretch}{1.05}
\caption{Reward-free training on DSR: SePT versus EM-FT in terms of AVG. The methods use different sampling temperature schedules. Full results are in Appendix~\ref{app:recovery}.}\label{tab:original-grpo}
\setlength{\tabcolsep}{4pt}
\begin{tabular}{@{}lrr@{}}\toprule
Model & SePT & EM-FT\\\midrule
Qwen2.5-Math-7B &55.0&51.3\\
DeepSeek-Math-7B-Instruct &33.9&32.5\\\bottomrule
\end{tabular}
\end{table}
\paragraph{Reward-free entropy minimization.}
EM-FT \citep{agarwal2025unreasonable} minimizes token-level entropy on self-generated contexts, whereas SePT fits sampled tokens by likelihood.
SePT achieves higher AVG on both tested models; see Table~\ref{tab:original-grpo} and Figure~\ref{fig:em}.

\paragraph{Reward-based training.}
We compare SePT with GRPO, which uses verifiable rewards and eight responses per prompt, on both DSR and OTM (Table~\ref{tab:full-training-recipes}). GRPO achieves higher AVG on DSR for all three tested models. On OTM, SePT comes within 1.4 and 0.4 AVG points of GRPO on Qwen-Math and DeepSeek-Math-Instruct, respectively, and exceeds it on Llama-Instruct.

\Needspace{9\baselineskip}
\subsection{Mechanistic analysis of SePT}\label{sec:components}
Section~\ref{sec:method} describes how sampling temperature and generator updates shape the learning signal. Temperature and generator-update controls assess their effects on predictions and learning. Transfer controls test the teaching value of a frozen SePT generator for a student at initial weights; continuation controls test whether further generator updates help an already SePT-trained student. Prefix interventions assess the usefulness of generated reasoning without weight updates. In the training controls, the \emph{generator} supplies responses and the \emph{student} learns from them.

\Needspace{5\baselineskip}
\subsubsection{Temperature control and generator evolution}\label{sec:generator-evolution}
Table~\ref{tab:recipe_controls} compares SePT with two ablations: SePT (Offline), which trains on a fixed dataset generated once by the initial model, and SePT with both sampling and training temperatures fixed at one.
With online refresh retained, the cold-to-unit schedule achieves an AVG of 55.0, compared with 44.6 under constant unit temperature.
\begin{table}[htbp]
\centering\small
\caption{Temperature and data-refresh ablations on Qwen2.5-Math-7B trained on DSR. Offline likelihood-training settings are given in Appendix~\ref{app:offline-protocol}.}
\label{tab:recipe_controls}
\setlength{\tabcolsep}{5pt}
\renewcommand{\arraystretch}{1.05}
\begin{tabular}{@{}lrrrr@{}}\toprule
\textbf{Condition} & \textbf{Pass@1} & \textbf{Pass@8} & \textbf{Pass@32} & \textbf{AVG}\\\midrule
Baseline & 22.7 & 47.3 & 61.0 & 43.7 \\
SePT (Offline) & 21.0 & 50.2 & 65.2 & 45.5 \\
SePT ($\tau_s=\tau_t=1$ throughout) & 19.3 & 50.1 & 64.3 & 44.6 \\
SePT (cold-to-unit schedule) & 39.5 & 57.7 & 67.9 & 55.0 \\
\bottomrule\end{tabular}
\end{table}

To distinguish learned changes from decoding-temperature adjustment, we compare the initial and SePT-trained models' next-token predictions at identical prefixes. Positive decoding temperatures preserve token rankings at a fixed prefix (Section~\ref{sec:temperature-dynamics}). At seven of the 240 tested prefixes, the most likely token changes (Appendix~\ref{app:rank-change}), showing that these learned changes cannot be reproduced by adjusting decoding temperature alone.

\begin{figure}[!htb]
\centering
\includegraphics[width=0.8\linewidth]{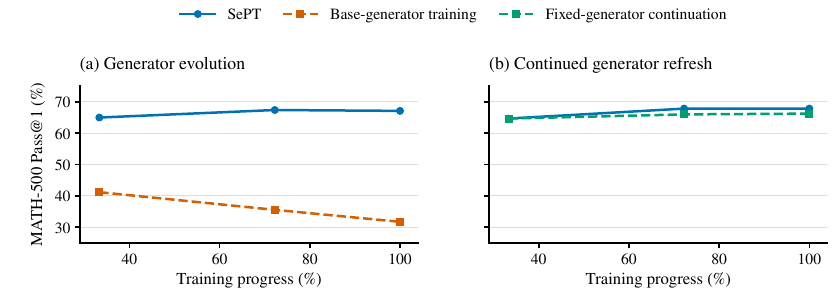}
\caption{Generator-update controls on MATH-500. Both experiments generate fresh training responses each round. (a) Students start from the initial model and learn from either an evolving generator (SePT) or the frozen initial generator. (b) Both students resume from the same SePT checkpoint after 300 refreshes, with identical optimizer state. One continues updating its generator; the other uses a generator frozen at that SePT checkpoint. Both students continue training. Protocols are given in Appendix~\ref{app:source-details}.}
\label{fig:controls}
\end{figure}

Whether the updated generator provides better supervision is a separate empirical question. SePT (Offline) achieves an AVG of 45.5 versus 55.0 for SePT, but also differs in response reuse and optimization settings. To isolate generator updates, \emph{Base-generator training} draws fresh responses from the fixed initial model, with student initialization, prompt stream, temperature schedule and optimizer settings matched to SePT. SePT maintains its gains while this student deteriorates on MATH-500 (Figure~\panelref{fig:controls}{a}). At the final assessment, the six-benchmark macro Pass@1 gap is 20.16 points [17.92,22.57] (Appendix~\ref{app:source-details}).
The early MATH-500 advantage also repeats under an independent matched training seed (Appendix~\ref{app:source-details}).
\par

\begin{samepage}
\subsubsection{Training on SePT-generated responses}\label{sec:source-transfer}
The generator-update control shows that refreshing the source improves learning. In \emph{SePT-generator transfer}, a frozen SePT-trained generator supplies unfiltered responses to a student initialized from the Base Model. This tests whether the acquired responses can teach a student without transferring the generator's weights. Under SePT's sampling temperature schedule, transfer improves MATH-500 Pass@1 by 24.25 points over Base-generator training (Table~\ref{tab:source_studies}, first block). The gain is concentrated in first-draw accuracy, with a much smaller Pass@8 difference (Appendix~\ref{app:source-details}).
\par
\end{samepage}
\begin{table}[!htb]
\centering\small
\caption{Effect of the response generator on student performance. Students train on DSR and are evaluated on MATH-500 with eight responses per question at $\tau_{\mathrm{eval}}=1$. $\theta_0$ denotes the initial model and $\theta_\star$ the SePT checkpoint after 300 refreshes. Within each block, student update counts are matched and evaluation endpoints are fixed. Protocols are given in Appendix~\ref{app:study-settings}.}
\label{tab:source_studies}
\setlength{\tabcolsep}{4pt}
\renewcommand{\arraystretch}{1.05}
\begin{tabular}{@{}lllrr@{}}\toprule
\textbf{Condition} & \textbf{Student starts at} & \textbf{Generator} & \textbf{Pass@1} & \textbf{Pass@8}\\\midrule
\multicolumn{5}{@{}l}{\textit{Training from initial weights: scheduled sampling temperature}}\\
SePT & $\theta_0$ & Evolving student & 65.0 & 87.6\\
Base-generator training & $\theta_0$ & Frozen $\theta_0$ & 40.7 & 85.2\\
SePT-generator transfer & $\theta_0$ & Frozen $\theta_\star$ & 64.9 & 86.4\\\midrule
\multicolumn{5}{@{}l}{\textit{Training from initial weights: fixed sampling temperature $\tau_s=0.3$}}\\
Base-generator training & $\theta_0$ & Frozen $\theta_0$ & 50.2 & 84.6\\
SePT-generator transfer & $\theta_0$ & Frozen $\theta_\star$ & 67.1 & 87.4\\\midrule
\multicolumn{5}{@{}l}{\textit{Continued training: scheduled sampling temperature, primary assessment}}\\
SePT & $\theta_\star$ & Evolving student & 67.8 & 87.4\\
Fixed-generator continuation & $\theta_\star$ & Frozen $\theta_\star$ & 66.0 & 87.4\\
\bottomrule\end{tabular}
\end{table}

Could cooling the initial generator provide equally useful training responses? We sample the frozen initial and SePT generators at $\tau_s=0.3$, using the temperature selected for the Baseline and matching student initialization, response counts and optimization steps. At this shared temperature, the student trained on SePT responses achieves 16.95 points higher MATH-500 Pass@1 than the student trained on initial-model responses (Table~\ref{tab:source_studies}, second block). To characterize the training responses used by these students, we audit responses from 1,024 matched DSR encounters. At the same sampling temperature of 0.3, the SePT generator produces more correct answers (49.41\% versus 37.21\%), more extractable answers (95.12\% versus 81.05\%), and fewer length-truncated responses (Appendix~\ref{app:source-decoder-transfer}).

\begin{samepage}
The continuation control asks whether generator updates remain beneficial for an already SePT-trained student. Both students resume from the same SePT checkpoint after 300 refreshes, with identical optimizer state. SePT continues updating its generator; \emph{Fixed-generator continuation} freezes the generator at that checkpoint while continuing to train the student (Table~\ref{tab:source_studies}, final block; Figure~\panelref{fig:controls}{b}). Continued refresh gives a modest additional gain at the primary assessment, while the frozen generator sustains much of the benefit (Appendix~\ref{app:source-details}).
\par
\end{samepage}

\begin{samepage}
\subsubsection{SePT prefixes help an unchanged model}\label{sec:prefix-utility}
Section~\ref{sec:source-transfer} shows that SePT-generated responses can train a student from initial weights. We test their usefulness without weight updates by supplying reasoning prefixes as context for solution completion. A \emph{donor} generates the beginning of a response, and a \emph{completer} receives the question and this prefix to finish the solution. We use the Base Model, its temperature-selected Baseline and SePT in both roles, reusing each prefix across completers without correctness-based selection. Holding the completer fixed lets us compare the usefulness of the different donors' prefixes.
\par
\end{samepage}
\begin{table}[!htb]
\centering\small
\caption{Prefix effects on 107 MATH-500 questions eligible for all three donors. SePT uses checkpoint $\theta_\star$ from Table~\ref{tab:source_studies}. In both donor and completer roles, Base Model and SePT use $\tau_{\mathrm{eval}}=1$, while Baseline uses $\tau_{\mathrm{eval}}=0.3$. Each cell reports answer accuracy (\%) over four continuations per question, except the no-prefix reference, which uses eight. Full protocols are given in Appendix~\ref{app:source-content}.}\label{tab:crossed-prefix}\label{tab:baseline-prefix}
\setlength{\tabcolsep}{8pt}
\renewcommand{\arraystretch}{1.05}
\begin{subtable}{\linewidth}
\centering
\captionsetup{position=top,labelfont=bf,textfont=bf,skip=4pt,hypcap=true}
\caption{Crossed donors and completers}\label{tab:crossed-prefix-donors}
\begin{tabular}{@{}lrrr@{}}\toprule
 & \multicolumn{3}{c}{Continuation model}\\\cmidrule(lr){2-4}
Prefix donor & Base Model ($\tau_{\mathrm{eval}}=1$) & Baseline ($\tau_{\mathrm{eval}}=0.3$) & SePT ($\tau_{\mathrm{eval}}=1$)\\\midrule
Base Model &35.98 &40.42 &40.65\\
Baseline &47.43 &55.37 &54.21\\
SePT &55.14 &69.16 &67.29\\\bottomrule
\end{tabular}
\end{subtable}
\par\medskip
\begin{subtable}{\linewidth}
\centering
\captionsetup{position=top,labelfont=bf,textfont=bf,skip=4pt,hypcap=true}
\caption{Prefix controls with the Baseline completer}\label{tab:crossed-prefix-length}
\begin{tabular}{@{}lr@{}}\toprule
Prefix source & Continuation accuracy\\\midrule
No supplied prefix & 57.83\\
Baseline-generated prefix & 55.37\\
SePT-generated prefix, length-matched & 68.46\\\bottomrule
\end{tabular}
\end{subtable}
\end{table}
\Needspace{5\baselineskip}
Table~\subpanelref{tab:crossed-prefix}{tab:crossed-prefix-donors} shows that SePT prefixes give the highest observed continuation accuracy with every completer, including both decoding settings of the initial model. The Baseline donor tests whether cooling the initial model can supply equally useful prefixes. Cooling helps, but with the Baseline completer fixed, SePT prefixes raise accuracy from 55.37\% to 69.16\%. Thus the learned generator supplies useful context beyond what the initial model produces at its selected decoding temperature.

The full-prefix gain is concentrated on questions where the Baseline donor supplies shorter prefixes. To test whether additional prefix length explains this advantage, we shorten the SePT prefixes to the corresponding Baseline lengths while retaining all 107 questions and the same Baseline completer. At matched lengths, SePT prefixes retain a 13.08-point advantage over Baseline prefixes, showing that longer prefixes alone do not explain the improvement (Table~\subpanelref{tab:crossed-prefix}{tab:crossed-prefix-length}).

To establish a benefit beyond replacing Baseline prefixes, we compare with answering from scratch. The same Baseline completer scores 57.83\% without a prefix, versus 55.37\% with its own prefixes. Length-matched SePT prefixes reach 68.46\%, improving on independent answering by 10.63 points [4.44,17.17] (Appendix~\ref{app:baseline-prefix}).

Together, these experiments show that SePT-generated responses can both train a student from initial weights and help an unchanged model complete solutions.

\subsection{Generality beyond mathematics}\label{sec:generality}
To assess capability retention after mathematical training, we evaluate Qwen-Math SePT on five general-capability benchmarks. Its average score is 35.0, compared with 34.9 for the initial model, with small task-level changes in either direction (Appendix~\ref{app:retention}).

To test learning from non-mathematical data, we train Qwen2.5-7B on ARC, LogiQA and StrategyQA questions and evaluate transfer to LongBench's 2WikiMultihopQA, HotpotQA and MuSiQue tasks \citep{bai2024longbench}. Macro F1 rises from 11.67 for the initial model to 14.44 after SePT. All three tasks improve, and the macro gain has a paired-question 95\% confidence interval of [1.55,4.09] F1 points. Appendix~\ref{app:nonmath} gives task-level results and evaluation details.

For code generation, we train Qwen3-4B-Instruct-2507 on rStar-Coder \citep{liu2025rstarcoder} and evaluate on the LiveCodeBench v6 increment \citep{jain2024livecodebench} and MBPP+ \citep{liu2023evalplus}. The sampling temperature starts above one and decreases toward one. CodeScore averages Pass@1 and Pass@5 across both benchmarks, using five responses per question. SePT achieves a 1.33-point CodeScore gain over the temperature-selected Coding Baseline, with a 1.19-point gain in a fixed-duration repeat. Both runs yield positive point estimates for short-duration adaptation. Longer runs highlight training-duration sensitivity, discussed in Section~\ref{sec:boundaries}. Appendix~\ref{app:coding} reports detailed scores, confidence intervals, protocols and trajectories.

\subsection{Robustness and scope}\label{sec:boundaries}
Section~\ref{sec:performance} reports sensitivity to training-data order and alternative decoding settings. Overlap between training and evaluation data could inflate measured gains. On MATH-500 questions unflagged by the overlap audit, SePT-generator transfer retains a 24.71-point Pass@1 advantage over Base-generator training. Separately, training Qwen-Math on the lower-overlap OTM dataset yields an AVG of 55.2, compared with 55.0 on DSR (Table~\ref{tab:full-training-recipes}). Appendix~\ref{app:overlap} details the audit, subgroup results and a further run trained after removing all candidate overlaps.

In the Llama-Instruct generator control, 410 of 4,000 responses from Base-generator training reach the length limit, compared with only eight from SePT. This raises the possibility that the performance gap reflects answers cut off before completion. We therefore continue all truncated responses in both groups, increasing the total allowance from 3,072 to 6,144 tokens. SePT retains a 9.38-point Pass@1 advantage over Base-generator training. Even assigning every still-truncated SePT response an incorrect outcome and every still-truncated control response a correct outcome leaves a positive gap (Appendix~\ref{app:source-budget}).

The prefix experiments demonstrate that SePT-generated reasoning helps an unchanged model reach correct answers, including at matched prefix lengths (Section~\ref{sec:prefix-utility}). Their focus is the usefulness of the supplied context for completing solutions. Verifying every intermediate step and excluding all semantic answer information lie beyond this evaluation's scope (Appendix~\ref{app:baseline-prefix}).

Because SePT learns from unfiltered responses, we test whether fitting an incorrect answer can coexist with useful transfer. We deliberately corrupt the final answers in otherwise unchanged responses. Training on these responses increases both the likelihood of the incorrect endings on training questions and the likelihood of pre-answer content on held-out questions. Fitting full responses produces greater pre-answer transfer than fitting only their answer-bearing tails (Appendix~\ref{app:corruption}). This transfer persists when fitting pre-answer content alone, which also reduces the likelihood of corrupted endings on training questions. This measures likelihood transfer, not an improvement in answer accuracy, and complements observations of cross-problem improvement despite reinforcing incorrect training answers in intrinsic-reward RLVR \citep{he2026far}. The behavior of natural misconceptions remains an open question.

Training duration also affects the benefit of self-generated supervision. Our coding runs improve early but deteriorate with extended training (Appendix~\ref{app:coding-hot}). Related work reports entropy collapse accompanied by performance saturation in RL \citep{cui2025entropy}, and early gains followed by degradation in intrinsic-reward RLVR \citep{he2026far}. These findings motivate studying how to sustain useful adaptation through sampling and optimization choices. Whether the same mechanisms explain SePT's coding trajectories remains open.

\section{Discussion and related work}\label{sec:discussion}
The generator controls distinguish the value of learned weights from that of their responses. Resetting the student to its initial weights shows that useful supervision can pass through unfiltered responses alone, establishing a role for SePT's evolving training data (Section~\ref{sec:components}).

This perspective builds on self-training and same-architecture distillation \citep{wang2022self,furlanello2018born}. STaR uses known answers to select or rationalize responses \citep{zelikman2022star}, TTRL derives consensus rewards \citep{zuo2025ttrl}, and on-policy distillation supplies teacher feedback on student-generated contexts \citep{agarwal2024onpolicy}. SePT learns from all sampled responses without these quality signals (Table~\ref{tab:methods}). Our frozen-generator transfer study uses distillation to test the resulting data's value, rather than adding a stage to SePT. Its cost includes training the generator and student (Appendix~\ref{app:source-details}).

Entropy minimization offers a closely related reward-free approach \citep{agarwal2025unreasonable}. At unit training temperature, a sampling temperature slightly below one aligns the sequence-sum reference's mean logit update with entropy minimization to first order (Appendix~\ref{app:reference-proofs}). The objectives nevertheless differ: SePT fits sampled tokens, while entropy minimization acts on the student's full next-token distribution. Agarwal et al.'s stronger gains on Qwen than Llama likewise motivate studying which starting models support useful self-training.

Related work amplifies existing preferences through spurious rewards \citep{shao2025spurious} and model-derived quality feedback \citep{huang2024self}. SePT fits unfiltered sampled responses. Fixed-prefix rank changes distinguish its updates from scalar decoding-temperature adjustment, while transfer experiments establish the usefulness of its responses. Sequence-power sampling also exploits existing preferences \citep{karan2025reasoning}, motivating comparisons beyond tokenwise temperature sweeps. Reporting both first-draw accuracy and coverage helps assess these approaches \citep{yue2025does}.

\begin{samepage}
\section{Conclusion}
SePT demonstrates that unfiltered self-generated responses can support reasoning improvements beyond a temperature-tuned no-training baseline. The generator and prefix interventions show that the trained model's responses provide useful content for both learning and solution completion. Together, these findings establish that reward-free self-training can improve not only a model's predictions but also the supervision it supplies. A promising next step is to identify which starting models and sampling schedules sustain this process across tasks and longer training runs.
\label{main-end}
\par
\end{samepage}

\bibliographystyle{unsrtnat}
\bibliography{main}
\clearpage\appendix
\raggedbottom
\makeatletter\setlength{\@fptop}{0pt}\makeatother
\section{Experimental settings and model-selection details}\label{app:study-settings}
This appendix specifies the evaluation metrics, model interfaces and generator-control protocols. The common training recipe is in Appendix~\ref{app:original-details}.

\subsection{Mathematical evaluation}

\paragraph{Pass@$k$ estimator.}\label{app:passk}
For a question with $c$ correct responses among $n$ sampled responses, we use the estimator of \citet{chen2021evaluating}:
\[
\widehat{\mathrm{Pass@}k}=1-\frac{\binom{n-c}{k}}{\binom nk},\qquad 1\le k\le n,
\]
with $\binom{n-c}{k}=0$ when $n-c<k$.
Thus Pass@1 is $c/n$ and Pass@$n$ is the indicator of at least one correct response.
We average questions within each benchmark and then weight the six benchmark means equally.
AVG then averages the three budget means for the same selected configuration.

\begin{table}[ht]
\centering\small
\caption{Mathematical evaluation settings for every model in Table~\ref{tab:overall}. Instruction placement and answer format are matched between Baseline and SePT. SePT is evaluated at $\tau_{\mathrm{eval}}=1$.}\label{tab:official-selections}
\setlength{\tabcolsep}{4pt}
\begin{tabular}{@{}lllr@{}}\toprule
Model & Instruction in & Final answer & Baseline $\tau_{\mathrm{eval}}$\\\midrule
Qwen2.5-7B & System message & Boxed &0.5\\
Qwen2.5-7B-Instruct & User message & Boxed &0.9\\
Qwen2.5-Math-7B & System message & Boxed &0.3\\
Qwen3-8B-Base & User message & Boxed &0.6\\\midrule
DeepSeek-Math-7B & User message & Unrestricted &0.6\\
DeepSeek-Math-7B-Instruct & System message & Boxed &0.6\\\midrule
Llama-3.1-8B & User message & Unrestricted &0.6\\
Llama-3.1-8B-Instruct & System message & Boxed &0.5\\\midrule
Phi-4-mini-Instruct & User message & Boxed &0.2\\\bottomrule
\end{tabular}
\end{table}
All prompts request step-by-step reasoning. ``Boxed'' requests a final answer in \texttt{\textbackslash boxed\{\}}; ``unrestricted'' imposes no answer delimiter. Instruction placement follows Table~\ref{tab:official-selections}, with model-specific chat templates detailed in Appendix~\ref{app:original-details}. The Llama-Instruct generator control instead uses the unrestricted user prompt in Appendix~\ref{app:source-budget}.

\paragraph{Model and temperature selection.}\label{app:selection-recovery}
The mathematical experiments use the three-epoch training protocol in Appendix~\ref{app:original-details}. The Baseline decoding temperature and SePT model are selected by benchmark AVG, with one configuration supplying all reported sampling budgets.
Mathematical positive-temperature candidates are spaced by 0.1, with model-specific ranges shown in Appendix~\ref{app:original-details}. Coding uses a separate candidate grid and selection metric, specified in Appendix~\ref{app:coding}.

\subsection{Source identities and controlled endpoints}
Let $\theta_0$ denote the initial Qwen2.5-Math-7B model and $\theta_\star$ the SePT generator after 300 refreshes (600 optimizer updates). This fixed checkpoint differs from the AVG-selected model in the main performance comparison. Each refresh generates 128 responses for two updates on 64-response minibatches. Table~\ref{tab:source-design} distinguishes the student and generator in each control.

\begin{table}[ht]\centering\small
\caption{Generator-control design on Qwen2.5-Math-7B. MATH-500 assessments use 500 questions, eight responses per question and evaluation temperature one.}\label{tab:source-design}
\setlength{\tabcolsep}{4pt}\renewcommand{\arraystretch}{1.12}
\begin{tabularx}{\linewidth}{@{}>{\raggedright\arraybackslash}p{.21\linewidth}>{\raggedright\arraybackslash}p{.17\linewidth}>{\raggedright\arraybackslash}X>{\raggedright\arraybackslash}p{.21\linewidth}@{}}\toprule
\textbf{Comparison} & \textbf{Student starts at} & \textbf{Response generators} & \textbf{Assessment refreshes}\\\midrule
Generator evolution, Figure~\panelref{fig:controls}{a} & $\theta_0$, fresh optimizer & Evolving student versus fixed $\theta_0$ & 300, 650, 900; six-benchmark check at 900\\
\addlinespace
Independent seed and transfer, Table~\ref{tab:source_studies} & $\theta_0$, fresh optimizer & Evolving student, fixed $\theta_0$, or fixed $\theta_\star$ & Fixed endpoint 300 (600 updates)\\
\addlinespace
Continued refresh, Figure~\panelref{fig:controls}{b} & $\theta_\star$, shared full optimizer state & Evolving student versus fixed $\theta_\star$ & Primary 650; secondary 900, counted from initial training\\
\addlinespace
Common sampling temperature, Table~\ref{tab:source-decoder-endpoints} & $\theta_0$, fresh optimizer & Fixed $\theta_0$ versus fixed $\theta_\star$, both at $\tau_s=0.3$ & Fixed endpoint 300 (600 updates)\\\bottomrule
\end{tabularx}
\end{table}
The first three comparisons retain SePT's sampling temperature schedule. The independent-seed study fixes its endpoint in advance; intermediate $n=4$ diagnostics do not select it. The two figure panels use separate evaluation responses. Continued refresh tests a prespecified two-point Pass@1 materiality/equivalence margin. The Llama response-budget intervention has a different prompt and is described with its results in Appendix~\ref{app:source-budget}.

\subsection{Uncertainty estimates}\label{app:statistics}
Square brackets following an estimate give its paired-question bootstrap 95\% confidence interval. Question-bootstrap analyses retain the compared configurations and all response draws for each question, resampling questions within each benchmark. For selected-model comparisons, the selected checkpoint and decoding temperature are held fixed during resampling. Intervals are pointwise unless a simultaneous adjustment is explicitly specified. This procedure neither reruns model selection nor estimates between-run variability; Appendix~\ref{app:shuffle-seeds} separately summarizes observed variation across data-order training runs. The corruption assay instead resamples its counterbalanced question pairs, and Appendix~\ref{app:crossed-prefix} additionally reports adjusted 97.5\% intervals for two primary contrasts.

\subsection{Data sources and training recipe}
\begingroup
The principal mechanism model is \href{https://huggingface.co/Qwen/Qwen2.5-Math-7B}{Qwen/Qwen2.5-Math-7B}.
DSR uses \href{https://huggingface.co/datasets/agentica-org/DeepScaleR-Preview-Dataset}{DeepScaleR-Preview-Dataset}.
OTM denotes math questions from \href{https://huggingface.co/datasets/open-thoughts/OpenThoughts3-1.2M}{OpenThoughts3-1.2M}, not its full mixed-domain response dataset.
The evaluation benchmarks are listed in Section~\ref{sec:settings}. Appendix~\ref{app:overlap} reports the training--evaluation overlap audit. Starting-model families include Qwen \citep{yang2024qwen2}, DeepSeek-Math \citep{shao2024deepseekmath} and Llama 3 \citep{llama3}.
\par\endgroup

\begin{table}[ht]\centering\small
\caption{Shared protocol for the independent-seed and generator-transfer studies in Table~\ref{tab:source-design}. Each question encounter supplies one fresh response, used without correctness filtering.}\label{tab:source-recipe}
\begin{tabularx}{\linewidth}{lX}\toprule
Setting & Specification\\\midrule
Preprocessing & 40,315 DSR rows; 40,306 retained by the 1,024-token prompt filter; identical matched question order.\\
Prompt & System: ``Please reason step by step, and put your final answer within \texttt{\textbackslash boxed\{\}}.'' User: problem; model tokenizer chat rendering.\\
Training & One raw response per encounter; 128 responses/refresh; two 64-response AdamW updates; 3,072-token response cap.\\
Schedule & LR $10^{-7}$, 10-refresh warmup then constant; sampling temperature $0.6+0.4(r-1)/628$ at refresh $r$ through this endpoint; training temperature one. \\
Sources & SePT refreshes its generator; Base-generator training retains the initial generator; SePT-generator transfer freezes the acquired generator and starts the student from the initial model with a fresh optimizer.\\
Evaluation & Three evenly spaced diagnostic MATH-500 panels with $n=4$; separate primary $n=8$ endpoint, temperature one; no endpoint reselection; paired question intervals conditional on checkpoints.\\
Comparison to Table~\ref{tab:overall} & Six-suite $n=32$, one AVG-selected checkpoint or no-training temperature for all reported budgets; distinct from the fixed endpoint.\\\bottomrule
\end{tabularx}
\end{table}

\subsection{Scoring conventions}
Answer correctness requires successful extraction and verifier acceptance. Diagnostic analyses also retain verifier verdicts without the extraction requirement. Correctness conditional on extraction is a ratio of these outcomes. Separately generated evaluations may produce different responses even with matched random seeds. Appendix~\ref{app:source-budget} specifies the continuation procedure for responses reaching the length limit.

\subsection{Inference allowances and the decoding reference}\label{app:finite-budget-scope}
The main comparison measures improvement under the stated per-question response and sampling allowances, with model-specific prompts and answer checking matched within each pair.
Let $\delta$ collect decoder settings and let $\mathcal B$ specify the sampling and response limits.
The no-training reference selects one configuration from the recorded finite family:
\[
\widehat\delta\in\mathop{\rm argmax}_{\delta\in\mathcal D_{\rm tested}}
\widehat{\rm AVG}(\theta_0,\delta;\mathcal B).
\]
All reported sampling budgets use that same selected configuration.
The reference covers the tested decoder family at the stated budget; additional temperatures, filtering settings or inference-time search may yield stronger comparisons.
In particular, sequence-level search can access distributions different from locally temperature-scaled conditionals \citep{karan2025reasoning}.

Matching response allowances does not imply identical realized tokens, FLOPs or wall time.
Appendix~\ref{app:original-compute} gives per-round recipe FLOPs; the accuracy comparisons do not optimize total training, selection and deployment cost.

\paragraph{Local decoder-family observation.}
At a fixed context, suppose the initial logits have a unique maximum token $a_0$.
Every positive scalar temperature preserves that maximum.
Standard top-$k$ and top-$p$ filtering retain the highest-ranked token and preserve the order of surviving tokens after normalization.
Hence any composition of these operations, without additional logit modifications, still has $a_0$ as its maximum; mixtures of such distributions satisfy $q(a_0)\ge q(a)$ for every token $a$.
A robust acquired-model top-token reversal therefore rejects exact conditional equivalence to this local decoder family at that context, beyond a sampled grid of scalar temperatures.
This applies to the seven recorded robust reversals in Appendix~\ref{app:rank-change} under the stated assumptions.
The argument concerns conditional token ordering, rather than an upper bound on task accuracy; arbitrary logit biases, constraints and lookahead search lie outside this decoder family.
\subsection{Robustness to additional decoding settings}\label{app:decoder-stress}
To test additional decoding choices, we evaluate Qwen2.5-Math-7B on 128 MATH-500 questions with eight responses per configuration. Questions and decoder settings are fixed before generation. The initial-model candidates add top-$p$, top-$k$, their combination, and a neighboring temperature to the Baseline. SePT uses the AVG-selected checkpoint from Table~\ref{tab:overall} at $\tau_{\mathrm{eval}}=1$. Prompts, answer checking and the 3,072-token response allowance are matched, with all responses retained.

\begin{table}[htbp]
\centering\small
\caption{Standalone decoder comparison on the fixed 128-question panel. $\tau_{\mathrm{eval}}$ denotes evaluation decoding temperature. Scores are percentages. Mean tokens and cap counts describe the 1,024 responses in each row; ``none'' means no candidate filtering.}\label{tab:decoder-stress}
\setlength{\tabcolsep}{5pt}
\begin{tabular}{@{}llrrrrr@{}}\toprule
Model & Filtering & $\tau_{\mathrm{eval}}$ & Pass@1 & Pass@8 & Mean tokens & At cap\\\midrule
Initial & none &0.30&53.91&79.69&1,094.5&179\\
Initial & top-$p=0.95$ &0.30&52.64&78.91&1,078.3&173\\
Initial & top-$k=50$ &0.30&53.42&79.69&1,110.2&187\\
Initial & both &0.30&54.10&77.34&1,095.5&182\\
Initial & none &0.25&55.27&78.12&1,062.3&159\\\midrule
SePT & none &1.00&\textbf{75.39}&\textbf{87.50}&696.2&16\\\bottomrule
\end{tabular}
\end{table}

The panel's prespecified reference criterion is the mean of Pass@1 and Pass@8, rather than the six-benchmark AVG used for the main evaluation.
The full-support initial decoder at $\tau_{\mathrm{eval}}=0.3$ remains the highest-scoring reference under this criterion.
Relative to it, SePT gains 21.48 points in Pass@1 (paired 95\% interval: 15.23 to 27.83) and 7.81 points in Pass@8 (0.78 to 14.84).
The $\tau_{\mathrm{eval}}=0.25$ candidate has the highest initial-model Pass@1, but SePT still exceeds it by 20.12 points (13.96 to 26.46).
SePT's point gains over all five initial decoders range from 20.12 to 22.75 points in Pass@1 and from 7.81 to 10.16 points in Pass@8; each corresponding paired interval is positive.
The gains persist across all tested decoders on this 128-question panel. The question-bootstrap intervals hold configurations fixed and do not adjust for selecting the best observed reference.

SePT produces shorter responses and reaches the common limit less often (Table~\ref{tab:decoder-stress}). This diagnostic generated 6.28 million tokens and used 0.59 A800 GPU-hours, including model loading but excluding prior training and selection.

\paragraph{Answer delivery and correctness.}
We separately measure answer extractability and correctness among extractable answers.
The selected initial decoder yields 833 extractable answers out of 1,024, of which 552 are correct; SePT yields 1,008 extractable answers, of which 772 are correct.
Thus extractability rises from 81.35\% to 98.44\%, while correctness among extractable responses rises from 66.27\% to 76.59\%.
These conditioning sets differ, so this is not a causal mediation analysis.
An optimistic saved-sample bound that makes every incorrect length-censored response correct, leaving the rest unchanged, reaches at most 71.58\% Pass@1 across the five initial profiles.
Repairing only unextractable responses analogously reaches at most 73.63\%, both below SePT's observed 75.39\%.
These saved-sample bounds apply to each repair separately, not their combination or other decoders.

\section{Complete mathematical results and sampling budgets}\label{app:recovery}
\paragraph{Model-specific evaluation settings.}
All nine models use 32 responses per question and the matched prompts in Table~\ref{tab:official-selections}. Tables~\ref{tab:family-qwen}--\ref{tab:family-phi} give benchmark-level results for the Base Model, Baseline and SePT.

\paragraph{Paired uncertainty across all nine models.}
Table~\ref{tab:math-selected-avg} reports fixed-selection uncertainty for every model in Table~\ref{tab:overall}. Each comparison uses 1,545 matched questions: 500 MATH-500, 40 AMC23, 30 AIME 2024, 30 AIME 2025, 673 OlympiadBench and 272 Minerva. The AMC results use AMC23 only. We resample paired questions independently within each benchmark with 20,000 replicates, following Appendix~\ref{app:statistics}.

\begin{table}[ht]
\centering\small
\caption{Aggregate mathematical performance for all nine starting models. The difference is SePT minus Baseline in percentage points; square brackets give paired-question bootstrap 95\% confidence intervals.}\label{tab:math-selected-avg}
\setlength{\tabcolsep}{5pt}
\begin{tabular}{lrrr}\toprule
Model & Baseline AVG & SePT AVG & Difference [95\% CI]\\\midrule
Qwen2.5-7B &44.072&50.741&$+6.669$ [$+4.615$,$+8.775$]\\
Qwen2.5-7B-Instruct &53.587&54.104&$+0.517$ [$-1.198$,$+2.242$]\\
Qwen2.5-Math-7B &43.664&55.030&$+11.366$ [$+8.486$,$+14.220$]\\
Qwen3-8B-Base &54.850&55.766&$+0.916$ [$-0.910$,$+2.729$]\\
DeepSeek-Math-7B &16.849&17.203&$+0.354$ [$-1.841$,$+2.577$]\\
DeepSeek-Math-7B-Instruct &32.697&33.903&$+1.206$ [$-0.647$,$+3.072$]\\
Llama-3.1-8B &17.992&19.480&$+1.488$ [$-0.186$,$+3.059$]\\
Llama-3.1-8B-Instruct &38.919&37.751&$-1.168$ [$-3.141$,$+0.840$]\\
Phi-4-mini-Instruct &46.272&47.048&$+0.775$ [$-0.862$,$+2.469$]\\
\bottomrule\end{tabular}
\end{table}

Qwen2.5-7B and Qwen2.5-Math-7B have AVG-gain intervals entirely above zero, supporting their gains over the temperature-selected Baseline. The remaining seven intervals include zero; Llama-3.1-8B-Instruct has a negative AVG-difference point estimate.

\paragraph{Sampling-budget profiles.}
An exploratory analysis of the same 13,905 paired question records resolves Pass@1, Pass@8 and Pass@32 using the same selected configuration across budgets. Table~\ref{tab:budget-profiles} complements aggregate AVG with the gains at each sampling budget.

\begin{table}[ht]
\centering\footnotesize
\caption{Sampling-budget profiles for all nine starting models at the selected configurations. Values are SePT minus Baseline in percentage points; square brackets give paired-question bootstrap 95\% confidence intervals. The lower panel compares the first-draw gain with the coverage gain using the same paired resamples.}\label{tab:budget-profiles}
\setlength{\tabcolsep}{2pt}
\begin{tabular}{lrrr}\toprule
Model & $\Delta$ Pass@1 & $\Delta$ Pass@8 & $\Delta$ Pass@32\\\midrule
Qwen2.5-7B & \begin{tabular}[c]{@{}r@{}}$+10.977$\\{}[$+9.636$,$+12.415$]\end{tabular} & \begin{tabular}[c]{@{}r@{}}$+5.852$\\{}[$+3.935$,$+7.857$]\end{tabular} & \begin{tabular}[c]{@{}r@{}}$+3.180$\\{}[$-0.921$,$+7.418$]\end{tabular}\\[3pt]
Qwen2.5-7B-Instruct & \begin{tabular}[c]{@{}r@{}}$-0.298$\\{}[$-1.033$,$+0.413$]\end{tabular} & \begin{tabular}[c]{@{}r@{}}$+0.168$\\{}[$-1.560$,$+1.814$]\end{tabular} & \begin{tabular}[c]{@{}r@{}}$+1.680$\\{}[$-2.052$,$+5.413$]\end{tabular}\\[3pt]
Qwen2.5-Math-7B & \begin{tabular}[c]{@{}r@{}}$+16.807$\\{}[$+14.013$,$+19.602$]\end{tabular} & \begin{tabular}[c]{@{}r@{}}$+10.389$\\{}[$+6.781$,$+14.024$]\end{tabular} & \begin{tabular}[c]{@{}r@{}}$+6.902$\\{}[$+3.196$,$+10.756$]\end{tabular}\\[3pt]
Qwen3-8B-Base & \begin{tabular}[c]{@{}r@{}}$+5.003$\\{}[$+4.048$,$+5.999$]\end{tabular} & \begin{tabular}[c]{@{}r@{}}$+0.330$\\{}[$-1.156$,$+1.829$]\end{tabular} & \begin{tabular}[c]{@{}r@{}}$-2.585$\\{}[$-6.457$,$+1.281$]\end{tabular}\\[3pt]
DeepSeek-Math-7B & \begin{tabular}[c]{@{}r@{}}$+0.018$\\{}[$-0.350$,$+0.412$]\end{tabular} & \begin{tabular}[c]{@{}r@{}}$-0.194$\\{}[$-1.950$,$+1.615$]\end{tabular} & \begin{tabular}[c]{@{}r@{}}$+1.238$\\{}[$-3.503$,$+6.070$]\end{tabular}\\[3pt]
DeepSeek-Math-7B-Instruct & \begin{tabular}[c]{@{}r@{}}$+0.524$\\{}[$-0.005$,$+1.094$]\end{tabular} & \begin{tabular}[c]{@{}r@{}}$+0.842$\\{}[$-0.865$,$+2.545$]\end{tabular} & \begin{tabular}[c]{@{}r@{}}$+2.252$\\{}[$-1.607$,$+6.232$]\end{tabular}\\[3pt]
Llama-3.1-8B & \begin{tabular}[c]{@{}r@{}}$+1.010$\\{}[$+0.664$,$+1.354$]\end{tabular} & \begin{tabular}[c]{@{}r@{}}$+1.760$\\{}[$+0.270$,$+3.179$]\end{tabular} & \begin{tabular}[c]{@{}r@{}}$+1.694$\\{}[$-1.901$,$+5.131$]\end{tabular}\\[3pt]
Llama-3.1-8B-Instruct & \begin{tabular}[c]{@{}r@{}}$-0.447$\\{}[$-1.139$,$+0.212$]\end{tabular} & \begin{tabular}[c]{@{}r@{}}$-1.117$\\{}[$-2.796$,$+0.619$]\end{tabular} & \begin{tabular}[c]{@{}r@{}}$-1.940$\\{}[$-6.164$,$+2.252$]\end{tabular}\\[3pt]
Phi-4-mini-Instruct & \begin{tabular}[c]{@{}r@{}}$-1.540$\\{}[$-2.617$,$-0.445$]\end{tabular} & \begin{tabular}[c]{@{}r@{}}$+1.250$\\{}[$-0.497$,$+2.987$]\end{tabular} & \begin{tabular}[c]{@{}r@{}}$+2.615$\\{}[$-0.599$,$+5.933$]\end{tabular}\\[3pt]
\bottomrule\end{tabular}
\medskip
\begin{tabular}{lr}\toprule
Model & $D_{1,32}=\Delta_1-\Delta_{32}$ [95\% CI]\\\midrule
Qwen2.5-7B & $+7.797$ [$+3.552$,$+11.884$]\\
Qwen2.5-7B-Instruct & $-1.978$ [$-5.687$,$+1.714$]\\
Qwen2.5-Math-7B & $+9.904$ [$+5.849$,$+13.832$]\\
Qwen3-8B-Base & $+7.588$ [$+3.780$,$+11.361$]\\
DeepSeek-Math-7B & $-1.220$ [$-5.844$,$+3.313$]\\
DeepSeek-Math-7B-Instruct & $-1.728$ [$-5.551$,$+2.029$]\\
Llama-3.1-8B & $-0.684$ [$-4.007$,$+2.767$]\\
Llama-3.1-8B-Instruct & $+1.493$ [$-2.582$,$+5.575$]\\
Phi-4-mini-Instruct & $-4.156$ [$-7.506$,$-0.871$]\\
\bottomrule\end{tabular}
\end{table}

The paired question draws retain both arms and all three budget coordinates together, then average six benchmark means equally.
The direct $D_{1,32}$ interval is computed from the coordinate difference within each shared draw, not from independent marginal errors.
Qwen3's improvement is concentrated in Pass@1, with a negative Pass@32 point estimate.
Llama-3.1-8B has positive Pass@1 and Pass@8 differences with both conditional 95\% intervals entirely above zero, as well as positive AVG and Pass@32 point estimates. These results concern the base checkpoint, distinct from Llama-3.1-8B-Instruct and its separate mechanism panel.
Phi has a negative Pass@1 difference despite positive higher-budget point estimates. DeepSeek-Math's first-draw point estimate is nearly unchanged.
These contrasts use each method's reported decoder, so they measure the combined effect of model and decoding choices. AVG combines Pass@1 with nonlinear coverage measures; heterogeneous changes across questions can therefore improve one aggregate without improving another (Appendix~\ref{app:mechanism-proofs}).

\subsection{Sensitivity to training-data order}\label{app:shuffle-seeds}
We evaluate three DSR training runs for Qwen2.5-Math-7B and six for DeepSeek-Math-7B-Instruct, including those in Table~\ref{tab:overall}. Only the data-shuffling seed is varied, with the recipe, AVG selection rule and evaluation protocol fixed within each model. Table~\ref{tab:shuffle-seeds} reports all nine runs.

\begin{table}[htbp]
\centering\small
\caption{Data-order sensitivity on DSR. Each run uses a different data-shuffling seed and the same AVG checkpoint-selection rule. Pass@$k$ is the six-benchmark mean. The reported run is the one in Table~\ref{tab:overall}; the remaining rows include all additional runs in this analysis. Summary rows give mean $\pm$ sample standard deviation across runs, not confidence intervals.}
\label{tab:shuffle-seeds}
\setlength{\tabcolsep}{5pt}
\begin{tabular}{@{}lrrrr@{}}\toprule
Run & Pass@1 & Pass@8 & Pass@32 & AVG\\\midrule
\multicolumn{5}{@{}l}{\textit{Qwen2.5-Math-7B (3 runs; Baseline AVG 43.7)}}\\
Reported & 39.5 & 57.7 & 67.9 & 55.0\\
Repeat 1 & 35.3 & 58.1 & 68.5 & 54.0\\
Repeat 2 & 37.8 & 56.6 & 66.5 & 53.6\\
\cmidrule(lr){1-5}
Mean $\pm$ SD & $37.5\pm2.1$ & $57.5\pm0.8$ & $67.6\pm1.0$ & $54.2\pm0.7$\\
\midrule
\multicolumn{5}{@{}l}{\textit{DeepSeek-Math-7B-Instruct (6 runs; Baseline AVG 32.7)}}\\
Reported & 15.9 & 35.5 & 50.3 & 33.9\\
Repeat 1 & 15.7 & 34.9 & 48.8 & 33.1\\
Repeat 2 & 15.6 & 35.2 & 49.0 & 33.3\\
Repeat 3 & 15.7 & 35.2 & 48.9 & 33.3\\
Repeat 4 & 15.3 & 35.4 & 50.3 & 33.6\\
Repeat 5 & 15.8 & 35.5 & 50.2 & 33.8\\
\cmidrule(lr){1-5}
Mean $\pm$ SD & $15.7\pm0.2$ & $35.3\pm0.2$ & $49.6\pm0.8$ & $33.5\pm0.3$\\
\bottomrule
\end{tabular}
\end{table}

All nine runs exceed their corresponding Baseline in AVG, showing that the improvement persists across data orders. Qwen-Math's Pass@1 varies more than AVG. The reported standard deviations describe data-order variability, not all sources of training randomness or question-bootstrap uncertainty. An independent matched training-seed test of the generator control is reported in Appendix~\ref{app:source-details}.

\subsection{Detailed benchmark and training-set results}
\begin{table}[ht]
\centering\small
\caption{SePT and GRPO across two training-question datasets (AVG). Benchmark-level results appear in Tables~\ref{tab:appendix_table2_otm_dsr_breakdown} and~\ref{tab:appendix_table6_limitations_otm_breakdown}.}\label{tab:full-training-recipes}
\begin{tabular}{@{}lrrrr@{}}\toprule
 & \multicolumn{2}{c}{DSR} & \multicolumn{2}{c}{OTM}\\
\cmidrule(lr){2-3}\cmidrule(lr){4-5}
Model & SePT & GRPO & SePT & GRPO\\\midrule
Qwen2.5-Math-7B &55.0&59.1&55.2&56.6\\
DeepSeek-Math-7B-Instruct &33.9&35.6&33.0&33.4\\
Llama-3.1-8B-Instruct &37.8&41.2&38.2&36.8\\\bottomrule
\end{tabular}
\end{table}
In the tables below, ``Mean Pass@$k$'' denotes the six-benchmark mean at a single sampling budget, not the three-budget AVG reported in the main text. Scores are rounded independently, so averaging displayed values need not reproduce the last digit of an aggregate.
Base Model denotes the initial model evaluated at temperature one.
{\small\setlength{\tabcolsep}{3pt}
\begin{longtable}{@{}lrrrrrrrrr@{}}
\caption{Qwen models: benchmark-level Pass@$k$ for the Base Model, Baseline and SePT. Mean Pass@$k$ averages the six benchmarks at each sampling budget.}\label{tab:family-qwen}\label{tab:appendix_table1_workable_breakdown}\label{tab:additional-full}\\
\toprule & \multicolumn{3}{c}{Base Model} & \multicolumn{3}{c}{Baseline} & \multicolumn{3}{c}{SePT}\\
\cmidrule(lr){2-4}\cmidrule(lr){5-7}\cmidrule(lr){8-10}
Benchmark & $k=1$ & $k=8$ & $k=32$ & $k=1$ & $k=8$ & $k=32$ & $k=1$ & $k=8$ & $k=32$\\\midrule
\endfirsthead
\multicolumn{10}{l}{\emph{Qwen results (continued)}}\\
\toprule & \multicolumn{3}{c}{Base Model} & \multicolumn{3}{c}{Baseline} & \multicolumn{3}{c}{SePT}\\
\cmidrule(lr){2-4}\cmidrule(lr){5-7}\cmidrule(lr){8-10}
Benchmark & $k=1$ & $k=8$ & $k=32$ & $k=1$ & $k=8$ & $k=32$ & $k=1$ & $k=8$ & $k=32$\\\midrule
\endhead
\bottomrule\endfoot
\multicolumn{10}{l}{\textbf{Qwen2.5-7B}}\\*
MATH-500 & 26.6 & 75.9 & 91.8 & 44.2 & 83.5 & 92.6 & 70.5 & 89.1 & 93.6\\*
AMC-23 & 15.9 & 58.6 & 80.0 & 33.8 & 75.1 & 90.0 & 45.6 & 78.2 & 90.0\\*
Minerva & 5.8 & 27.6 & 50.0 & 10.2 & 34.8 & 50.4 & 25.5 & 50.3 & 60.3\\*
OlympiadBench & 12.1 & 41.1 & 57.5 & 25.7 & 53.6 & 66.7 & 34.9 & 57.7 & 68.2\\*
AIME-24 & 2.8 & 13.9 & 30.0 & 8.9 & 26.9 & 46.7 & 9.8 & 24.0 & 36.7\\*
AIME-25 & 1.4 & 8.0 & 23.3 & 5.0 & 18.5 & 26.7 & 7.3 & 28.4 & 43.3\\*
Mean Pass@$k$ & 10.7 & 37.5 & 55.4 & 21.3 & 48.7 & 62.2 & 32.3 & 54.6 & 65.3\\
\midrule
\multicolumn{10}{l}{\textbf{Qwen2.5-7B-Instruct}}\\*
MATH-500 & 73.6 & 90.1 & 94.2 & 74.7 & 90.4 & 94.8 & 74.9 & 90.6 & 95.6\\*
AMC-23 & 49.7 & 79.8 & 90.0 & 51.1 & 80.6 & 92.5 & 49.8 & 81.5 & 97.5\\*
Minerva & 36.2 & 54.4 & 62.9 & 37.1 & 53.6 & 59.6 & 36.8 & 54.2 & 61.4\\*
OlympiadBench & 37.2 & 60.8 & 71.2 & 38.0 & 60.4 & 70.4 & 38.0 & 59.8 & 69.5\\*
AIME-24 & 10.7 & 24.7 & 33.3 & 13.2 & 30.8 & 53.3 & 11.6 & 26.6 & 43.3\\*
AIME-25 & 6.8 & 23.4 & 30.0 & 7.0 & 23.7 & 33.3 & 8.2 & 27.9 & 46.7\\*
Mean Pass@$k$ & 35.7 & 55.5 & 63.6 & 36.8 & 56.6 & 67.3 & 36.6 & 56.8 & 69.0\\
\midrule
\multicolumn{10}{l}{\textbf{Qwen2.5-Math-7B}}\\*
MATH-500 & 30.4 & 78.2 & 91.2 & 50.2 & 79.2 & 89.4 & 77.2 & 90.1 & 93.0\\*
AMC-23 & 18.4 & 61.7 & 82.5 & 40.8 & 74.4 & 85.0 & 59.0 & 81.4 & 90.0\\*
Minerva & 6.9 & 31.0 & 50.4 & 10.6 & 32.6 & 47.1 & 35.5 & 53.6 & 61.8\\*
OlympiadBench & 9.3 & 36.6 & 56.6 & 16.5 & 40.9 & 57.7 & 39.6 & 60.6 & 69.1\\*
AIME-24 & 7.9 & 32.2 & 53.3 & 12.1 & 35.6 & 53.3 & 14.0 & 33.2 & 53.3\\*
AIME-25 & 2.1 & 12.6 & 26.7 & 5.9 & 21.4 & 33.3 & 11.7 & 27.5 & 40.0\\*
Mean Pass@$k$ & 12.5 & 42.0 & 60.1 & 22.7 & 47.3 & 61.0 & 39.5 & 57.7 & 67.9\\
\midrule
\multicolumn{10}{l}{\textbf{Qwen3-8B-Base}}\\*
MATH-500 & 53.7 & 89.1 & 94.0 & 72.6 & 91.4 & 95.4 & 78.5 & 91.5 & 94.8\\*
AMC-23 & 33.4 & 77.1 & 90.0 & 50.9 & 82.9 & 90.0 & 58.2 & 84.1 & 92.5\\*
Minerva & 21.5 & 53.2 & 65.1 & 31.9 & 57.2 & 65.1 & 38.1 & 58.8 & 65.1\\*
OlympiadBench & 24.2 & 56.0 & 68.4 & 37.8 & 61.8 & 72.1 & 42.3 & 62.6 & 71.3\\*
AIME-24 & 5.6 & 21.3 & 36.7 & 12.0 & 31.7 & 56.7 & 15.1 & 33.3 & 46.7\\*
AIME-25 & 3.2 & 16.2 & 30.0 & 10.0 & 27.9 & 40.0 & 12.9 & 24.6 & 33.3\\*
Mean Pass@$k$ & 23.6 & 52.2 & 64.0 & 35.9 & 58.8 & 69.9 & 40.9 & 59.2 & 67.3\\
\end{longtable}
}
{\small\setlength{\tabcolsep}{3pt}
\begin{longtable}{@{}lrrrrrrrrr@{}}
\caption{DeepSeek models: benchmark-level Pass@$k$ for the Base Model, Baseline and SePT. Mean Pass@$k$ averages the six benchmarks at each sampling budget.}\label{tab:family-deepseek}\\
\toprule & \multicolumn{3}{c}{Base Model} & \multicolumn{3}{c}{Baseline} & \multicolumn{3}{c}{SePT}\\
\cmidrule(lr){2-4}\cmidrule(lr){5-7}\cmidrule(lr){8-10}
Benchmark & $k=1$ & $k=8$ & $k=32$ & $k=1$ & $k=8$ & $k=32$ & $k=1$ & $k=8$ & $k=32$\\\midrule
\endfirsthead
\multicolumn{10}{l}{\emph{DeepSeek results (continued)}}\\
\toprule & \multicolumn{3}{c}{Base Model} & \multicolumn{3}{c}{Baseline} & \multicolumn{3}{c}{SePT}\\
\cmidrule(lr){2-4}\cmidrule(lr){5-7}\cmidrule(lr){8-10}
Benchmark & $k=1$ & $k=8$ & $k=32$ & $k=1$ & $k=8$ & $k=32$ & $k=1$ & $k=8$ & $k=32$\\\midrule
\endhead
\bottomrule\endfoot
\multicolumn{10}{l}{\textbf{DeepSeek-Math-7B}}\\*
MATH-500 & 4.5 & 26.0 & 52.8 & 6.6 & 34.3 & 62.8 & 7.0 & 35.0 & 62.2\\*
AMC-23 & 2.0 & 13.3 & 35.0 & 3.6 & 23.2 & 55.0 & 3.4 & 19.1 & 42.5\\*
Minerva & 1.7 & 10.9 & 25.4 & 3.7 & 19.5 & 37.5 & 3.0 & 16.5 & 36.0\\*
OlympiadBench & 1.3 & 8.8 & 23.8 & 2.2 & 13.1 & 29.0 & 2.1 & 12.5 & 27.6\\*
AIME-24 & 0.0 & 0.0 & 0.0 & 0.3 & 2.5 & 10.0 & 0.5 & 4.2 & 16.7\\*
AIME-25 & 0.2 & 1.5 & 3.3 & 0.0 & 0.0 & 0.0 & 0.5 & 4.2 & 16.7\\*
Mean Pass@$k$ & 1.6 & 10.1 & 23.4 & 2.7 & 15.4 & 32.4 & 2.8 & 15.2 & 33.6\\
\midrule
\multicolumn{10}{l}{\textbf{DeepSeek-Math-7B-Instruct}}\\*
MATH-500 & 38.1 & 69.7 & 83.2 & 42.7 & 71.5 & 84.2 & 42.4 & 71.2 & 83.6\\*
AMC-23 & 15.0 & 44.7 & 70.0 & 16.7 & 44.7 & 65.0 & 18.5 & 48.7 & 72.5\\*
Minerva & 17.6 & 44.5 & 59.6 & 18.6 & 44.0 & 59.2 & 19.7 & 44.4 & 56.2\\*
OlympiadBench & 9.8 & 31.4 & 49.0 & 12.4 & 35.0 & 49.9 & 12.7 & 34.0 & 49.5\\*
AIME-24 & 0.7 & 5.1 & 13.3 & 1.4 & 9.3 & 23.3 & 1.4 & 9.3 & 23.3\\*
AIME-25 & 0.3 & 2.3 & 6.7 & 0.5 & 3.5 & 6.7 & 0.7 & 5.5 & 16.7\\*
Mean Pass@$k$ & 13.6 & 33.0 & 47.0 & 15.4 & 34.6 & 48.1 & 15.9 & 35.5 & 50.3\\
\end{longtable}
}
{\small\setlength{\tabcolsep}{3pt}
\begin{longtable}{@{}lrrrrrrrrr@{}}
\caption{Llama models: benchmark-level Pass@$k$ for the Base Model, Baseline and SePT. Mean Pass@$k$ averages the six benchmarks at each sampling budget.}\label{tab:family-llama}\\
\toprule & \multicolumn{3}{c}{Base Model} & \multicolumn{3}{c}{Baseline} & \multicolumn{3}{c}{SePT}\\
\cmidrule(lr){2-4}\cmidrule(lr){5-7}\cmidrule(lr){8-10}
Benchmark & $k=1$ & $k=8$ & $k=32$ & $k=1$ & $k=8$ & $k=32$ & $k=1$ & $k=8$ & $k=32$\\\midrule
\endfirsthead
\multicolumn{10}{l}{\emph{Llama results (continued)}}\\
\toprule & \multicolumn{3}{c}{Base Model} & \multicolumn{3}{c}{Baseline} & \multicolumn{3}{c}{SePT}\\
\cmidrule(lr){2-4}\cmidrule(lr){5-7}\cmidrule(lr){8-10}
Benchmark & $k=1$ & $k=8$ & $k=32$ & $k=1$ & $k=8$ & $k=32$ & $k=1$ & $k=8$ & $k=32$\\\midrule
\endhead
\bottomrule\endfoot
\multicolumn{10}{l}{\textbf{Llama-3.1-8B}}\\*
MATH-500 & 4.8 & 25.7 & 49.6 & 13.6 & 43.4 & 63.8 & 15.9 & 43.4 & 62.4\\*
AMC-23 & 3.3 & 22.2 & 60.0 & 3.5 & 22.5 & 50.0 & 4.7 & 28.1 & 65.0\\*
Minerva & 1.7 & 10.4 & 22.1 & 4.5 & 19.6 & 34.9 & 6.3 & 23.0 & 39.7\\*
OlympiadBench & 1.3 & 8.4 & 21.4 & 2.6 & 12.7 & 27.0 & 3.5 & 15.7 & 28.8\\*
AIME-24 & 0.0 & 0.0 & 0.0 & 0.3 & 2.5 & 10.0 & 0.4 & 2.8 & 6.7\\*
AIME-25 & 0.0 & 0.0 & 0.0 & 0.3 & 2.5 & 10.0 & 0.1 & 0.8 & 3.3\\*
Mean Pass@$k$ & 1.9 & 11.1 & 25.5 & 4.1 & 17.2 & 32.6 & 5.1 & 19.0 & 34.3\\
\midrule
\multicolumn{10}{l}{\textbf{Llama-3.1-8B-Instruct}}\\*
MATH-500 & 38.9 & 70.1 & 82.2 & 48.9 & 76.2 & 86.2 & 49.2 & 75.8 & 85.8\\*
AMC-23 & 15.4 & 50.6 & 75.0 & 24.1 & 56.6 & 80.0 & 22.9 & 56.6 & 82.5\\*
Minerva & 19.0 & 43.2 & 54.4 & 26.6 & 47.5 & 59.2 & 25.6 & 45.4 & 54.4\\*
OlympiadBench & 10.0 & 30.2 & 46.7 & 15.6 & 38.4 & 53.3 & 16.2 & 38.9 & 54.4\\*
AIME-24 & 4.1 & 22.1 & 43.3 & 7.5 & 24.4 & 43.3 & 5.7 & 18.2 & 30.0\\*
AIME-25 & 0.1 & 0.8 & 3.3 & 0.3 & 2.5 & 10.0 & 0.5 & 4.0 & 13.3\\*
Mean Pass@$k$ & 14.6 & 36.2 & 50.8 & 20.5 & 40.9 & 55.3 & 20.0 & 39.8 & 53.4\\
\end{longtable}
}
{\small\setlength{\tabcolsep}{3pt}
\begin{longtable}{@{}lrrrrrrrrr@{}}
\caption{Phi models: benchmark-level Pass@$k$ for the Base Model, Baseline and SePT. Mean Pass@$k$ averages the six benchmarks at each sampling budget.}\label{tab:family-phi}\\
\toprule & \multicolumn{3}{c}{Base Model} & \multicolumn{3}{c}{Baseline} & \multicolumn{3}{c}{SePT}\\
\cmidrule(lr){2-4}\cmidrule(lr){5-7}\cmidrule(lr){8-10}
Benchmark & $k=1$ & $k=8$ & $k=32$ & $k=1$ & $k=8$ & $k=32$ & $k=1$ & $k=8$ & $k=32$\\\midrule
\endfirsthead
\multicolumn{10}{l}{\emph{Phi results (continued)}}\\
\toprule & \multicolumn{3}{c}{Base Model} & \multicolumn{3}{c}{Baseline} & \multicolumn{3}{c}{SePT}\\
\cmidrule(lr){2-4}\cmidrule(lr){5-7}\cmidrule(lr){8-10}
Benchmark & $k=1$ & $k=8$ & $k=32$ & $k=1$ & $k=8$ & $k=32$ & $k=1$ & $k=8$ & $k=32$\\\midrule
\endhead
\bottomrule\endfoot
\multicolumn{10}{l}{\textbf{Phi-4-mini-Instruct}}\\*
MATH-500 & 56.9 & 83.5 & 90.6 & 68.8 & 84.6 & 89.6 & 67.7 & 86.7 & 92.6\\*
AMC-23 & 35.1 & 64.7 & 75.0 & 39.0 & 70.1 & 90.0 & 40.1 & 70.7 & 82.5\\*
Minerva & 32.0 & 52.4 & 61.0 & 35.1 & 48.0 & 54.0 & 33.3 & 52.0 & 58.8\\*
OlympiadBench & 26.4 & 49.0 & 60.5 & 34.8 & 52.8 & 60.6 & 30.7 & 52.5 & 62.7\\*
AIME-24 & 5.6 & 18.3 & 33.3 & 8.5 & 19.9 & 26.7 & 7.5 & 20.9 & 33.3\\*
AIME-25 & 3.1 & 13.2 & 20.0 & 6.5 & 17.4 & 26.7 & 4.2 & 17.4 & 33.3\\*
Mean Pass@$k$ & 26.5 & 46.8 & 56.7 & 32.1 & 48.8 & 57.9 & 30.6 & 50.0 & 60.5\\
\end{longtable}
}

{\small\setlength{\tabcolsep}{3pt}
\begin{longtable}{@{}p{.19\linewidth}p{.17\linewidth}p{.13\linewidth}p{.07\linewidth}rrr@{}}
\caption{SePT and GRPO on DSR and OTM.}\label{tab:appendix_table2_otm_dsr_breakdown}\\
\toprule
\textbf{Model} & \textbf{Benchmark} & \textbf{Method} & \textbf{Data} & \textbf{$k=1$} & \textbf{$k=8$} & \textbf{$k=32$} \\
\midrule
\endfirsthead
\toprule
\textbf{Model} & \textbf{Benchmark} & \textbf{Method} & \textbf{Data} & \textbf{$k=1$} & \textbf{$k=8$} & \textbf{$k=32$} \\
\midrule
\endhead
\bottomrule\endfoot

Qwen2.5-Math-7B & MATH-500 & Baseline & \textemdash{} & 50.2 & 79.2 & 89.4 \\
\cmidrule(lr){3-7}
 &  & GRPO & OTM & 76.3 & 90.4 & 94.2 \\
 &  &  & DSR & 80.9 & 90.4 & 93.8 \\
\cmidrule(lr){3-7}
 &  & SePT & OTM & 78.1 & 90.7 & 93.8 \\
 &  &  & DSR & 77.2 & 90.1 & 93.0 \\
\midrule
 & AMC-23 & Baseline & \textemdash{} & 40.8 & 74.4 & 85.0 \\
\cmidrule(lr){3-7}
 &  & GRPO & OTM & 58.8 & 79.8 & 90.0 \\
 &  &  & DSR & 63.8 & 81.6 & 92.5 \\
\cmidrule(lr){3-7}
 &  & SePT & OTM & 62.0 & 83.9 & 92.5 \\
 &  &  & DSR & 59.0 & 81.4 & 90.0 \\
\midrule
 & Minerva & Baseline & \textemdash{} & 10.6 & 32.6 & 47.1 \\
\cmidrule(lr){3-7}
 &  & GRPO & OTM & 34.0 & 51.7 & 61.0 \\
 &  &  & DSR & 37.3 & 50.9 & 57.7 \\
\cmidrule(lr){3-7}
 &  & SePT & OTM & 35.8 & 52.9 & 58.8 \\
 &  &  & DSR & 35.5 & 53.6 & 61.8 \\
\midrule
 & Olympiad & Baseline & \textemdash{} & 16.5 & 40.9 & 57.7 \\
\cmidrule(lr){3-7}
 &  & GRPO & OTM & 40.1 & 60.2 & 69.7 \\
 &  &  & DSR & 41.6 & 60.6 & 68.8 \\
\cmidrule(lr){3-7}
 &  & SePT & OTM & 40.6 & 60.3 & 68.5 \\
 &  &  & DSR & 39.6 & 60.6 & 69.1 \\
\midrule
 & AIME-24 & Baseline & \textemdash{} & 12.1 & 35.6 & 53.3 \\
\cmidrule(lr){3-7}
 &  & GRPO & OTM & 15.3 & 42.8 & 66.7 \\
 &  &  & DSR & 24.6 & 53.4 & 66.7 \\
\cmidrule(lr){3-7}
 &  & SePT & OTM & 14.1 & 33.8 & 46.7 \\
 &  &  & DSR & 14.0 & 33.2 & 53.3 \\
\midrule
 & AIME-25 & Baseline & \textemdash{} & 5.9 & 21.4 & 33.3 \\
\cmidrule(lr){3-7}
 &  & GRPO & OTM & 12.8 & 29.3 & 46.7 \\
 &  &  & DSR & 14.6 & 34.2 & 50.0 \\
\cmidrule(lr){3-7}
 &  & SePT & OTM & 14.5 & 30.1 & 36.7 \\
 &  &  & DSR & 11.7 & 27.5 & 40.0 \\
\midrule
 & Mean Pass@$k$ & Baseline & \textemdash{} & 22.7 & 47.3 & 61.0 \\
\cmidrule(lr){3-7}
 &  & GRPO & OTM & 39.5 & 59.0 & 71.4 \\
 &  &  & DSR & 43.8 & 61.8 & 71.6 \\
\cmidrule(lr){3-7}
 &  & SePT & OTM & 40.8 & 58.6 & 66.2 \\
 &  &  & DSR & 39.5 & 57.7 & 67.9 \\
\midrule
\parbox[c]{2.35cm}{\raggedright DeepSeek-Math\\-7B-Instruct} & MATH-500 & Baseline & \textemdash{} & 42.7 & 71.5 & 84.2 \\
\cmidrule(lr){3-7}
 &  & GRPO & OTM & 44.2 & 71.5 & 83.0 \\
 &  &  & DSR & 47.6 & 73.6 & 84.4 \\
\cmidrule(lr){3-7}
 &  & SePT & OTM & 42.6 & 71.3 & 84.2 \\
 &  &  & DSR & 42.4 & 71.2 & 83.6 \\
\midrule
 & AMC-23 & Baseline & \textemdash{} & 16.7 & 44.7 & 65.0 \\
\cmidrule(lr){3-7}
 &  & GRPO & OTM & 19.9 & 49.0 & 75.0 \\
 &  &  & DSR & 21.2 & 55.0 & 77.5 \\
\cmidrule(lr){3-7}
 &  & SePT & OTM & 17.9 & 47.9 & 65.0 \\
 &  &  & DSR & 18.5 & 48.7 & 72.5 \\
\midrule
 & Minerva & Baseline & \textemdash{} & 18.6 & 44.0 & 59.2 \\
\cmidrule(lr){3-7}
 &  & GRPO & OTM & 21.7 & 46.1 & 60.3 \\
 &  &  & DSR & 22.3 & 46.6 & 59.2 \\
\cmidrule(lr){3-7}
 &  & SePT & OTM & 18.9 & 42.9 & 55.1 \\
 &  &  & DSR & 19.7 & 44.4 & 56.2 \\
\midrule
 & Olympiad & Baseline & \textemdash{} & 12.4 & 35.0 & 49.9 \\
\cmidrule(lr){3-7}
 &  & GRPO & OTM & 13.9 & 34.3 & 47.7 \\
 &  &  & DSR & 15.6 & 37.4 & 51.6 \\
\cmidrule(lr){3-7}
 &  & SePT & OTM & 12.9 & 34.0 & 48.4 \\
 &  &  & DSR & 12.7 & 34.0 & 49.5 \\
\midrule
 & AIME-24 & Baseline & \textemdash{} & 1.4 & 9.3 & 23.3 \\
\cmidrule(lr){3-7}
 &  & GRPO & OTM & 1.7 & 8.3 & 16.7 \\
 &  &  & DSR & 2.1 & 10.1 & 20.0 \\
\cmidrule(lr){3-7}
 &  & SePT & OTM & 1.7 & 9.5 & 16.7 \\
 &  &  & DSR & 1.4 & 9.3 & 23.3 \\
\midrule
 & AIME-25 & Baseline & \textemdash{} & 0.5 & 3.5 & 6.7 \\
\cmidrule(lr){3-7}
 &  & GRPO & OTM & 0.2 & 1.7 & 6.7 \\
 &  &  & DSR & 0.4 & 3.3 & 13.3 \\
\cmidrule(lr){3-7}
 &  & SePT & OTM & 0.6 & 5.0 & 20.0 \\
 &  &  & DSR & 0.7 & 5.5 & 16.7 \\
\midrule
 & Mean Pass@$k$ & Baseline & \textemdash{} & 15.4 & 34.6 & 48.1 \\
\cmidrule(lr){3-7}
 &  & GRPO & OTM & 16.9 & 35.1 & 48.2 \\
 &  &  & DSR & 18.2 & 37.7 & 51.0 \\
\cmidrule(lr){3-7}
 &  & SePT & OTM & 15.8 & 35.1 & 48.2 \\
 &  &  & DSR & 15.9 & 35.5 & 50.3 \\
\end{longtable}
}

{\small\setlength{\tabcolsep}{3pt}
\begin{longtable}{@{}p{.25\linewidth}p{.13\linewidth}p{.18\linewidth}rrr@{}}
\caption{SePT and EM-FT benchmark results (Figure~\ref{fig:em}).}\label{tab:appendix_table3_em_dsr_breakdown}\\
\toprule
\textbf{Model} & \textbf{Method} & \textbf{Benchmark} & \textbf{$k=1$} & \textbf{$k=8$} & \textbf{$k=32$} \\
\midrule
\endfirsthead
\toprule
\textbf{Model} & \textbf{Method} & \textbf{Benchmark} & \textbf{$k=1$} & \textbf{$k=8$} & \textbf{$k=32$} \\
\midrule
\endhead
\bottomrule\endfoot

Qwen2.5-Math-7B & Baseline & MATH-500 & 50.2 & 79.2 & 89.4 \\
 &  & AMC-23 & 40.8 & 74.4 & 85.0 \\
 &  & Minerva & 10.6 & 32.6 & 47.1 \\
 &  & Olympiad & 16.5 & 40.9 & 57.7 \\
 &  & AIME-24 & 12.1 & 35.6 & 53.3 \\
 &  & AIME-25 & 5.9 & 21.4 & 33.3 \\
 &  & Mean Pass@$k$ & 22.7 & 47.3 & 61.0 \\
\cmidrule(lr){2-6}
 & SePT & MATH-500 & 77.2 & 90.1 & 93.0 \\
 &  & AMC-23 & 59.0 & 81.4 & 90.0 \\
 &  & Minerva & 35.5 & 53.6 & 61.8 \\
 &  & Olympiad & 39.6 & 60.6 & 69.1 \\
 &  & AIME-24 & 14.0 & 33.2 & 53.3 \\
 &  & AIME-25 & 11.7 & 27.5 & 40.0 \\
 &  & Mean Pass@$k$ & 39.5 & 57.7 & 67.9 \\
\cmidrule(lr){2-6}
 & EM-FT & MATH-500 & 61.1 & 89.4 & 94.4 \\
 &  & AMC-23 & 47.6 & 80.2 & 87.5 \\
 &  & Minerva & 18.7 & 47.9 & 60.7 \\
 &  & Olympiad & 29.4 & 57.8 & 71.0 \\
 &  & AIME-24 & 15.6 & 44.2 & 53.3 \\
 &  & AIME-25 & 6.8 & 21.5 & 36.7 \\
 &  & Mean Pass@$k$ & 29.9 & 56.8 & 67.3 \\
\midrule
\parbox[c]{2.35cm}{\raggedright DeepSeek-Math\\-7B-Instruct} & Baseline & MATH-500 & 42.7 & 71.5 & 84.2 \\
 &  & AMC-23 & 16.7 & 44.7 & 65.0 \\
 &  & Minerva & 18.6 & 44.0 & 59.2 \\
 &  & Olympiad & 12.4 & 35.0 & 49.9 \\
 &  & AIME-24 & 1.4 & 9.3 & 23.3 \\
 &  & AIME-25 & 0.5 & 3.5 & 6.7 \\
 &  & Mean Pass@$k$ & 15.4 & 34.6 & 48.1 \\
\cmidrule(lr){2-6}
 & SePT & MATH-500 & 42.4 & 71.2 & 83.6 \\
 &  & AMC-23 & 18.5 & 48.7 & 72.5 \\
 &  & Minerva & 19.7 & 44.4 & 56.2 \\
 &  & Olympiad & 12.7 & 34.0 & 49.5 \\
 &  & AIME-24 & 1.4 & 9.3 & 23.3 \\
 &  & AIME-25 & 0.7 & 5.5 & 16.7 \\
 &  & Mean Pass@$k$ & 15.9 & 35.5 & 50.3 \\
\cmidrule(lr){2-6}
 & EM-FT & MATH-500 & 43.1 & 70.7 & 83.6 \\
 &  & AMC-23 & 18.5 & 51.1 & 72.5 \\
 &  & Minerva & 19.4 & 43.7 & 55.1 \\
 &  & Olympiad & 12.9 & 35.1 & 50.1 \\
 &  & AIME-24 & 1.0 & 6.8 & 16.7 \\
 &  & AIME-25 & 0.1 & 0.8 & 3.3 \\
 &  & Mean Pass@$k$ & 15.9 & 34.7 & 46.9 \\
\end{longtable}
}

{\small\setlength{\tabcolsep}{3pt}
\begin{longtable}{@{}p{.27\linewidth}p{.28\linewidth}rrr@{}}
\caption{SePT compared with likelihood training on a fixed self-generated dataset.}\label{tab:appendix_table4_offline_sft_breakdown}\\
\toprule
\textbf{Benchmark} & \textbf{Method} & \textbf{$k=1$} & \textbf{$k=8$} & \textbf{$k=32$} \\
\midrule
\endfirsthead
\toprule
\textbf{Benchmark} & \textbf{Method} & \textbf{$k=1$} & \textbf{$k=8$} & \textbf{$k=32$} \\
\midrule
\endhead
\bottomrule\endfoot

MATH-500 & Baseline & 50.2 & 79.2 & 89.4 \\
 & SePT & 77.2 & 90.1 & 93.0 \\
 & SePT (Offline) & 47.6 & 85.7 & 94.2 \\
\midrule
AMC-23 & Baseline & 40.8 & 74.4 & 85.0 \\
 & SePT & 59.0 & 81.4 & 90.0 \\
 & SePT (Offline) & 36.4 & 75.4 & 90.0 \\
\midrule
Minerva & Baseline & 10.6 & 32.6 & 47.1 \\
 & SePT & 35.5 & 53.6 & 61.8 \\
 & SePT (Offline) & 10.9 & 38.7 & 56.6 \\
\midrule
Olympiad & Baseline & 16.5 & 40.9 & 57.7 \\
 & SePT & 39.6 & 60.6 & 69.1 \\
 & SePT (Offline) & 15.7 & 45.3 & 63.4 \\
\midrule
AIME-24 & Baseline & 12.1 & 35.6 & 53.3 \\
 & SePT & 14.0 & 33.2 & 53.3 \\
 & SePT (Offline) & 11.6 & 38.1 & 53.3 \\
\midrule
AIME-25 & Baseline & 5.9 & 21.4 & 33.3 \\
 & SePT & 11.7 & 27.5 & 40.0 \\
 & SePT (Offline) & 3.4 & 18.3 & 33.3 \\
\midrule
Mean Pass@$k$ & Baseline & 22.7 & 47.3 & 61.0 \\
 & SePT & 39.5 & 57.7 & 67.9 \\
 & SePT (Offline) & 21.0 & 50.2 & 65.2 \\
\end{longtable}
}

{\small\setlength{\tabcolsep}{3pt}
\begin{longtable}{@{}p{.27\linewidth}p{.28\linewidth}rrr@{}}
\caption{Temperature-dynamics ablation.}\label{tab:appendix_table5_decoupling_breakdown}\\
\toprule
\textbf{Benchmark} & \textbf{Method} & \textbf{$k=1$} & \textbf{$k=8$} & \textbf{$k=32$} \\
\midrule
\endfirsthead
\toprule
\textbf{Benchmark} & \textbf{Method} & \textbf{$k=1$} & \textbf{$k=8$} & \textbf{$k=32$} \\
\midrule
\endhead
\bottomrule\endfoot

MATH-500 & Baseline & 50.2 & 79.2 & 89.4 \\
 & $\tau_s = \tau_t$ & 45.4 & 86.5 & 94.0 \\
 & Cold-to-unit schedule & 77.2 & 90.1 & 93.0 \\
\midrule
AMC-23 & Baseline & 40.8 & 74.4 & 85.0 \\
 & $\tau_s = \tau_t$ & 31.1 & 76.5 & 90.0 \\
 & Cold-to-unit schedule & 59.0 & 81.4 & 90.0 \\
\midrule
Minerva & Baseline & 10.6 & 32.6 & 47.1 \\
 & $\tau_s = \tau_t$ & 9.8 & 37.6 & 56.6 \\
 & Cold-to-unit schedule & 35.5 & 53.6 & 61.8 \\
\midrule
Olympiad & Baseline & 16.5 & 40.9 & 57.7 \\
 & $\tau_s = \tau_t$ & 15.1 & 47.1 & 65.1 \\
 & Cold-to-unit schedule & 39.6 & 60.6 & 69.1 \\
\midrule
AIME-24 & Baseline & 12.1 & 35.6 & 53.3 \\
 & $\tau_s = \tau_t$ & 11.1 & 38.3 & 56.7 \\
 & Cold-to-unit schedule & 14.0 & 33.2 & 53.3 \\
\midrule
AIME-25 & Baseline & 5.9 & 21.4 & 33.3 \\
 & $\tau_s = \tau_t$ & 3.3 & 14.6 & 23.3 \\
 & Cold-to-unit schedule & 11.7 & 27.5 & 40.0 \\
\midrule
Mean Pass@$k$ & Baseline & 22.7 & 47.3 & 61.0 \\
 & $\tau_s = \tau_t$ & 19.3 & 50.1 & 64.3 \\
 & Cold-to-unit schedule & 39.5 & 57.7 & 67.9 \\
\end{longtable}
}

{\small\setlength{\tabcolsep}{3pt}
\begin{longtable}{@{}p{.19\linewidth}p{.17\linewidth}p{.13\linewidth}p{.07\linewidth}rrr@{}}
\caption{Llama-Instruct results on DSR and OTM.}\label{tab:appendix_table6_limitations_otm_breakdown}\\
\toprule
\textbf{Model} & \textbf{Benchmark} & \textbf{Method} & \textbf{Data} & \textbf{$k=1$} & \textbf{$k=8$} & \textbf{$k=32$} \\
\midrule
\endfirsthead
\toprule
\textbf{Model} & \textbf{Benchmark} & \textbf{Method} & \textbf{Data} & \textbf{$k=1$} & \textbf{$k=8$} & \textbf{$k=32$} \\
\midrule
\endhead
\bottomrule\endfoot

\parbox[c]{2.0cm}{\raggedright Llama-3.1\\-8B-Instruct} & MATH-500 & Base Model & \textemdash{} & 38.9 & 70.1 & 82.2 \\
 &  & Baseline & \textemdash{} & 48.9 & 76.2 & 86.2 \\
\cmidrule(lr){3-7}
 &  & SePT & DSR & 49.2 & 75.8 & 85.8 \\
 &  &  & OTM & 47.9 & 74.0 & 83.4 \\
\cmidrule(lr){3-7}
 &  & GRPO & DSR & 55.6 & 78.1 & 87.0 \\
 &  &  & OTM & 47.7 & 76.2 & 86.0 \\
\midrule
 & AMC-23 & Base Model & \textemdash{} & 15.4 & 50.6 & 75.0 \\
 &  & Baseline & \textemdash{} & 24.1 & 56.6 & 80.0 \\
\cmidrule(lr){3-7}
 &  & SePT & DSR & 22.9 & 56.6 & 82.5 \\
 &  &  & OTM & 25.7 & 60.0 & 82.5 \\
\cmidrule(lr){3-7}
 &  & GRPO & DSR & 33.4 & 61.7 & 75.0 \\
 &  &  & OTM & 24.2 & 58.4 & 77.5 \\
\midrule
 & Minerva & Base Model & \textemdash{} & 19.0 & 43.2 & 54.4 \\
 &  & Baseline & \textemdash{} & 26.6 & 47.5 & 59.2 \\
\cmidrule(lr){3-7}
 &  & SePT & DSR & 25.6 & 45.4 & 54.4 \\
 &  &  & OTM & 24.3 & 44.6 & 55.5 \\
\cmidrule(lr){3-7}
 &  & GRPO & DSR & 29.4 & 50.0 & 59.6 \\
 &  &  & OTM & 25.4 & 48.5 & 59.6 \\
\midrule
 & Olympiad & Base Model & \textemdash{} & 10.0 & 30.2 & 46.7 \\
 &  & Baseline & \textemdash{} & 15.6 & 38.4 & 53.3 \\
\cmidrule(lr){3-7}
 &  & SePT & DSR & 16.2 & 38.9 & 54.4 \\
 &  &  & OTM & 15.7 & 36.5 & 49.8 \\
\cmidrule(lr){3-7}
 &  & GRPO & DSR & 21.4 & 40.1 & 51.4 \\
 &  &  & OTM & 15.6 & 35.9 & 49.0 \\
\midrule
 & AIME-24 & Base Model & \textemdash{} & 4.1 & 22.1 & 43.3 \\
 &  & Baseline & \textemdash{} & 7.5 & 24.4 & 43.3 \\
\cmidrule(lr){3-7}
 &  & SePT & DSR & 5.7 & 18.2 & 30.0 \\
 &  &  & OTM & 6.0 & 21.5 & 36.7 \\
\cmidrule(lr){3-7}
 &  & GRPO & DSR & 10.9 & 29.9 & 43.3 \\
 &  &  & OTM & 5.2 & 19.4 & 30.0 \\
\midrule
 & AIME-25 & Base Model & \textemdash{} & 0.1 & 0.8 & 3.3 \\
 &  & Baseline & \textemdash{} & 0.3 & 2.5 & 10.0 \\
\cmidrule(lr){3-7}
 &  & SePT & DSR & 0.5 & 4.0 & 13.3 \\
 &  &  & OTM & 0.8 & 5.7 & 16.7 \\
\cmidrule(lr){3-7}
 &  & GRPO & DSR & 0.7 & 4.3 & 10.0 \\
 &  &  & OTM & 0.1 & 0.8 & 3.3 \\
\midrule
 & Mean Pass@$k$ & Base Model & \textemdash{} & 14.6 & 36.2 & 50.8 \\
 &  & Baseline & \textemdash{} & 20.5 & 40.9 & 55.3 \\
\cmidrule(lr){3-7}
 &  & SePT & DSR & 20.0 & 39.8 & 53.4 \\
 &  &  & OTM & 20.1 & 40.4 & 54.1 \\
\cmidrule(lr){3-7}
 &  & GRPO & DSR & 25.2 & 44.0 & 54.4 \\
 &  &  & OTM & 19.7 & 39.9 & 50.9 \\
\end{longtable}
}

\section{Training--evaluation overlap and sensitivity}\label{app:overlap}
Our audit extracts training and evaluation question text and takes the union of five candidate checks: exact normalized text; digit-insensitive exact matching and contiguous 32-word overlap following Light-R1 \citep{wen2025light}; normalized Indel similarity at least 75 and contiguous 13-token overlap under the Qwen2 tokenizer following OpenThoughts \citep{guha2025openthoughts}.
Candidate pairs were reviewed: exact duplicates and clear long-span overlaps were retained; digit-insensitive, token and Indel matches required problem-specific overlap.
Answer-format boilerplate, generic definitions, numeric runs, indexed-variable notation and diagram code alone were rejected.
Reviewed-positive MATH matches include near-duplicate or number-changed templates.

\par
\begin{table}[ht]
\centering\small
\caption{Reviewed training--evaluation matches. Counts are unique within each row; totals de-duplicate training rows across benchmarks. The audit covers all 675 OlympiadBench questions; the matched evaluation panels in Appendix~\ref{app:recovery} contain 673, explaining the totals of 1,547 versus 1,545.}\label{tab:restored-overlap}
\begin{tabular}{llrrrr}\toprule
Set & Benchmark & Train rows & Train rate & Eval. questions & Eval. rate\\\midrule
DSR & MATH-500 &145&0.36\%&67/500&13.40\%\\
 & AMC-23 &0&0.00\%&0/40&0.00\%\\
 & AIME-24 &0&0.00\%&0/30&0.00\%\\
 & AIME-25 &0&0.00\%&0/30&0.00\%\\
 & Minerva &0&0.00\%&0/272&0.00\%\\
 & OlympiadBench &18&0.04\%&17/675&2.52\%\\
 & Total &163&0.40\%&84/1547&5.43\%\\\midrule
OTM & MATH-500 &0&0.00\%&0/500&0.00\%\\
 & AMC-23 &0&0.00\%&0/40&0.00\%\\
 & AIME-24 &0&0.00\%&0/30&0.00\%\\
 & AIME-25 &0&0.00\%&0/30&0.00\%\\
 & Minerva &0&0.00\%&0/272&0.00\%\\
 & OlympiadBench &11&0.03\%&3/675&0.44\%\\
 & Total &11&0.03\%&3/1547&0.19\%\\\bottomrule
\end{tabular}
\end{table}
\par

\paragraph{Mapping the audit to the source-control questions.}
The reviewed labels identify 67 flagged and 433 unflagged MATH-500 questions in the generator controls. All 145 corresponding DSR training rows survive the Qwen prompt-length filter (40,315 raw rows, 40,306 retained). We reuse these labels without further review. ``Unflagged'' means no match under this audit, not verified absence of prior exposure.

\paragraph{Fixed-checkpoint subgroup results.}
We retain all generations, scores and selection decisions and resample paired question blocks within each group (10,000 draws), sharing draws across arms and metrics.
\par
\begin{table}[ht]\centering\small
\caption{Source-control accuracy on the full, audit-flagged and unflagged MATH-500 populations. Each model contributes eight responses per question; scores are percentages.}\label{tab:overlap-subgroups}
\begin{tabular}{llrrr}\toprule
Panel & Metric & SePT & \shortstack{Base-generator\\training} & \shortstack{SePT-generator\\transfer}\\\midrule
Full (500) & Pass@1 & 65.025 & 40.650 & 64.900\\
Full (500) & Pass@8 & 87.600 & 85.200 & 86.400\\
Flagged (67) & Pass@1 & 66.604 & 47.015 & 68.284\\
Flagged (67) & Pass@8 & 88.060 & 86.567 & 86.567\\
Unflagged (433) & Pass@1 & 64.781 & 39.665 & 64.376\\
Unflagged (433) & Pass@8 & 87.529 & 84.988 & 86.374\\
\bottomrule\end{tabular}
\end{table}
\par
\begin{table}[ht]\centering\small
\caption{Paired source-control contrasts within the overlap groups in Table~\ref{tab:overlap-subgroups}. Differences and paired-question 95\% intervals are Pass@1 points.}\label{tab:overlap-contrasts}
\begin{tabularx}{\linewidth}{l>{\raggedright\arraybackslash}Xrr}\toprule
Panel & Contrast & Pass@1 gap & 95\% interval\\\midrule
Full & SePT--Base-generator training & 24.375 & [21.650,27.125]\\
Full & SePT-generator transfer--Base-generator training & 24.250 & [21.400,27.050]\\
Full & SePT-generator transfer--SePT & -0.125 & [-1.850,1.600]\\
Flagged & SePT--Base-generator training & 19.590 & [13.060,26.119]\\
Flagged & SePT-generator transfer--Base-generator training & 21.269 & [14.552,27.799]\\
Flagged & SePT-generator transfer--SePT & 1.679 & [-3.172,6.530]\\
Unflagged & SePT--Base-generator training & 25.115 & [22.084,28.204]\\
Unflagged & SePT-generator transfer--Base-generator training & 24.711 & [21.622,27.858]\\
Unflagged & SePT-generator transfer--SePT & -0.404 & [-2.252,1.415]\\
\bottomrule\end{tabularx}
\end{table}
\par

The transfer advantage persists on unflagged questions. These exploratory intervals hold the checkpoints and audit labels fixed, following Appendix~\ref{app:statistics}.

\paragraph{Deterministic deletion sensitivity.}
Keep all three checkpoints, all eight responses per question and their persisted scores fixed.
For each contrast let $d_q\in[-1,1]$ be the paired difference in mean correctness, $N=500$, and remove any $M=67$ questions $C$.
Then
\[
\bar d_{\bar C}=\frac{N\bar d-\sum_{q\in C}d_q}{N-M}
\ \geq\ \frac{N\bar d-M}{N-M}.
\]
Sorting the observed $d_q$ and deleting the 67 largest gives the exact minimum over all such removed sets, separately for each contrast.
\par
\begin{table}[ht]\centering\small
\caption{Worst-case deletion sensitivity for MATH-500 Pass@1. Each contrast removes any 67 of 500 questions while keeping responses and checkpoints fixed. Bounds are percentage points, not confidence intervals.}\label{tab:overlap-deletion}
\begin{tabular}{lrrr}\toprule
Contrast & Full gap (pp) & Range-only bound & Exact minimum\\\midrule
SePT--Base-generator training &24.375&12.673&16.166\\
SePT-generator transfer--Base-generator training &24.250&12.529&16.022\\\bottomrule
\end{tabular}
\end{table}
\par
No set of 67 questions accounts for the entire observed gap. This is a sensitivity check on fixed evaluations, not a retraining experiment: each contrast removes its own worst-case set, which may differ from the audit-flagged set.

\paragraph{Training on the lower-overlap OTM dataset.}
The Qwen2.5-Math-7B comparison is summarized in Table~\ref{tab:overlap-training-sets}.
\par
\begin{table}[ht]\centering\small
\caption{Mathematical performance of Qwen2.5-Math-7B trained on DSR or the lower-overlap OTM questions. The Baseline is selected by decoding temperature; all scores are percentages.}\label{tab:overlap-training-sets}
\begin{tabular}{lrrrr}\toprule
Setting & Pass@1 & Pass@8 & Pass@32 & AVG\\\midrule
Baseline &22.7&47.3&61.0&43.7\\
SePT on DSR &39.5&57.7&67.9&55.0\\
SePT on OTM &40.8&58.6&66.2&55.2\\\bottomrule
\end{tabular}
\end{table}
\par
SePT also improves when trained on OTM, which has fewer reviewed-positive matches. This tests a different training-question dataset, complementing the fixed-checkpoint subgroup analysis above.

\subsection{Training after removing all candidate overlaps}
A separate SePT run used a broader, conservative filter: it removed every candidate overlap before manual false-positive review, including all subsequently reviewed-positive training questions. Potentially harmless candidates were therefore excluded as well.
Recorded $n=32$, temperature-one evaluations yield the following six-suite means (extra AMC and response-length metrics excluded):
\par
\begin{table}[ht]\centering\small
\caption{Candidate-removed DSR training: approximate mathematical scores reconstructed from rounded evaluation logs. The initial model uses temperature one, not the temperature-selected Baseline.}\label{tab:candidate-removed}
\begin{tabular}{lrrrr}\toprule
Recorded state & Pass@1 & Pass@8 & Pass@32 & AVG\\\midrule
Initial, temperature one &12.63&42.22&59.48&38.11\\
Terminal model &37.60&59.22&68.88&55.23\\\bottomrule
\end{tabular}\end{table}
\par
The table reports initial and terminal states, without checkpoint selection. Scores are reconstructed from rounded logs, so paired confidence intervals are unavailable. The initial model uses temperature one, and only part of the sampling temperature schedule is recorded. This run supports improvement after broad candidate removal, but is not a matched replication of the main recipe or a comparison against the temperature-selected Baseline.

\section{Training with fixed and evolving generators}\label{app:source-details}
The generator controls use the protocol in Appendix~\ref{app:study-settings}. Figure~\ref{fig:controls} assesses 300, 650 and 900 total refreshes, shown as percentages of training duration. Continued-training controls start from the checkpoint at 300 refreshes.

\paragraph{Source evolution improves on fresh initial-generator samples.}
Base-generator training draws fresh responses from the initial checkpoint, matching SePT's prompts, order, sampling temperature schedule, response cap and student optimizer.
On the common 500-prompt MATH-500 panel, SePT/Base-generator training Pass@1 is 64.975/41.175 at source acquisition, 67.375/35.550 at the primary late assessment, and 67.100/31.725 at the final assessment.
The corresponding gaps are 23.800 [21.050,26.650], 31.825 [28.825,34.925] and 35.375 [32.800,38.075] percentage points.
From source acquisition to the final assessment, SePT gains 2.125 points while Base-generator training loses 9.450; both changes contribute to the growing gap.
At the fixed final assessment, the six-benchmark macro differences are +20.164 Pass@1 [17.917,22.568] and +14.588 Pass@8 [10.031,19.121].
The control tests generator updates jointly, including changes in both visited prefixes and token targets.
\paragraph{An additional seed reproduces the early first-draw contrast.}
An independent training seed repeats SePT and Base-generator training at the prespecified early endpoint, using the shared protocol in Table~\ref{tab:source-recipe}.
SePT/Base-generator training reaches 65.025/40.650 MATH-500 Pass@1, a +24.375 [21.650,27.125] point gap. Pass@8 is 87.600/85.200, with gap +2.400 [$-0.600$,5.400], giving clearer evidence for first-draw accuracy than for eight-sample coverage.
Intervals use 10,000 paired-question bootstrap draws (Appendix~\ref{app:statistics}). The endpoint is fixed before evaluation and repeats the early comparison, not the full training schedule.
\paragraph{An acquired frozen source teaches a reset student.}
To test the training value of SePT-generated responses without inheriting the generator's parameters, we freeze the SePT-trained generator and reset the student and optimizer to the same initialization as Base-generator training.
SePT-generator transfer uses the same independent-seed prompt order, temperature schedule and training budget, fitting all generated responses.
Its fixed MATH-500 endpoint gives Pass@1/Pass@8 of 64.900/86.400.
Relative to Base-generator training, the Pass@1 gain is +24.250 [21.400,27.050] points; relative to SePT it is $-0.125$ [$-1.850$,1.600] points.
The latter interval does not establish equivalence; transfer minus Base-generator training gives a Pass@8 gap of +1.200 [$-1.800$,4.200] points.
All intervals use the same paired-question draws. This same-architecture distillation test \citep{furlanello2018born} shows that a frozen SePT generator supplies useful supervision to a student that does not inherit its weights.

\paragraph{Cost of training and transferring the generator.}
Table~\ref{tab:source-token-cost} separates student training from prior generator acquisition, which must be included in the full transfer cost. Token counts exclude diagnostics and model selection and do not measure FLOPs or wall time.
\par
\begin{table}[ht]
\centering\small
\caption{Response-token counts for the independent-seed generator controls, in millions. Each procedure uses 300 refreshes and 600 updates. Counts include each generated response once and exclude the three diagnostic panels per student. ``\textemdash'' denotes evaluation tokens excluded from this accounting.}\label{tab:source-token-cost}
\begin{tabular}{@{}lrr@{}}\toprule
Procedure & Training tokens & Endpoint evaluation tokens\\\midrule
SePT &37.887&3.161\\
Base-generator training &44.685&4.072\\
SePT-generator transfer &35.228&3.138\\\midrule
Prior generator acquisition &38.295&\multicolumn{1}{c}{\textemdash}\\\bottomrule
\end{tabular}
\end{table}
\par

\paragraph{Testing continued refresh after source acquisition.}
Fixed-generator continuation freezes the acquired SePT generator while the student continues from the same model and optimizer state as the comparison arm, with learning rate $10^{-7}$ and the same temperature schedule.
SePT/Fixed-generator continuation evaluation panels give 67.825/65.975 at the primary late assessment and 67.825/66.225 at the final assessment.
The gaps are +1.850 [0.100,3.575] and +1.600 [$-0.575$,3.750].
The prespecified materiality margin is two points: neither superiority beyond two points nor equivalence within $\pm2$ points is established.
The frozen generator sustains much of the performance, with a smaller observed benefit from continued refresh.

\subsection{Source identity under the Baseline decoder}\label{app:source-decoder-transfer}
To test whether cooling the initial model provides equally useful supervision, we freeze the initial Qwen2.5-Math-7B model and the SePT generator used in the transfer study. Both sample at $\tau_s=0.3$, using the temperature selected for the Baseline, and supply 38,400 raw responses on the same DSR question encounters. Students start from identical initial weights and fresh optimizers. Every response is consumed once, without correctness, format or length filtering.

Each run uses 300 refreshes and 600 optimizer calls, global response-token-mean likelihood at $\tau_t=1$, AdamW with learning rate $10^{-7}$ and a ten-refresh warmup. The prompt/response allowances are 1,024/3,072 tokens. The source sampler has full support and ordinary end-of-sequence stopping. Endpoint evaluation uses all 500 MATH-500 questions, eight responses per question and $\tau_{\mathrm{eval}}=1$; no checkpoint is selected in this comparison.

\par
\begin{table}[ht]
\centering\small
\caption{Fixed-endpoint students taught by different frozen sources at the same sampling temperature $\tau_s=0.3$. Differences and paired question-bootstrap intervals are percentage points.}\label{tab:source-decoder-endpoints}
\begin{tabular}{@{}lrr@{}}\toprule
Training source & Pass@1 & Pass@8\\\midrule
Initial model & 50.15 & 84.60\\
Acquired model & 67.10 & 87.40\\
Acquired minus initial & $16.95\ [14.48,19.48]$ & $2.80\ [0.00,5.60]$\\\bottomrule
\end{tabular}
\end{table}
\par

The SePT generator retains a substantial Pass@1 advantage at this common sampling temperature. Intervals use 10,000 paired-question bootstrap draws (Appendix~\ref{app:statistics}). Temperature $0.3$ was selected for inference, not student learning, and equal temperatures need not yield equal entropy. Because the scheduled-temperature study uses different evaluation and saving cadences, the two studies should not be interpreted as a matched temperature intervention.

\paragraph{What the sources generated.}
We grade responses from 1,024 matched DSR encounters selected before grading (Table~\ref{tab:source-quality}). The SePT generator supplies more correct and extractable answers with fewer length cutoffs. These offline labels describe the improved training data without selecting or weighting responses or isolating the cause of the student gain.
\par
\begin{table}[ht]
\centering\small
\caption{Responses from the initial and SePT-trained generators on 1,024 matched DSR encounters, both sampled at $\tau_s=0.3$. Counts are followed by percentages.}\label{tab:source-quality}
\begin{tabular}{@{}lrr@{}}\toprule
Response property & Initial generator & SePT-trained generator\\\midrule
Correct answer &381 (37.21\%)&506 (49.41\%)\\
Extractable answer &830 (81.05\%)&974 (95.12\%)\\
Response-length cutoff &159 (15.53\%)&34 (3.32\%)\\
Mean length (tokens) &1,144.52&884.81\\\bottomrule
\end{tabular}
\end{table}
\par

Across all 38,400 responses, the acquired source supplies 33,015,377 supervised tokens versus 42,919,911 from the initial source. Source generation, fitting and endpoint evaluations use 44.72 assigned GPU-hours, excluding prior generator acquisition. Thus the stronger student receives fewer tokens, but the complete transfer pipeline also incurs the acquisition cost.

\paragraph{Fitting the same acquired contexts with different objectives.}
We compare the acquired-source likelihood student with a fresh student trained
on the identical response tokens, contexts, masks and minibatches by minimizing
its own full-vocabulary token entropy. Both use the same initialization,
global response-token denominator, optimizer, warmup and 600-update budget.
For the fixed acquired prefixes $h_{it}=(q_i,y_{i,<t})$, the two losses are
\[
L_{\mathrm{NLL}}=-\frac{1}{T_B}\sum_{i,t}\log p_\theta(y_{it}\mid h_{it}),
\qquad
L_H=\frac{1}{T_B}\sum_{i,t}H\!\left(p_\theta(\cdot\mid h_{it})\right),
\]
where $T_B$ counts response tokens in the optimizer minibatch. Entropy is
computed over the full student vocabulary at $\tau_t=1$.
This fixed-context ablation isolates the objective, unlike the full EM-FT comparison, which also changes the generated contexts.

\par
\begin{table}[ht]
\centering\small
\caption{Identical acquired contexts and fitting budget, different objectives.
The primary contrast is likelihood minus entropy Pass@1; Pass@8 is secondary.
Differences and paired 95\% question-bootstrap intervals are percentage points.}
\label{tab:identical-context-objectives}
\begin{tabular}{@{}lrr@{}}\toprule
Fitting objective & Pass@1 & Pass@8\\\midrule
Realized-token likelihood & 67.10 & 87.40\\
Token entropy & 66.93 & 83.20\\
Likelihood minus entropy & $0.18\ [-1.78,2.13]$ & $4.20\ [2.00,6.60]$\\\bottomrule
\end{tabular}
\end{table}
\par

The two students produce 2,684 and 2,677 correct responses out of 4,000, but
solve 437 and 416 questions at least once, respectively; 29 questions are
solved only by the likelihood student and eight only by the entropy student.
Both objectives yield similar observed first-draw accuracy, while likelihood fitting solves more questions within eight attempts. The entropy arm
clips gradients on 82.5\% of updates, with unclipped norms ranging from 0.56
to 9.58. These are fixed-optimizer results, without coefficient or learning-rate
selection. Entropy fitting and endpoint evaluation add 8.46 assigned GPU-hours;
source acquisition is shared. Fitting alone uses 7.80 assigned GPU-hours for
likelihood and 8.01 for entropy. These recorded costs describe the two implementations,
not a general computational ranking of the objectives.

\subsection{Response-budget and sampling temperature diagnostics}\label{app:source-budget}
We test whether extending truncated responses closes the Llama generator-control gap, then compare cooling only the next token with cooling the entire remaining response.

\begin{table}[ht]
\centering\small
\caption{Llama-Instruct response-budget comparison. Differences are Pass@1 points with paired question-bootstrap 95\% intervals; absolute arm accuracies and the continuation procedure are given below.}\label{tab:budget-control}
\begin{tabular}{p{0.60\linewidth}r}\toprule
Contrast & Difference [95\% CI]\\\midrule
Llama SePT--Base-generator training, cap 3,072 &9.825 [7.725,11.925]\\
Llama SePT--Base-generator training, cap 6,144 &9.375 [7.300,11.426]\\
\bottomrule\end{tabular}
\end{table}
\begin{table}[ht]
\centering\small
\caption{Fixed-prefix source probes. Differences are correctness points with paired question-bootstrap 95\% intervals over 60 shared questions, using model-specific prompts requesting boxed answers.}\label{tab:controls}
\begin{tabular}{p{0.60\linewidth}r}\toprule
Contrast & Difference [95\% CI]\\\midrule
Qwen cold first action, common suffix policy &2.500 [$-1.042$,6.250]\\
Qwen cold complete suffix &3.542 [0.625,6.562]\\
Llama cold first action, common suffix policy &1.875 [$-2.083$,6.042]\\
Llama cold complete suffix &1.042 [$-1.458$,3.542]\\
Qwen--Llama complete-suffix interaction &2.500 [$-2.083$,6.875]\\
\bottomrule\end{tabular}
\end{table}

\paragraph{The Llama advantage persists with a larger response budget.}
The Llama-Instruct experiment uses a user-only reasoning prompt without a boxed-answer requirement and symbolic answer verification, identically in both arms.
The user message is ``Question: \{problem\} Answer: Let's think step by step.'' The 1,024-token prompt filter retains 40,309 of 40,315 DSR rows and removes two of 675 OlympiadBench prompts; the initial model context is 4,096 tokens.
We evaluate all 500 MATH-500 questions with eight responses each and retain exact token IDs for continuation.
All 418 responses ending at the response limit (8 SePT, 410 Base-generator training), including already-correct responses, are continued from their exact prefixes with a total budget of 6,144 tokens.
SePT/Base-generator training Pass@1 changes from 44.275/34.450 to 44.275/34.900.
Base-generator training has 25 incorrect-to-correct and seven correct-to-incorrect transitions. The extended-response score replaces the original score, retaining both gains and losses.
Six SePT and 188 Base-generator training responses remain censored.
Assigning worst-case correctness to responses still at the length limit gives terminal-gap bounds [4.800,9.650] points; the lower-bound estimate has interval [2.525,7.100].
Completed-response outcomes stay fixed. For responses still at the limit, the lower bound assigns SePT/Base-generator outcomes 0/1 and the upper bound 1/0; paired-question resampling propagates uncertainty.
The advantage therefore persists under both the extension procedure and the worst-case bound for remaining truncations.

\paragraph{First-action and complete-suffix interventions.}
The probes use the initial Qwen2.5-Math-7B and Llama-3.1-8B-Instruct models, not the trained source-control endpoints.
The 64 MATH-500 questions were fixed before intervention generation.
One temperature-one reference per model and question fixes the candidate reasoning states.
The prefix rule requires at least 32 generated tokens, a position within 20--80\% of the reference before its detected answer region, and a high/low-entropy position gap of at most 10\% of the response.
It prefers punctuation, newline or equation boundaries, then maximizes the entropy difference subject to the recorded eligibility rules.
Qwen retains 60 eligible questions and Llama 64; the contrasts in \Cref{tab:controls} use their 60-question intersection and the high-entropy state for each model.
Shared question identities do not imply identical token prefixes across models.
The sampling temperatures are 0.6 for Qwen and 0.9 for Llama, each compared with sampling at temperature one; both use full-support sampling and prompts requesting boxed answers, unlike the Llama source-control training prompt.
The first-action probe draws eight repetitions per temperature arm, changes only the next-token law and completes the remainder at temperature one.
The complete-suffix probe draws 16 repetitions per arm with the assigned temperature throughout the remainder.
Both retain a 3,072-token total response budget including the fixed prefix.
The reported interaction analysis uses 10,000 paired question-bootstrap draws.
Its cross-model contrast is specific to these model-dependent states and temperatures, not an isolated architecture effect.
Qwen's complete-suffix raw-correctness effect is positive at the sampled prefixes (\Cref{tab:controls}). The direct Qwen-minus-Llama interaction and the Qwen suffix-minus-first-action contrast, +1.042 [$-3.438$,5.523], both have intervals spanning zero. These comparisons do not establish a difference between models or intervention types.

\section{Useful source content}\label{app:source-content}
\subsection{Crossing acquired reasoning contexts with continuation models}\label{app:crossed-prefix}
Using the initial Qwen2.5-Math-7B model and the frozen SePT generator from the transfer study, we test whether changing the supplied prefix or the completing model improves final-answer accuracy. No additional training is performed.

\paragraph{Paired intervention.}
We select 128 MATH-500 questions before generating or grading responses.
Each model supplies one prefix per question at $\tau_{\mathrm{eval}}=1$, capped at 128 tokens and cut before the first recognized final-answer marker or terminal token.
Both prefixes must retain at least 32 tokens; 120 question pairs qualify, and the eight excluded pairs are not replaced.
Adding the cooler initial donor retains 107 common questions under the same eligibility rule (Appendix~\ref{app:baseline-prefix}).
Each fixed prompt-plus-prefix is completed four times by the initial model at $\tau_{\mathrm{eval}}=1$, the acquired model at $\tau_{\mathrm{eval}}=1$, and the initial model at its existing Baseline temperature $\tau_{\mathrm{eval}}=0.3$.
All policies retain full sampling support and a total response budget of 3,072 tokens, including the prefix.
The 2,880 complete responses are checked using the same mathematical answer-verification procedure.
Table~\ref{tab:crossed-prefix-120} reports this two-donor comparison on all 120 eligible questions; Table~\subpanelref{tab:crossed-prefix}{tab:crossed-prefix-donors} reports the three-donor comparison on their 107-question common subset. Both measure continuation accuracy, not the full-policy benchmark metric.
\par
\begin{table}[ht]
\centering\small
\caption{Crossed donors and completers on 120 paired questions. Both donors use $\tau_{\mathrm{eval}}=1$. Each cell averages four continuations per question; scores are conditional continuation accuracy (\%).}\label{tab:crossed-prefix-120}
\setlength{\tabcolsep}{8pt}
\begin{tabular}{@{}lrrr@{}}\toprule
 & \multicolumn{3}{c}{Continuation model}\\\cmidrule(lr){2-4}
Prefix donor & Base Model ($\tau_{\mathrm{eval}}=1$) & SePT ($\tau_{\mathrm{eval}}=1$) & Baseline ($\tau_{\mathrm{eval}}=0.3$)\\\midrule
Base Model &34.2 &41.5 &39.8\\
SePT &58.1 &70.2 &72.5\\\bottomrule
\end{tabular}
\end{table}
\par

\paragraph{Useful contexts and common-context improvement.}
SePT prefixes improve accuracy with the initial completer fixed, and the SePT completer improves accuracy with the initial prefixes fixed. The crossed design separates these two contributions.
Let $M_{ij}$ denote accuracy with prefix source $i$ and completer $j$ at $\tau_{\mathrm{eval}}=1$, where zero denotes initial and one acquired.
For donor-prefix law $d_i$ and remaining-budget success probability $V_j$, $M_{ij}=\mathbb{E}_q\mathbb{E}_{h\sim d_i(\cdot\mid q)}V_j(q,h)$, conditional on paired eligibility.
The primary prefix effect $M_{10}-M_{00}$ is 23.96 [16.67,31.46] points; the primary common-context completer effect $M_{01}-M_{00}$ is 7.29 [2.92,11.88] points.
The corresponding multiplicity-adjusted 97.5\% intervals are [15.63,32.50] and [2.29,12.50].
All intervals resample paired questions, retaining both prefixes and every continuation together; they are conditional on these model states and sampled prefixes.
With the acquired completer, the prefix effect is 28.75 [22.29,35.62] points.
With acquired prefixes, the completer effect is 12.08 [7.08,17.29] points.
The directly estimated interaction is 4.79 [$-1.26$,10.83] points, and the joint change is 36.04 [28.75,43.54] points.
The joint change has two exact paths:
\[
M_{11}-M_{00}=(M_{10}-M_{00})+(M_{11}-M_{10})
             =(M_{01}-M_{00})+(M_{11}-M_{01}).
\]
Averaging the two prefix contributions and the two completer contributions gives symmetric components whose sum is the joint change.
The symmetric prefix and completer components are respectively 26.35 [20.21,32.81] and 9.69 [6.04,13.54] points.

\paragraph{Temperature reference and answer overlap.}
Relative to initial completion at $\tau_{\mathrm{eval}}=0.3$, acquired completion at $\tau_{\mathrm{eval}}=1$ differs by 1.67 [$-2.71$,6.25] points on initial prefixes and $-2.29$ [$-6.46$,2.08] points on acquired prefixes.
The lower-temperature decoder therefore reaches similar point estimates in this conditional comparison.
After freezing the prefix bank, a literal reference-answer overlap check flags at least one prefix in 14 question pairs.
Such matches can be given or intermediate values, while absence of a match does not exclude semantic answer disclosure.
On the remaining 106 pairs, initial-completer accuracy is 31.4\% with initial prefixes and 55.4\% with acquired prefixes, a post-hoc difference of 24.06 [16.51,31.84] points; the primary analysis retains all 120 pairs.
Additional post-hoc sensitivity analyses use the same paired-question procedure: restricting both prefixes to exactly 128 tokens retains 106 pairs and a 20.75 [13.21,28.54]-point initial-completer prefix effect; excluding the previously reviewed training--evaluation overlap flags retains 105 pairs and a 25.48 [17.38,33.57]-point effect.
Combining equal prefix length, no literal answer match, and no reviewed overlap flag retains 83 pairs and a 19.58 [11.14,28.01]-point effect.
These overlapping, post-hoc subgroups reuse the same generations and audit labels. They test sensitivity to the stated filters, not absence of semantic answer disclosure or prior exposure.
The prefix effect also remains positive under the already selected initial completer at $\tau_{\mathrm{eval}}=0.3$: 32.71 [25.00,40.63] points; the primary contrasts remain unchanged.

\subsection{Acquired contexts beyond a cooler initial source}\label{app:baseline-prefix}
To test whether cooling the initial donor supplies equally useful prefixes, we generate one initial-model prefix at its Baseline temperature $\tau_{\mathrm{eval}}=0.3$ for each of the 120 questions above. The same eligibility rules retain 107 questions without replacement. Each cell in Table~\subpanelref{tab:crossed-prefix}{tab:crossed-prefix-donors} uses four continuations per question (428 responses), reusing the $\tau_{\mathrm{eval}}=1$ donor results on this subset. Prompts, verification, full sampling support and the total 3,072-token allowance remain unchanged.

\paragraph{Acquired prefixes outperform the cooler source.}
The primary contrast, specified before generating the cooler-source prefixes, holds the initial $\tau_{\mathrm{eval}}=0.3$ completer fixed and changes only the prefix source from initial $\tau_{\mathrm{eval}}=0.3$ to acquired $\tau_{\mathrm{eval}}=1$.
Accuracy rises from 55.37\% to 69.16\%, a 13.79 [6.07,21.73]-point gain.
With the initial and acquired completers at $\tau_{\mathrm{eval}}=1$, the corresponding secondary gains are 7.71 [0.47,14.95] and 13.08 [5.61,20.79] points.
SePT prefixes therefore improve continuation under all three completers, beyond prefixes from the temperature-tuned initial donor.

\paragraph{Cooling also improves the initial source.}
Compared with initial prefixes generated at $\tau_{\mathrm{eval}}=1$, cooler initial prefixes improve accuracy by 11.45 [3.97,18.93] points with the initial $\tau_{\mathrm{eval}}=1$ completer, 13.55 [6.07,21.03] with the SePT $\tau_{\mathrm{eval}}=1$ completer, and 14.95 [6.78,23.36] with the Baseline $\tau_{\mathrm{eval}}=0.3$ completer. Cooling helps, while SePT prefixes add value beyond this reference.
Square brackets in this subsection give paired-question bootstrap 95\% intervals, retaining all nine cells and four continuations per cell together.
The secondary intervals are nominal, without multiplicity adjustment.

\paragraph{Matching prefix length within each question.}
To distinguish content from additional context, we shorten each acquired prefix to the length of the corresponding cooler prefix, retaining the same 107 questions and initial $\tau_{\mathrm{eval}}=0.3$ completer.
For 19 questions the acquired prefix is shortened; the remaining 88 prefixes already match lengths and their continuations are reused, as are all cooler-prefix cells.
Truncation uses the existing token sequence, with no new instruction or answer-quality selection, and the total response allowance remains 3,072 tokens including the prefix.

\par
\begin{table}[h]
\centering\small
\caption{Supplied-context controls on the same 107 questions with the initial $\tau_{\mathrm{eval}}=0.3$ completer. Prefix and cue conditions use four continuations per question; the no-prefix condition uses eight. All share a 3,072-token total response allowance.}\label{tab:prefix-length-control}
\begin{tabular}{@{}lr@{}}\toprule
Prefix condition & Continuation accuracy (\%)\\\midrule
Cooler initial &55.37\\
Acquired, matched to cooler prefix length &68.46\\
Acquired, full 128-token prefix &69.16\\
No supplied prefix &57.83\\
Generic assistant-start cue &55.37\\\bottomrule
\end{tabular}
\end{table}
\par

Let $B_q$, $A_{m_q}$ and $A_{128}$ denote these respective contexts, and $V(q,h)$ the fixed completer's success probability with the remaining response allowance.
The matched-content and additional-context contrasts are
\[
D_{\rm matched}=\mathbb E_q[V(q,A_{m_q})-V(q,B_q)],\qquad
D_{\rm more}=\mathbb E_q[V(q,A_{128})-V(q,A_{m_q})].
\]
Matching length retains a 13.08 [5.37,21.03]-point acquired-prefix advantage.
Retaining the additional acquired-prefix tokens changes accuracy by 0.70 [$-0.70$,2.10] points.
The two differences sum to the full 13.79-point donor contrast, with most of the gain retained at matched lengths. Paired resampling jointly retains all three conditions and the fixed prefix bank.

\paragraph{Naturally equal-length and shorter-prefix subgroups.}
Length matching changes the supplied prefix within the same 107 questions. A post-hoc subgroup analysis instead divides those questions by the cooler donor's natural stopping behavior (Table~\ref{tab:prefix-subgroups}). The 19 shorter-prefix questions contribute 12.38 of the full 13.79-point donor contrast; 18 reach an answer marker and one a terminal boundary.
Unlike the within-question length control, these subgroups contain different questions. The equal-length analysis in Appendix~\ref{app:crossed-prefix} uses two donors at $\tau_{\mathrm{eval}}=1$, a different comparison.
\par
\begin{table}[ht]
\centering\small
\caption{Post-hoc subgroups of the full-prefix comparison, using the initial $\tau_{\mathrm{eval}}=0.3$ completer. Accuracies are percentages; differences have paired-question 95\% intervals. Unlike length matching, this analysis retains each donor's supplied prefix and partitions the questions.}\label{tab:prefix-subgroups}
\begin{tabular}{@{}lrrr@{}}\toprule
Question group & SePT prefix & Cooler initial prefix & Difference\\\midrule
Both prefixes 128 tokens ($n=88$) &65.34&63.64&1.70 [$-4.83$,8.52]\\
Cooler prefix shorter ($n=19$) &86.84&17.11&69.74 [53.95,84.21]\\\bottomrule
\end{tabular}
\end{table}
\par

\paragraph{Scope of the prefix intervention.}
Prefixes are delimited by token length and answer markers, not by semantic correctness. These boundaries do not guarantee answer-free content. The intervention measures usefulness for continuation, rather than correctness of every intermediate statement.

\paragraph{Comparison with no supplied response prefix.}
For the same 107 questions, we reuse standalone responses generated by the initial model at $\tau_{\mathrm{eval}}=0.3$ in Appendix~\ref{app:decoder-stress}. Question text, prompt tokens, tokenizer, grader and response allowance match, but generation occasions and random seeds differ. The standalone condition uses eight responses per question versus four for each prefix condition (Table~\ref{tab:prefix-length-control}).
The length-matched acquired versus standalone difference is 10.63 [4.44,17.17] points under a paired question bootstrap.
This post-hoc comparison supports usefulness relative to no supplied prefix on the shared question bank. It excludes the cost of training the donor and generating prefixes.

\paragraph{Exploratory comparison with one generic assistant-start cue.}
Replacing the acquired prefix with one fixed 19-token task-start reminder gives 55.37\% accuracy on the same 107 questions (Table~\ref{tab:prefix-length-control}).
The length-matched acquired prefix exceeds this cue by 13.08 [7.71,18.69] points. The cue alone gives no clear improvement over no prefix: $-2.45$ [$-8.76$,3.97] points.
This exploratory comparison tests one cue of a different length, not generic prompting in general. The exact cue and execution records are retained with the research artifacts.

\paragraph{Sensitivity and response length.}
On the 92 questions unflagged by the existing reviewed overlap record, the acquired-versus-cooler donor contrast is 13.86 [5.43,22.55] points; excluding the four natural early endings leaves 103 questions and 12.62 [4.85,20.87] points.
These post-hoc subsets reuse the primary evaluations and have nominal paired-question intervals.
With the initial $\tau_{\mathrm{eval}}=0.3$ completer, the acquired and cooler prefix conditions average 766.10 and 1,085.46 logical prefix-plus-suffix tokens per completion, respectively.
The higher accuracy uses fewer emitted tokens on average; these lengths do not account for training or donor costs.

\paragraph{Coverage and interpretation.}
The cooler donor's answer markers occur in 31 of 120 prefixes; the mechanical truncation rule leaves 13 prefixes below the minimum length.
Four generated prefixes end naturally within the short allowance, and two of these complete answers are correct under separate descriptive grading.
Those verdicts do not enter the continuation scores. Results concern completion from eligible supplied prefixes, with the models and prefix bank fixed during resampling.

\subsection{Fixed-context token ordering}\label{app:rank-change}
The continuation experiments test the usefulness of generated contexts. We now hold the context identical for both models and test whether their conditional token distributions differ beyond scalar temperature rescaling.
For a fixed context $h$, a positive scalar rescaling
$z_\star(h)=a z_0(h)+c(h)\mathbf{1}$, $a>0$, preserves every strict token rank, even if $a$ varies with the context.
We score full-vocabulary next-token logits on the 240 prefixes generated at $\tau_{\mathrm{eval}}=1$ in the 120-question, two-donor study in Appendix~\ref{app:crossed-prefix}.
Repeating the initial-model forward pass yields identical stored logits.
Seven contexts have a top-token reversal with both competing-token margins exceeding 0.125: five of 120 initial prefixes and two of 120 acquired prefixes.
One additional small-margin reversal is not counted as robust.
A label-free fit of a single inverse temperature over $[0.1,10]$ gives $a=1.104$ and mean forward KL residual 0.0462 nats.
These reversals establish changes beyond scalar temperature rescaling. The continuation experiments separately test usefulness; they do not attribute the accuracy gain to the reversed tokens.

\section{Theoretical analysis and proofs}\label{app:reference-proofs}
The analysis below connects the likelihood target to continuation value, generator refresh and shared-parameter fitting.
The reference objective assumes positive temperatures, a finite horizon,
positive fixed support and a teacher frozen within the round. Include all
modeled sampled actions, including emitted EOS; an imposed cap adds no
unsampled EOS probability. Full-softmax temperature identities do not cover
support-changing top-$k$/top-$p$ truncation without qualification.
The logit identity also gives
$z_i^*(\zeta)-z_j^*(\zeta)=(\tau_t/\tau_s)[z_i^{\rm old}(\zeta)-z_j^{\rm old}(\zeta)]$.
\begin{proposition}[Sequence-level KL decomposition]\label{prop:osft_seq}
Let $d_b(\zeta)$ be the expected number of visits to prefix $\zeta$ under the frozen source. For a constant $C_b$ independent of the student,
\[
L_b(\theta)=C_b+\sum_\zeta d_b(\zeta)\KL\bigl(b(\cdot\mid\zeta)\|p_\theta(\cdot\mid\zeta)\bigr).
\]
\end{proposition}
\begin{proof}[Proof of Proposition~\ref{prop:osft_seq}]
Factor each response probability into modeled conditional actions, exchange the finite expectation and sum, and group visits by prefix.
This gives $L_b=\sum_\zeta d_b(\zeta)\sum_a-b(a\mid\zeta)\log p_\theta(a\mid\zeta)$.
Use $H(b,p)=H(b)+\KL(b\|p)$; the summed entropies form $C_b$, independent of the student.
\end{proof}
\paragraph{Ideal temperature-ratio target.}
If each visited conditional is fitted independently and exactly, the sequence-sum objective gives $p_\theta^*=b$, hence
\[
z^*(\zeta)=\frac{\tau_t}{\tau_s}z_{\rm old}(\zeta)+c(\zeta)\mathbf1.
\]
\begin{proof}
For each positive-weight independent conditional, the KL is minimized uniquely in probability space at $p_\theta=b$.
Equality of the two softmax distributions means their logit vectors divided by temperature differ only by an all-token constant.
Multiply by $\tau_t$ and subtract any two components.
No existence of a shared parameter vector attaining every pointwise optimum is assumed.
\end{proof}
Under these assumptions, repeated independent exact fitting remains equivalent to temperature scaling on the fitted prefix support. With full support and the same response horizon in each round, successive temperature ratios multiply.
To connect this target to task performance, let continuation value denote the expected terminal utility after choosing a token and then following a fixed continuation policy. For a correctness indicator, this is the probability of eventually reaching a correct answer.
\begin{lemma}[Conditional source value]\label{lem:mechanism-source}
At a fixed prefix, let $p(a)>0$, $b_\alpha(a)=p(a)^\alpha/Z_\alpha$, $Z_\alpha=\sum_a p(a)^\alpha$, $\alpha=\tau_t/\tau_s$.
For bounded continuation value $V$ held fixed,
\[
\mathbb E_{b_\alpha}V-\mathbb E_pV=
\frac{\operatorname{Cov}_p(V,p^{\alpha-1})}{\mathbb E_p[p^{\alpha-1}]},\qquad
\partial_\alpha\mathbb E_{b_\alpha}V=\operatorname{Cov}_{b_\alpha}(V,\log p).
\]
\end{lemma}
\begin{proof}[Proof of the conditional source-value lemma]
Write $\mathbb E_{b_\alpha}V=\mathbb E_p[p^{\alpha-1}V]/\mathbb E_p[p^{\alpha-1}]$ and subtract $\mathbb E_pV$.
Differentiating the normalized finite sum yields the log-probability covariance identity.
\end{proof}
Tokenwise tempering is not global sequence-power sampling \citep{karan2025reasoning}:
$b^{\rm tok}_\alpha(o\mid q)=p(o\mid q)^\alpha/\prod_t Z_\alpha(\zeta_t)$.
The product of prefix-dependent normalizers generally varies across trajectories.

\paragraph{Connection to entropy minimization.}
At $\tau_t=1$, for fixed positive $p=\operatorname{softmax}(z)$ and $\alpha=1+\epsilon$, Taylor expansion gives
\[
\begin{aligned}
b_{1+\epsilon}-p
&=\epsilon p\odot(\log p-\mathbb E_p\log p)+O(\epsilon^2)\\
&=-\epsilon\nabla_z H(p)+O(\epsilon^2).
\end{aligned}
\]
For a fixed strictly positive probability vector $p$, this expansion holds as $\epsilon$ tends to zero.
The local directions agree to first order, while the finite objectives differ: likelihood fits sampled tokens and entropy minimization differentiates the full conditional distribution. Appendix~\ref{app:source-decoder-transfer} compares the objectives on identical contexts. Token averaging changes the expected teacher as derived in Appendix~\ref{app:implementation-objective}.

\subsection{Cumulative source value and local shared projection}
\label{app:mechanism-proofs}
We use the sequence-sum objective and assumptions of Appendix~\ref{app:reference-proofs}. Smoothness through three derivatives suffices for the local Taylor remainders. The results connect sampling, generator refresh and local changes in utility; they do not guarantee finite-time accuracy improvement.

\begin{lemma}[Cumulative source value]\label{lem:mechanism-cumulative}
For fixed initial prefix $\zeta_0$, terminal utility $R\in[0,1]$, target
continuation $p$ and cold continuation $b_\alpha$, let $Q^p(\zeta,a)$ be
the expected terminal utility after action $a$ followed by $p$. Then
\[
J(b_\alpha\mid\zeta_0)-J(p\mid\zeta_0)=
\sum_\zeta d_{b_\alpha}^{\zeta_0}(\zeta)
\frac{\operatorname{Cov}_{a\sim p(\cdot\mid\zeta)}
       (Q^p(\zeta,a),p(a\mid\zeta)^{\alpha-1})}
{\mathbb E_{a\sim p(\cdot\mid\zeta)}p(a\mid\zeta)^{\alpha-1}}.
\]
\end{lemma}
\begin{proof}
Let $V^p(\zeta)=\mathbb E_p Q^p(\zeta,a)$, with terminal value $R$ at EOS
or the cap. Telescoping $V^p$ along a $b_\alpha$ trajectory gives
$J(b_\alpha)-J(p)=\mathbb E_{b_\alpha}\sum_t
[Q^p(\zeta_t,a_t)-V^p(\zeta_t)]$.
Group by visited prefix and apply \Cref{lem:mechanism-source}.
This specializes the standard performance-difference identity \citep{schulman2015trust}.
\end{proof}
The identity applies separately to correctness, extraction and natural stopping. Correctness conditional on extraction is a ratio of expectations and requires separate treatment.

\paragraph{Fisher form of teacher mismatch.}
For teacher and student distributions at a common temperature and parameter anchor $\bar\theta$, define
\[
g(\theta_s,\theta_g)=\mathbb E_{q,o\sim p_{\theta_g}}
\nabla_{\theta_s}\log p_{\theta_s}(o\mid q),\quad
s=\left.\nabla_\theta\log p_\theta(o\mid q)\right|_{\bar\theta},\quad
F=\mathbb E_{p_{\bar\theta}}[ss^\top].
\]
Under the assumptions above,
\begin{equation}\label{eq:mechanism-fisher}
g(\bar\theta,\bar\theta)=0,\quad D_sg=-F,\quad D_gg=F,\qquad
g(\bar\theta+u,\bar\theta+v)=-F(u-v)
+O((\|u\|+\|v\|)^2).
\end{equation}
\begin{proof}
Differentiating total probability gives $\mathbb E[s]=0$ and then
$\mathbb E[\nabla^2\log p]=-F$ for the student derivative. Differentiating
the teacher density instead yields $\mathbb E[ss^\top]=F$.
Joint Taylor expansion proves the final identity.
\end{proof}
A memoryless fixed-preconditioner update has first-order direction
$-\eta MF(\theta_s-\theta_g)$. The derivatives cancel along the matched
teacher/student diagonal; a frozen teacher supplies local mismatch correction.
This local result does not characterize the full AdamW trajectory, which also depends on sampling, normalization, teacher lag and optimizer state. In particular, it does not identify the observed performance decline as parameter-space reversion.

\paragraph{Refresh changes exposure and conditional supervision.}
At fixed student $\theta$, write $s_\theta(\zeta,a)=
\nabla_\theta\log p_\theta(a\mid\zeta)$ and
$g(\theta,\phi)=\sum_\zeta d_\phi(\zeta)\mathbb E_{b_\phi}s_\theta$.
For two teachers,
\begin{align}
g(\theta,\phi_1)-g(\theta,\phi_0)
={}&\sum_\zeta(d_1-d_0)(\zeta)\mathbb E_{b_1}s_\theta(\zeta,a)\notag\\
&+\sum_\zeta d_0(\zeta)\sum_a(b_1-b_0)(a\mid\zeta)s_\theta(\zeta,a).
\label{eq:mechanism-refresh}
\end{align}
\begin{proof}
Add and subtract $\sum_\zeta d_0(\zeta)\mathbb E_{b_1}s_\theta$.
\end{proof}
The two terms separate changes in prefix exposure and conditional targets. The experiment changes both together. For token-averaged likelihood, use the effective teacher and weights from Appendix~\ref{app:implementation-objective}.

\paragraph{Anchored score-kernel identity.}
This identity relates the sampling-induced training direction to evaluation utility through gradient alignment across contexts.
For a fixed direction $g$ at the current student, Taylor expansion gives $J_1(\theta+\eta g)-J_1(\theta)=\eta\nabla J_1(\theta)^\top g+O(\eta^2)$. The expression below expands this inner product for the likelihood direction, including a fixed preconditioner $M$.
At student anchor $\theta=\theta_{\rm old}$ let $p$ be its training law at $\tau_t$
and $b$ the source law, with common positive support and the same fixed prompt distribution $\mathcal D$. For a training pair $X=(q_X,o_X)$, write $\pi_\theta(X;\tau_t)$ for $\pi_\theta(o_X\mid q_X;\tau_t)$.
Set $s_t(X)=\nabla_\theta\log\pi_\theta(X;\tau_t)$,
$w(X)=b(X)/p(X)-1$, and $g_S=\mathbb E_p[w(X)s_t(X)]$.
Independently of $X\sim p$, draw an evaluation prompt $q_Y\sim\mathcal D_{\rm eval}$
and response $o_Y\sim\pi_\theta(\cdot\mid q_Y;1)$, and write $Y=(q_Y,o_Y)$.
For bounded terminal reward $R(Y)$, define
$J_q(\theta)=\mathbb E_{o\sim\pi_\theta(\cdot\mid q;1)}R(q,o)$,
$J_1(\theta)=\mathbb E_{q\sim\mathcal D_{\rm eval}}J_q(\theta)$ and
$s_1(Y)=\nabla_\theta\log\pi_\theta(o_Y\mid q_Y;1)$.
The prompt law is fixed, and scores and expectations below are evaluated at the
anchor. For fixed positive-semidefinite preconditioner $M$,
\begin{equation}\label{eq:mechanism-kernel}
\left.\frac{dJ_1(\theta+\eta Mg_S)}{d\eta}\right|_0
=\mathbb E_{X,Y}
\left[w(X)(R(Y)-J_{q_Y})s_1(Y)^\top M s_t(X)\right].
\end{equation}
\begin{proof}
Score-function differentiation gives
$\nabla J_1=\mathbb E[(R-J_q)s_1]$.
At the stated anchor, $\mathbb E_p s_t=0$, hence
$\mathbb E_b s_t=\mathbb E_p[(b/p-1)s_t]$.
Substitute in $\nabla J_1^\top M g_S$ and use independent train/evaluation
trajectories.
\end{proof}
Away from the matched student anchor use the direct uncentered teacher score.
For the expected executed ratio, substitute the effective teachers and their
weights, not unmodified $b$. A favorable source is neither sufficient for
every finite shared update nor necessary for every useful local transfer.
Appendix~\ref{app:corruption} tests one such coexistence experimentally: pre-answer likelihood transfer while fitting a synthetically incorrect ending.

\paragraph{First-draw accuracy and aggregate coverage.}
The relationship between Pass@1 and AVG depends on how improvements are distributed across questions. For prompt success $u_q$, put
$f(u)=[u+1-(1-u)^8+1-(1-u)^{32}]/3$. Under independent draws and the six-suite
protocol, the chain rule gives
\[
\nabla J_{\rm AVG}=\frac16\sum_{d=1}^6\mathbb E_{q\in d}
[f'(u_q)\nabla u_q],\qquad
f'(u)=\frac{1+8(1-u)^7+32(1-u)^{31}}3.
\]
The weights are positive but differ across questions. An aggregate Pass@1 increase therefore need not imply AVG improvement when some questions deteriorate.

\subsection{Response-token averaging and the effective teacher}
\label{app:implementation-objective}
Equation~\ref{eq:reference} sums likelihood over each response. The generator controls instead average it over all response tokens in an optimizer minibatch. Random response lengths make this denominator random, changing the expected objective. Consequently, the zero expected gradient for matched teacher and student at equal temperatures need not hold for the token mean.

The teacher is fixed across each round's updates. Sampled EOS tokens are counted, and a hard cap adds no unsampled EOS. Distributed accumulation implements the global minibatch token mean. The separate corruption assay uses the normalization specified in Appendix~\ref{app:corruption}.

\begin{lemma}[Expected token ratio as effective-teacher projection]
\label{lem:mechanism-effective}
Condition on a fixed prompt batch $q_1,\ldots,q_B$. Independently draw
$O_i\sim b_i(\cdot\mid q_i)$ from fixed teachers with positive finite support.
Let $T_i=T(O_i)\ge1$ count all modeled response actions, including sampled EOS.
For $S_B=\sum_iT_i$, define
\[
\widehat L_B(\theta)=-\frac{\sum_i\log p_\theta(O_i\mid q_i)}{S_B},
\quad
\omega_i(o)=\mathbb E_{O_{-i}}\frac1{T(o)+\sum_{j\ne i}T(O_j)},
\]
\[
c_i=\mathbb E_{b_i}\omega_i(O_i)>0,\qquad
\widetilde b_i(o)=b_i(o)\omega_i(o)/c_i.
\]
Then
\begin{equation}\label{eq:mechanism-effective}
\mathbb E\widehat L_B(\theta)
=\sum_i c_i\left[H(\widetilde b_i)
+\operatorname{KL}(\widetilde b_i\|p_\theta(\cdot\mid q_i))\right],
\qquad c_i=\mathbb E[1/S_B].
\end{equation}
\end{lemma}
\begin{proof}
Expand the numerator and condition its $i$th term on $O_i=o$.
Integrating the other responses gives
$-\sum_{i,o}b_i(o)\omega_i(o)\log p_\theta(o\mid q_i)$.
Normalize each $b_i\omega_i$ and apply the cross-entropy identity.
The tower property gives $c_i=\mathbb E[1/S_B]$ for every occurrence in
the same conditioned batch; it can differ across batch compositions.
\end{proof}
This is the expected raw likelihood objective of a fixed teacher, not an
identity for an entire adaptive AdamW trajectory.

\paragraph{Effective prefix target.}
For a nonterminal prefix of occurrence $i$, let
$m_i(\zeta,a)=\mathbb E_{b_i}[\omega_i(O_i)\mid\zeta,a]$ and
$m_i(\zeta)=\mathbb E_{b_i}[\omega_i(O_i)\mid\zeta]$.
Marginalizing the tilted trajectory law to prefix and prefix--action events
and dividing yields
\[
\widetilde b_i(a\mid\zeta)
=b_i(a\mid\zeta)\frac{m_i(\zeta,a)}{m_i(\zeta)}.
\]
For a full-softmax teacher $b_i(\cdot\mid\zeta)=
\operatorname{softmax}(z_\phi(\zeta)/\tau_s)$, independently fitting the
effective conditional at $\tau_t$ therefore gives
\begin{equation}\label{eq:mechanism-logit-correction}
z_i^*(\zeta,a)=\frac{\tau_t}{\tau_s}z_\phi(\zeta,a)
+\tau_t\log m_i(\zeta,a)+\kappa_i(\zeta).
\end{equation}
This follows by taking log probabilities, with $\kappa_i$ an
action-independent shift. For repeated prompts, combine objective-weighted
prefix masses before normalizing:
\[
\widetilde b(a\mid q,\zeta)=
\frac{\sum_{i:q_i=q}c_i\widetilde d_i(\zeta)\widetilde b_i(a\mid\zeta)}
{\sum_{i:q_i=q}c_i\widetilde d_i(\zeta)}.
\]
Here $\widetilde d_i$ is prefix reach probability under the tilted law.
The formula follows by grouping cross-entropy contributions for the shared
conditional. For random batch compositions also integrate these weighted
contributions over the batch law; do not average conditionals unweighted.

\paragraph{Length and neutrality qualifications.}
For fixed distribution of other lengths, $\omega_i(T)$ decreases with $T$
because $1/(T+U)$ decreases for every possible sum $U$. A constant
denominator or fixed lengths give action-independent $m_i$, recovering
the original teacher and temperature-ratio optimum. Within one realized
batch the denominator is a common scalar; the effective-teacher statement
concerns expectation over random batches.
At matched student and teacher, finite-support score differentiation gives
\[
\mathbb E\left[\frac{\sum_i s_i}{S_B}\right]
=\nabla_\theta\mathbb E_{\prod_i p_\theta(\cdot\mid q_i)}[1/S_B],
\qquad s_i=\nabla_\theta\log p_\theta(O_i\mid q_i).
\]
Indeed, the joint sampling score is $\sum_i s_i$; apply the score-function
identity to $1/S_B$. This mathematical derivative does not backpropagate
through sampled discrete tokens. For one Bernoulli-logit model with a
length-one outcome of probability $p$ and a length-two outcome of probability
$1-p$, $\mathbb E[s/T]=p(1-p)/2>0$ although $\mathbb E[s]=0$.
In contrast, dividing the sequence-sum population objective by a fixed
teacher's population mean length only rescales it positively.

These identities describe how token averaging changes the expected target and equal-temperature gradient. They do not identify normalization as the cause of the empirical gains or imply that shorter responses are more correct.

\begingroup
\let\section\subsection
\section{A constructive shared-parameter example}
\label{app:constructive-projection}
This example shows how shared-parameter optimization can fit a temperature-scaled teacher at a training context yet produce predictions elsewhere that no scalar decoding temperature can reproduce.

Let $p_{a,b}(1\mid x)=\sigma(ax+b)$, where
$\sigma(u)=(1+e^{-u})^{-1}$, and initialize $(a,b)=(1,0)$.
At the sole training context $x=1$, the fixed initial teacher at temperature
$1/\alpha$, $\alpha>1$, samples action one with probability
$q=\sigma(\alpha)$. Fit all raw teacher samples in population, without
correctness feedback, at student temperature one:
\[
L(a,b)=-q\log\sigma(a+b)-(1-q)\log[1-\sigma(a+b)],
\qquad \partial_aL=\partial_bL=\sigma(a+b)-q.
\]
Euclidean gradient flow preserves $a-b=1$. Its sum $s=a+b$ obeys
$\dot s=-2[\sigma(s)-\sigma(\alpha)]$ and increases from $1$ to the
unique equilibrium $\alpha$. Thus its limiting parameters are
\[
a_*=(\alpha+1)/2,\qquad b_*=(\alpha-1)/2.
\]
The different parameters $(\alpha,0)$ also fit the teacher exactly at
$x=1$ and implement global temperature scaling. Fitting this training
distribution therefore does not uniquely determine behavior elsewhere;
the optimization trajectory selects between these solutions.

Now evaluate at $x=-\delta$, with
$0<\delta<(\alpha-1)/(\alpha+1)$, and stipulate that action one is correct.
For every positive initial-model decoding temperature $\tau$,
\[
p_{1,0}(1\mid-\delta;\tau)=\sigma(-\delta/\tau)<1/2,
\qquad \sup_{\tau>0}p_{1,0}(1\mid-\delta;\tau)=1/2.
\]
In contrast, the fitted model at temperature one has logit
\[
-a_*\delta+b_*=
\frac{(\alpha-1)-(\alpha+1)\delta}{2}>0,
\]
so its success probability exceeds that entire initial temperature family.
For $\alpha=2$ and $\delta=0.1$, initial success at temperature one is
$0.4750$, the temperature-family supremum is $0.5000$, and fitted success
is $0.5866$. The alternative global-scaling optimum gives $0.4502$.

This benefit requires alignment with the evaluation utility: if action zero
is correct instead, the same update reduces success from $0.5250$ to
$0.4134$. The example isolates extrapolation under population cross-entropy and Euclidean gradient flow. It demonstrates a possible outcome of shared fitting, while the LLM experiments test whether learned responses are useful in practice.

\endgroup
\section{General-reasoning training and transfer}\label{app:transfer-assays}\label{app:nonmath}
We train on non-mathematical reasoning questions and evaluate transfer to multi-hop question answering. The separate retention test below evaluates general capabilities after mathematical training.
The training set contains 12,349 prompts: 2,251 ARC-Easy and 1,119 ARC-Challenge \citep{clark2018arc}, 7,376 LogiQA \citep{liu2020logiqa} and 1,603 StrategyQA \citep{geva2021strategyqa}.
Each is a user message consisting of the question and choices when applicable, followed by ``Please reason step by step, and put your final answer within \texttt{\textbackslash boxed\{\}}.''
The starting model is Qwen2.5-7B.
The training configuration specifies LR $10^{-7}$ with a warmup of 10 generator refreshes, 128 prompts per refresh, two 64-response optimizer minibatches, one response per encounter, three epochs (288 refreshes planned), a 1,024-token prompt cap and a 3,072-token response cap.
Sampling temperature anneals linearly from 0.9 to one over two epochs, with training temperature fixed at one. The evaluation uses a fixed trained checkpoint.

The evaluation uses LongBench task version 5.0, its test splits and 200 questions per task.
Evaluation is zero-shot and greedy, with a 32-token generation cap, 32,768-token model context and middle truncation.
The evaluation prompt requests only the answer based on the supplied passages.
The base and SePT models are evaluated on identical documents, prompts and targets.

\begin{table}[h]
\centering\small
\caption{Non-mathematical transfer. F1 scores are percentages; differences include paired question-bootstrap 95\% confidence intervals.}\label{tab:nonmath}
\begin{tabular}{lrrr}\toprule
Task & Starting model & SePT & Difference [95\% CI]\\\midrule
2WikiMultihopQA &11.54&14.22&2.681 [0.861,4.784]\\
HotpotQA &17.37&21.56&4.189 [1.426,7.194]\\
MuSiQue &6.11&7.53&1.419 [0.098,3.161]\\
Macro &11.67&14.44&2.763 [1.553,4.086]\\\bottomrule
\end{tabular}
\end{table}
The bootstrap uses 20,000 replicates, resampling paired questions independently within each task and averaging the three task means equally.
The macro is computed from unrounded task means; subtracting displayed rounded cells need not reproduce its final digits.

\subsection{General capabilities after mathematical training}\label{app:retention}
We evaluate the initial Qwen2.5-Math-7B model and its SePT- and GRPO-trained counterparts using the Language Model Evaluation Harness \citep{eval-harness} on IFEval, BBH, GPQA, MuSR and MMLU-Pro \citep{zhou2023instructionfollowing,suzgun2022challenging,rein2023gpqa,sprague2024musrtestinglimitschainofthought,wang2024mmluprorobustchallengingmultitask}.
Table~\ref{tab:original-retention} shows a stable average score on these tasks, with small task-level changes in both directions.
\begin{table}[h]
\centering\small
\caption{General-capability retention of Qwen2.5-Math-7B after mathematical training.}\label{tab:original-retention}
\begin{tabular}{lrrrrrr}\toprule
Method & IFEval & BBH & GPQA & MuSR & MMLU-Pro & Mean\\\midrule
Initial model &23.4&47.5&29.9&41.4&32.1&34.9\\
SePT &23.6&47.3&30.6&41.5&32.2&35.0\\
GRPO &24.6&47.7&30.1&41.4&32.3&35.2\\\bottomrule
\end{tabular}
\end{table}

\section{Transfer while fitting a corrupted ending}\label{app:corruption}
We test whether fitting an incorrect ending can coexist with improved likelihood on pre-answer content for held-out questions. The diagnostic measures these two effects separately, rather than benchmark accuracy or correction of natural errors.
The experiment uses Qwen2.5-Math-7B responses from the first 16,000 rows of a MATH-500 $n=32$ sweep at sampling temperature 0.6.
We select responses whose final answers pass the verifier and are canonical unsigned integers. Counterbalanced question pairs swap a units digit in the final boxed answer, changing exactly one token per response while preserving the pair's changed-token histogram. Each corrupted answer fails verification.
The split contains 128 train, 32 development and 96 test questions, with whole pairs kept within disjoint splits.

\paragraph{Training targets and update.}
The four arms fit the complete response with a correct final answer, the complete corrupted response, its pre-answer content, or its answer-bearing tail. Loss masks isolate these targets while retaining the full causal context.
Stored token IDs are consumed without reapplying a chat template.
The full mask $F=[s,e)$ covers response content, excluding prompt, padding and chat-ending tokens.
The shared-prefix mask $P=[s,t)$ ends at the token boundary at or before the first standalone occurrence of the clean numeric answer before the final payload, or at the payload boundary if there is no earlier occurrence.
The answer-bearing tail $A=[t,e)$ partitions $F$ with $P$ and can contain valid content: it is not a pure wrong-token mask.
Correct-full uses clean tokens and $F$; the other arms use the corrupted sequence with $F$, $P$ or $A$, retaining full causal context.
Each arm starts from the same base checkpoint and fresh AdamW state and makes two 64-example optimizer updates, with decay coefficients $(0.9,0.999)$, $\epsilon=10^{-8}$, weight decay 0.01 and global gradient clipping at 1.0.
All eight saved pre-clip norms exceed 1.
The optimized loss averages per-GPU microbatch ratios, each with the matched \emph{full-response} token denominator even for masked numerators; it is not the global-minibatch ratio of the source-control training runs.
The constant LR $10^{-6}$ was selected from $\{3\!\times\!10^{-8},10^{-7},3\!\times\!10^{-7},10^{-6}\}$ using train/development data: require finite metrics, positive correct-full development prefix gain and positive wrong-payload training gain for both full-corrupted and tail arms, then maximize full-corrupted development prefix gain (lower LR breaks ties).
Test measurements are excluded from selection. This two-update diagnostic uses no scheduler or inherited optimizer state, and its AdamW arm effects need not be additive.

\paragraph{Assay and uncertainty.}
Wrong-answer log-probability gain is the training-question mean of post-minus-pre summed log probability over the corrupted payload.
Transfer is the test-question mean of pre-minus-post token-mean NLL over the clean shared prefix.
The 95\% intervals resample whole counterbalanced pairs (64 train or 48 test pairs), using 20,000 replicates.
They are conditional on this single selected pulse, not training-seed or selection uncertainty.

\begin{table}[h]
\centering\small
\caption{Terminal-corruption diagnostic. Wrong-ending log-probability gains are in nats; pre-answer NLL reductions are in nats per token. Square brackets contain 95\% confidence intervals obtained by resampling counterbalanced question pairs.}\label{tab:corruption}
\begin{tabular}{p{0.32\linewidth}rr}\toprule
Training target & Wrong-ending logp gain & Pre-answer NLL reduction\\\midrule
Complete correct & $-0.0416$ [$-0.0640$,$-0.0195$] &0.00469 [0.00223,0.00751]\\
Complete corrupted &0.0804 [0.0613,0.1001]&0.00466 [0.00272,0.00696]\\
Pre-answer only &$-0.0264$ [$-0.0476$,$-0.0053$]&0.00451 [0.00189,0.00781]\\
Answer-bearing tail only &0.1806 [0.1556,0.2054]&$-0.00069$ [$-0.00219$,0.00067]\\\bottomrule
\end{tabular}
\end{table}
The complete-corrupted minus answer-bearing-tail transfer contrast is +0.00534 [0.00242,0.00887].
Paired measurements and token masks are retained with the research records.

\section{Comparison with related methods}
\subsection{Entropy-minimization training}\label{app:emft}
\begin{figure}[h]
\centering
\includegraphics[width=0.62\linewidth]{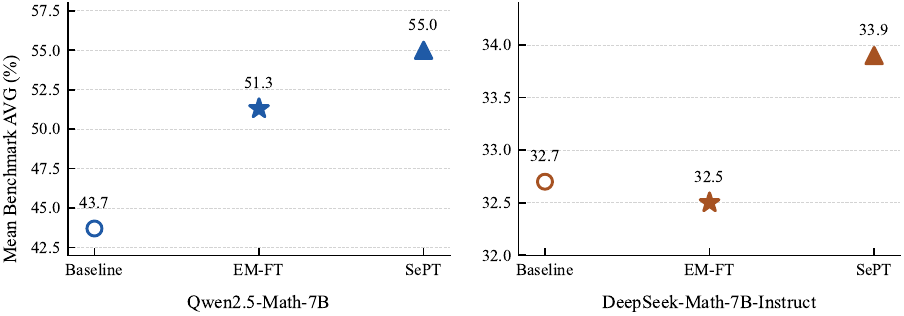}
\caption{AVG comparison of the Baseline, EM-FT and SePT on two starting models. Each point represents a separate evaluated configuration. SePT achieves higher AVG under the specified training recipes.}\label{fig:em}
\end{figure}

EM-FT uses the same configuration for Qwen2.5-Math-7B and DeepSeek-Math-7B-Instruct, sampling and training at temperature one.
Its objective minimizes the token-mean entropy of the next-token distribution on the sampled prefixes, with coefficient one and no KL regularizer.
It uses the same starting model as SePT, DSR prompts, one rollout, 128-prompt refreshes, 64-response optimizer minibatches, one optimization epoch per generated batch, LR $10^{-7}$ and a warmup of 10 generator refreshes, and 1,024/3,072 prompt/response caps.
Training uses eight GPUs with microbatch size four per GPU.
The comparison changes both the objective and sampling schedule: EM-FT stays at one, while SePT anneals from 0.6 to one. Figure~\ref{fig:em} and Table~\ref{tab:appendix_table3_em_dsr_breakdown} report these recipe results. Appendix~\ref{app:source-decoder-transfer} separately compares the objectives on identical SePT-generated responses.

\begin{table}[h]
\centering\small
\caption{Training signals and response generation in related methods and SePT controls. The lower block uses unfiltered responses and likelihood training; its variants differ in whether responses or generator weights are held fixed.}\label{tab:methods}
\setlength{\tabcolsep}{5pt}
\renewcommand{\arraystretch}{1.12}
\begin{tabularx}{\linewidth}{@{}>{\raggedright\arraybackslash}p{.23\linewidth}>{\raggedright\arraybackslash}p{.17\linewidth}>{\raggedright\arraybackslash}p{.29\linewidth}>{\raggedright\arraybackslash}X@{}}\toprule
\textbf{Method} & \textbf{Feedback} & \textbf{Training update} & \textbf{Response generator}\\\midrule
\textbf{STaR} & Ground-truth answers & Likelihood on selected or rationalized responses & Updated model\\\addlinespace
\textbf{TTRL} & Answer consensus & Consensus-reward RL & Updated model\\\addlinespace
\textbf{EM-FT} & None & Entropy minimization & Updated model\\\midrule
\textbf{SePT (Offline)} & None & Likelihood on a fixed dataset & Initial model; sampled once\\\addlinespace
\textbf{Base-generator training} & None & Likelihood on fresh responses & Fixed initial model\\\addlinespace
\textbf{Fixed-generator continuation} & None & Likelihood on fresh responses & Fixed trained model\\\addlinespace
\rowcolor{septwash}\textbf{SePT} & None & Likelihood on fresh responses & Updated model\\\bottomrule
\end{tabularx}
\end{table}

\section{Training configurations and illustrative responses}\label{app:original-details}
\subsection{Training recipe}
The method-comparison experiments use eight A800 GPUs, AdamW with weight decay 0.01, a ten-refresh warmup and three training epochs.
The recorded learning rates are $10^{-7}$ for SePT and $5\times10^{-7}$ for GRPO; the KL regularizer is disabled.
Each refresh samples 128 prompts, split into two 64-prompt optimization minibatches.
SePT uses one response per prompt, while GRPO uses eight; equal prompt counts therefore do not imply equal sequence or token counts.
The source-control experiments use four GPUs per arm; Appendix~\ref{app:study-settings} specifies their initialization and training horizons.

The SePT sampling temperature schedule, in epoch progress $u$, is
\[
\tau_s(u)=
\begin{cases}
\tau_{s,\mathrm{start}}+\dfrac{u}{2}(1-\tau_{s,\mathrm{start}}),&0\le u<2,\\
1,&2\le u\le3.
\end{cases}
\]
The starting value is 0.6 for Qwen2.5-Math-7B and DeepSeek-Math-7B-Instruct, and 0.9 for the other models; the training temperature remains one.
EM-FT uses the configuration in Appendix~\ref{app:emft}.

\paragraph{Fixed-data ablation.}\label{app:offline-protocol}
SePT (Offline) samples one response per DSR prompt from the initial Qwen2.5-Math-7B model at temperature 0.6, yielding 40,306 fixed pairs. A student initialized from the same model reuses them for three likelihood-training epochs. We test learning rates $5\times10^{-6}$ and $10^{-5}$; the reported run uses $10^{-5}$, 128-example optimizer batches and a 4,096-token prompt--response limit. Saved checkpoints are evaluated at temperature one with 32 responses per question and selected by six-benchmark AVG. Table~\ref{tab:appendix_table4_offline_sft_breakdown} gives the results. This ablation changes optimization settings and response reuse as well as generator updates; the matched Base-generator training control isolates generator updates while retaining fresh sampling.

\paragraph{Model-specific prompting.}
Qwen2.5 base/math models place the reasoning instruction in the system message, followed by the problem as the user message.
Qwen-Instruct retains its assistant identity system message and appends the reasoning instruction to the user's problem.
DeepSeek-Math-Instruct uses its own serialized system/user/assistant form.
Llama-Instruct in Table~\ref{tab:overall} uses a system reasoning instruction, without the stock knowledge-cutoff/date boilerplate.
All these reasoning instructions request step-by-step reasoning and a final boxed answer.
For Llama and DeepSeek base models, and for the Llama-Instruct generator control, we use the user-only SimpleRLZoo reasoning prompt \citep{zeng2025simplerl} without requiring a boxed answer (Appendix~\ref{app:source-budget}).
Qwen3 and Phi place the boxed-answer reasoning instruction in the user message, using their respective model templates.

\subsection{Compute accounting}\label{app:original-compute}
We use the dense-transformer approximation of $2PD$ FLOPs for a forward pass and $6PD$ for a training forward/backward pass, with $P=7\times10^9$ parameters and $D$ prompt-response tokens.
Writing $\bar D$ for the mean token count per generation--training round gives
\[
C_{\mathrm{SePT}}=8P\bar D_{\mathrm{SePT}},\qquad
C_{\mathrm{GRPO}}=10P\bar D_{\mathrm{GRPO}}.
\]
SePT counts generation and likelihood training; GRPO also counts a separately recomputed old-policy log-probability forward pass.
The KL regularizer is disabled, so no reference-model forward pass is included.
Costs average complete DSR runs, with two AdamW updates per generation--training round. They are approximate recipe-level FLOPs, not measured runtime or matched-compute comparisons.

\par
\begin{table}[ht]\centering\small
\caption{Estimated generation--training FLOPs per round, in $10^{15}$ operations. Each round has two optimizer updates; the ratio compares the specified recipes.}\label{tab:recipe-flops}
\begin{tabular}{lrrr}\toprule
Model & SePT & GRPO & GRPO/SePT\\\midrule
Qwen2.5-Math-7B &7.53&67.23&$8.93\times$\\
DeepSeek-Math-7B-Instruct &4.55&46.92&$10.32\times$\\\bottomrule
\end{tabular}
\end{table}
\par
The reset-student study additionally incurs the cost of acquiring its source; Appendix~\ref{app:source-details} reports those training-token counts separately.

\subsection{Decoding-temperature sweeps}
\begin{figure}[htbp]
\centering
\includegraphics[width=0.95\linewidth]{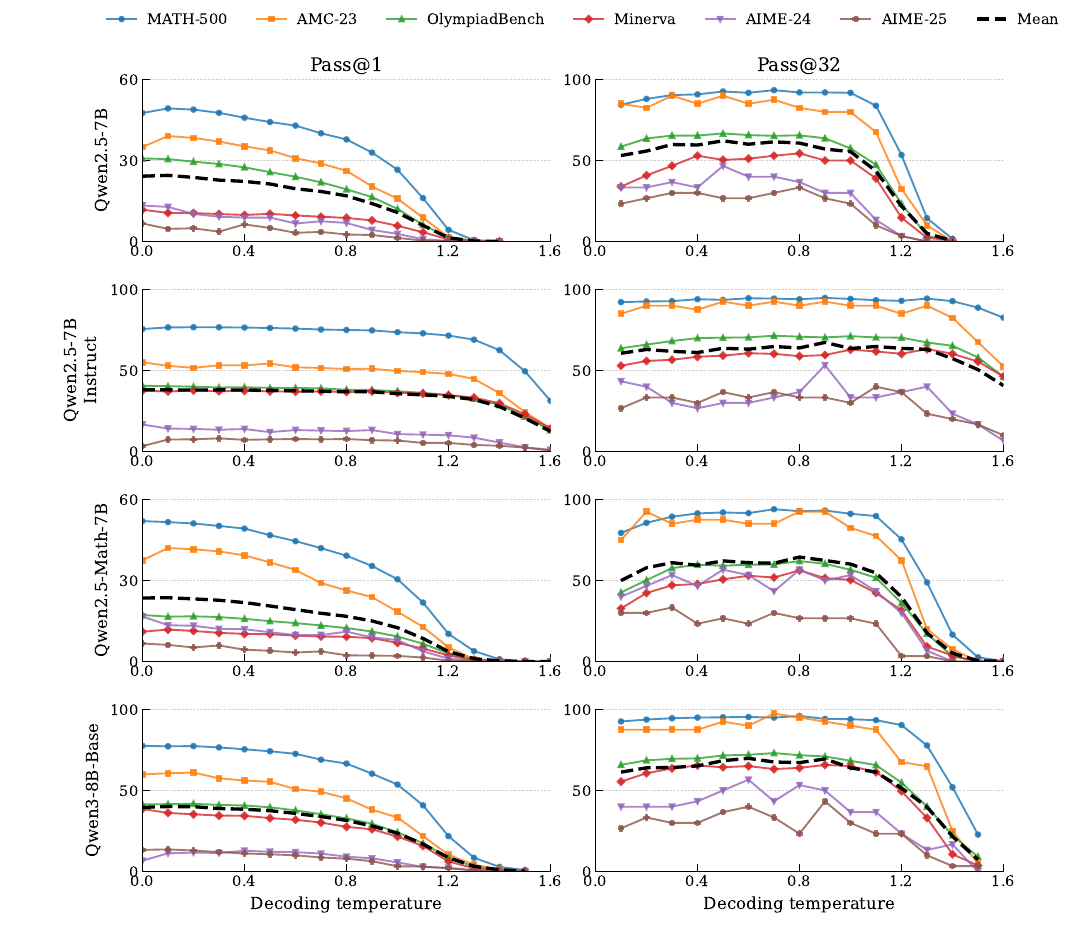}
\caption{Starting-model decoding-temperature sweeps: Qwen family. Columns show Pass@1 and Pass@32 (\%). Colored curves denote individual benchmarks; the dashed black curve is their mean. Pass@1 scales vary by model; Pass@32 shares a 0--100\% scale. Other model families appear in Figure~\ref{fig:additional-sweeps}.}\label{fig:original-other-sweeps}
\end{figure}

\begin{figure}[htbp]
\centering
\includegraphics[width=0.95\linewidth]{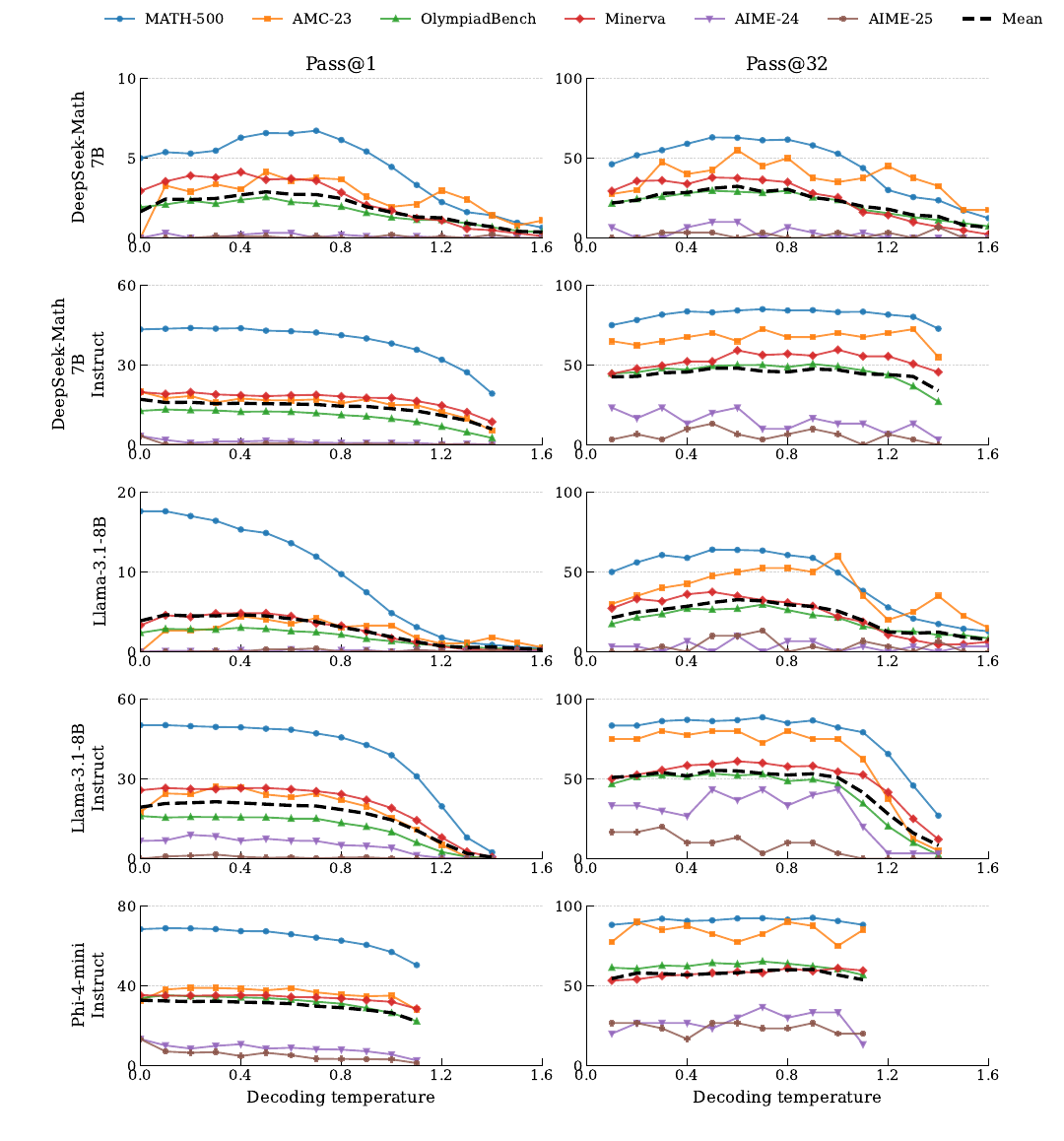}
\caption{Starting-model decoding-temperature sweeps: DeepSeek, Llama and Phi families. Columns show Pass@1 and Pass@32 (\%). Colored curves denote individual benchmarks; the dashed black curve is their mean. Pass@1 scales vary by model; Pass@32 shares a 0--100\% scale. The Qwen family appears in Figure~\ref{fig:original-other-sweeps}.}\label{fig:additional-sweeps}
\end{figure}

Figures~\ref{fig:original-other-sweeps} and~\ref{fig:additional-sweeps} show all nine starting-model sweeps. Prompts are fixed across temperatures and match the Baseline--SePT comparisons. Positive-temperature points use 32 responses per question; temperature zero reports only first-response accuracy. Curves connect measured points, with axes and colors explained in the captions.

\clearpage

\subsection{Full question and model responses}\label{app:original-case}
\paragraph{A trajectory-level illustration.}
\begin{figure}[t]
\centering
\includegraphics[width=0.93\linewidth]{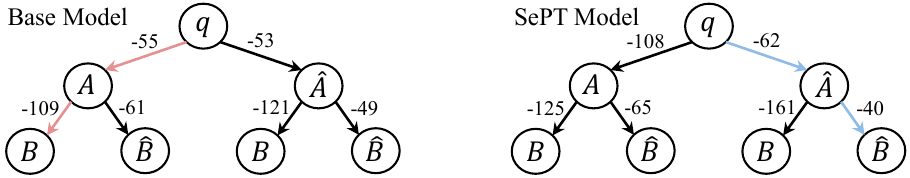}
\input{original/trace_probs}
\caption{Trajectory probabilities on a MATH-500 problem, evaluated at temperature one.
$q$ is the question; $A$ and $\hat A$ are the initial problem-setup segments of a failed base-model response and a successful SePT response, respectively.
$B$ and $\hat B$ are their subsequent reasoning and calculation segments, ending in the wrong answer 1 and the correct answer 4.
Each edge is a summed conditional log probability: $q\!\to\!A$ is $\log\pi(A\mid q)$ and $A\!\to\!B$ is $\log\pi(B\mid q,A)$, for example.
The left and right trees score the same segments under the base and SePT models; the equations give all twelve values explicitly.
Full responses appear in Figures~\ref{fig:qa_appendix_base}--\ref{fig:qa_appendix_osft}.}\label{fig:original-case}
\end{figure}
The question asks how many values can be obtained by inserting parentheses into $2\cdot3\cdot4\cdot5+1$, without rearranging terms.
Qwen2.5-Math-7B fails on all eight recorded attempts, whereas the displayed SePT response answers correctly.
Figure~\ref{fig:original-case} compares one failed response with that successful response, scoring both prefixes and both continuations under each model.

Before training, the alternative prefix $\hat A$ already has a slightly higher log probability than $A$ ($-53$ versus $-55$), but the margin is only two.
Conditioned on $\hat A$, the model also prefers the correct ending $\hat B$ over $B$ ($-49$ versus $-121$).
After SePT, the prefix margin grows to 46 and the latter suffix margin grows from 72 to 121.
The relative preferences already present in the initial model become stronger.

The displayed correct full trajectory has the same rounded log probability before and after training: $-53-49=-62-40=-102$.
The failed path instead changes from $-164$ to $-233$.
This case illustrates suppression of a competing path, with unchanged rounded probability for the displayed correct response. Aggregate accuracy depends on all correct responses, so this example alone does not explain the overall gain.

Figures~\ref{fig:qa_appendix_base} and~\ref{fig:qa_appendix_osft} give the full responses analyzed in Figure~\ref{fig:original-case}.
Blue backgrounds mark setup prefixes $A$ and $\hat A$; gray backgrounds mark the subsequent suffixes $B$ and $\hat B$.

\definecolor{colorA}{HTML}{EBF5FB}
\definecolor{colorB}{HTML}{F2F3F4}

\definecolor{colorAText}{HTML}{2874A6}
\definecolor{colorBText}{HTML}{4B5563}

\begin{figure}[h!]
\centering

\begin{tcolorbox}[
    colback=teal!2!white, 
    colframe=teal!20!white, 
    coltitle=black,
  title=\textbf{Example from MATH-500},
  fonttitle=\bfseries
]
\textbf{Problem}: The expression $2\cdot 3 \cdot 4\cdot 5+1$ is equal to 121, since multiplication is carried out before addition. However, we can obtain values other than 121 for this expression if we are allowed to change it by inserting parentheses. For example, we can obtain 144 by writing \[
(2\cdot (3\cdot 4)) \cdot (5+1) = 144.
\]In total, how many values can be obtained from the expression $2\cdot 3\cdot 4 \cdot 5 + 1$ by inserting parentheses? (Note that rearranging terms is not allowed, only inserting parentheses).

\end{tcolorbox}

\noindent
\begin{minipage}[ht]{\textwidth}
  \begin{tcolorbox}[
    bicolor,
    colback=colorA,
    colbacklower=colorB,
    colframe=teal!40!white, 
    coltitle=black,
    title=\textbf{Response (Base Model)},
    fonttitle=\bfseries,
  ]

To determine how many distinct values can be obtained from the expression \(2 \cdot 3 \cdot 4 \cdot 5 + 1\) by inserting parentheses, we need to consider all possible ways of inserting parentheses into the expression.

First, let's evaluate the expression without any parentheses:
\[2 \cdot 3 \cdot 4 \cdot 5 + 1 = 121\]

Now, we need to consider all possible ways of inserting parentheses.

\tcblower

There are five possible ways to insert one set of parentheses:
1. \(((2 \cdot 3) \cdot 4) \cdot 5 + 1\)
2. \((2 \cdot (3 \cdot 4)) \cdot 5 + 1\)
3. \((2 \cdot 3) \cdot (4 \cdot 5) + 1\)
4. \(2 \cdot ((3 \cdot 4) \cdot 5) + 1\)
5. \(2 \cdot (3 \cdot (4 \cdot 5)) + 1\)

However, the first and fifth expressions will be the same, as will the second and fourth. This leaves us with three unique ways of inserting parentheses, in addition to the original expression without parentheses.

So we need to evaluate each of these expressions:
1. \(((2 \cdot 3) \cdot 4) \cdot 5 + 1 = (6 \cdot 4) \cdot 5 + 1 = 24 \cdot 5 + 1 = 121\)
2. \((2 \cdot (3 \cdot 4)) \cdot 5 + 1 = (2 \cdot 12) \cdot 5 + 1 = 24 \cdot 5 + 1 = 121\)
3. \((2 \cdot 3) \cdot (4 \cdot 5) + 1 = 6 \cdot 20 + 1 = 120 + 1 = 121\)
4. \(2 \cdot ((3 \cdot 4) \cdot 5) + 1 = 2 \cdot (12 \cdot 5) + 1 = 2 \cdot 60 + 1 = 120 + 1 = 121\)
5. \(2 \cdot (3 \cdot (4 \cdot 5)) + 1 = 2 \cdot (3 \cdot 20) + 1 = 2 \cdot 60 + 1 = 120 + 1 = 121\)

From this, we see that all possible ways of inserting parentheses result in the value 121. Therefore, the only distinct value we can obtain is 121. 

Hence, the number of distinct values we can obtain from the expression \(2 \cdot 3 \cdot 4 \cdot 5 + 1\) by inserting parentheses is:
\[
\boxed{1}
\]

\end{tcolorbox}
\end{minipage}

\caption{Full question and the incorrect response generated by the base model, corresponding to the analysis in \Cref{fig:original-case}. The \textcolor{colorAText}{soft blue} and \textcolor{colorBText}{light gray} backgrounds distinguish the initial reasoning prefix ($A$) from the subsequent flawed suffix ($B$). This segmentation is because $B$ contains the core reasoning and calculation leading to the answer, while $A$ represents the initial problem setup.}
\label{fig:qa_appendix_base}
\end{figure}

\clearpage
\definecolor{colorA}{HTML}{EBF5FB}
\definecolor{colorB}{HTML}{F2F3F4}
\definecolor{colorAText}{HTML}{2874A6}
\definecolor{colorBText}{HTML}{4B5563}

\begin{figure}[h]
\centering

\begin{tcolorbox}[
    colback=teal!2!white, 
    colframe=teal!20!white, 
    coltitle=black,
  title=\textbf{Example from MATH-500},
  fonttitle=\bfseries
]
\textbf{Problem}: The expression $2\cdot 3 \cdot 4\cdot 5+1$ is equal to 121, since multiplication is carried out before addition. However, we can obtain values other than 121 for this expression if we are allowed to change it by inserting parentheses. For example, we can obtain 144 by writing \[
(2\cdot (3\cdot 4)) \cdot (5+1) = 144.
\]In total, how many values can be obtained from the expression $2\cdot 3\cdot 4 \cdot 5 + 1$ by inserting parentheses? (Note that rearranging terms is not allowed, only inserting parentheses).

\end{tcolorbox}

\noindent
\begin{minipage}[t]{\textwidth}
  \begin{tcolorbox}[
    bicolor,
    colback=colorA,
    colbacklower=colorB,
    colframe=teal!40!white, 
    coltitle=black,
    title=\textbf{Response (SePT Model)},
    fonttitle=\bfseries,
  ]

To determine how many different values can be obtained from the expression \(  2 \cdot 3 \cdot 4 \cdot 5 + 1\) by inserting parentheses, we need to consider all possible ways to group the terms. Let's break this down step by step.

First, let's list all the possible ways to insert parentheses in the expression \( 2 \cdot 3 \cdot 4 \cdot 5 + 1 \):

\tcblower

1. \((2 \cdot 3 \cdot 4 \cdot 5) + 1\)
2. \(2 \cdot (3 \cdot 4 \cdot 5 + 1)\)
3. \(2 \cdot 3 \cdot (4 \cdot 5 + 1)\)
4. \(2 \cdot (3 \cdot (4 \cdot 5 + 1))\)
5. \((2 \cdot 3 \cdot 4) \cdot (5 + 1)\)
6. \((2 \cdot 3) \cdot (4 \cdot (5 + 1))\)
7. \(2 \cdot ((3 \cdot 4) \cdot (5 + 1))\)
8. \(((2 \cdot 3) \cdot 4) \cdot (5 + 1)\)
9. \((2 \cdot (3 \cdot 4)) \cdot (5 + 1)\)

Now, let's evaluate each of these expressions:

1. \((2 \cdot 3 \cdot 4 \cdot 5) + 1 = 120 + 1 = 121\)
2. \(2 \cdot (3 \cdot 4 \cdot 5 + 1) = 2 \cdot (60 + 1) = 2 \cdot 61 = 122\)
3. \(2 \cdot 3 \cdot (4 \cdot 5 + 1) = 6 \cdot (20 + 1) = 6 \cdot 21 = 126\)
4. \(2 \cdot (3 \cdot (4 \cdot 5 + 1)) = 2 \cdot (3 \cdot 21) = 2 \cdot 63 = 126\)
5. \((2 \cdot 3 \cdot 4) \cdot (5 + 1) = 24 \cdot 6 = 144\)
6. \((2 \cdot 3) \cdot (4 \cdot (5 + 1)) = 6 \cdot (4 \cdot 6) = 6 \cdot 24 = 144\)
7. \(2 \cdot ((3 \cdot 4) \cdot (5 + 1)) = 2 \cdot (12 \cdot 6) = 2 \cdot 72 = 144\)
8. \(((2 \cdot 3) \cdot 4) \cdot (5 + 1) = (6 \cdot 4) \cdot 6 = 24 \cdot 6 = 144\)
9. \((2 \cdot (3 \cdot 4)) \cdot (5 + 1) = (2 \cdot 12) \cdot 6 = 24 \cdot 6 = 144\)

From the evaluations, we see that the possible values are 121, 122, 126, and 144. Therefore, there are 4 distinct values that can be obtained.

The final answer is \(\boxed{4}\).

\end{tcolorbox}
\end{minipage}

\caption{Full question and the SePT response ending in the correct answer, corresponding to \Cref{fig:original-case}. The \textcolor{colorAText}{soft blue} and \textcolor{colorBText}{light gray} backgrounds mark the setup prefix ($\hat{A}$) and subsequent reasoning suffix ($\hat{B}$). Answer correctness does not independently certify each intermediate step.}
\label{fig:qa_appendix_osft}
\end{figure}

\clearpage
\section{Code generation: adaptation and training stability}\label{app:coding}
SePT yields positive CodeScore point estimates with a hot sampling schedule in both the initial run and a fixed-duration repeat. The controls below examine response budgets, learning rates and training duration. Section~\ref{sec:boundaries} places the trajectories in the context of prior work.

We test SePT on code generation with Qwen3-4B-Instruct-2507, using 175 questions
from the LiveCodeBench v6 increment and 378 MBPP+ tasks.
All comparisons in Table~\ref{tab:coding-fullsupport} use full-support sampling
and a 16,384-token response allowance.
Pass@1 and Pass@5 are estimated from five responses per question using
Appendix~\ref{app:passk}; CodeScore is their equal-weight mean across the two
benchmarks, distinct from the mathematical AVG metric.
The Coding Baseline selects one temperature from
$\{0.2,0.4,0.7,1.0,1.2,1.6\}$ by CodeScore, choosing 1.2.
SePT in Table~\ref{tab:coding-fullsupport} is evaluated at temperature one.

\begin{table}[htbp]
\centering\small
\setlength{\tabcolsep}{5pt}
\caption{Coding performance under a common 16K evaluation allowance. LR denotes the configured peak. All rows use the same tasks and evaluation sample count. Checkpoint selection and warmup details are given below.}
\label{tab:coding-fullsupport}
\begin{tabular}{lrrr}
\toprule
Configuration & Pass@1 & Pass@5 & CodeScore \\
\midrule
Coding Baseline & 49.67 & 58.82 & 54.25 \\
SePT, 8K training & 49.70 & 58.51 & 54.11 \\
SePT, 16K training, LR $10^{-7}$ & 49.84 & 59.08 & 54.46 \\
SePT, 16K training, LR $10^{-6}$ & 50.33 & 59.50 & 54.92 \\
\bottomrule
\end{tabular}
\end{table}

\paragraph{Matched training and selection.}
The two 16K runs start from the same initial model with fresh optimizers, each
using 8,192 questions from the recorded rStar-Coder subset over 64 refreshes and 128 AdamW updates.
Each refresh generates 128 responses for two 64-response, token-mean updates,
without correctness feedback or filtering.
Both arms use ten-optimizer-update warmup, AdamW moments $(0.9,0.999)$, weight decay 0.01
and gradient clipping at one. Sampling temperature rises from 0.9 toward one
over 128 refresh intervals, without compressing the schedule at the stopping
budget, reaching 0.9492; training temperature stays at one.
The prompt/context limits are 2,048/32,768 tokens. No top-$p$/top-$k$ filtering
or artificial end token at the length cutoff is used.

We monitor 14 saved states on a fixed subset of 64 questions from each benchmark,
with one response per question; it is not held out.
The two best monitor Pass@1 states and the endpoint receive full five-response
evaluation. CodeScore selects among these, with earlier states preferred in ties.
The 8K-trained row evaluates a separately trained SePT model
under the same 16K decoder; it is not the matched learning-rate control.

\paragraph{Learning rate and training stability.}
The larger-peak-LR run's selected state is within warmup, after updates at zero
and $10^{-7}$; it has not reached the configured peak of $10^{-6}$.
Continued training at the higher configured rate deteriorates. The larger- minus smaller-LR contrast is $+0.45$ points $[-0.73,1.65]$.
Relative to the Coding Baseline, the smaller- and larger-LR runs gain
$0.22$ points $[-1.07,1.50]$ and $0.67$ points $[-0.44,1.78]$, respectively.
These small point differences are not resolved by the reported intervals.
The intervals resample paired questions within each benchmark and
condition on the evaluated configurations, as in Appendix~\ref{app:statistics}.

\Needspace{9\baselineskip}
\paragraph{Cold-source model changes on fixed responses.}
To distinguish changing inputs from changes in the model, we score 128 fixed
initial-model responses under the initial, warmup-selected and cold-source
endpoint models, without correctness selection.
Token-weighted cross-entropy is $0.3726$, $0.3726$ and $0.6906$ nats per token.
Mean full-vocabulary entropy at up to 16 evenly spaced prediction positions per
question is $0.2053$, $0.2055$ and $0.0678$ nats, respectively.
Thus the cold endpoint assigns lower likelihood to the same initial-model
responses while becoming more concentrated at these fixed prefixes.
This diagnostic separates fixed-content changes from declining training loss
on changing responses; likelihood and entropy are not correctness measures.

\paragraph{Response-budget diagnostic.}
We also grade the first 8,192 native response tokens of the initial and
8K-trained models' temperature-one outputs, alongside their complete 16K outputs.
The larger allowance increases CodeScore by $0.60$ points $[0.09,1.26]$ for the
initial model and $0.46$ points $[0.06,1.03]$ for the trained model.
Both correctness gains and losses enter the paired differences. The test compares two lengths of the same responses, without selecting the better score per response or varying training length.

\subsection{Hot-source adaptation and early stopping}\label{app:coding-hot}

We next reverse the sampling temperature direction while retaining full support:
$\tau_s(r)=1.2-0.2\min\{(r-1)/128,1\}$.
The model, questions, response allowance, training temperature, AdamW settings,
peak learning rate $10^{-6}$ and two updates per refresh remain unchanged.
Near equal sampling and training temperatures, hot and cold sources induce opposite first-order mean logit updates under the sequence-sum reference (Appendix~\ref{app:reference-proofs}).
A label-free 128-question pilot checks generation length and repetition before
training; every generated response is retained.

The 64-refresh run uses the same finite monitoring and selection procedure
described above. SePT attains a CodeScore increase of $1.33$ points
$[-0.35,3.03]$ over the Coding Baseline, while its prescribed endpoint falls
to $26.56\%$, below the matched cold-source endpoint by $22.93$ points
$[-26.62,-19.33]$. Early adaptation and continued training therefore have
different effects in this setting.

\begin{table}[htbp]
\centering\small
\setlength{\tabcolsep}{5pt}
\caption{Hot-source coding results in percent. The initial run and fixed-duration
repeat are shown against their respective Coding Baselines. Trained models
are evaluated at temperature one; Coding Baselines use temperature 1.2.
The final block gives the same-decoder reference, the other
shortlisted model and the longer-run endpoint. The repeat uses fresh training
and evaluation responses.}
\label{tab:coding-hot}
\begin{tabular}{lrrr}
\toprule
Configuration & Pass@1 & Pass@5 & CodeScore \\
\midrule
\multicolumn{4}{l}{\emph{Initial run}}\\
Coding Baseline & 49.67 & 58.82 & 54.25 \\
Hot-source SePT & 51.03 & 60.12 & 55.58 \\
\midrule
\multicolumn{4}{l}{\emph{Fixed-duration repeat}}\\
Fresh Coding Baseline & 49.68 & 58.93 & 54.30 \\
Hot-source SePT & 49.98 & 61.00 & 55.49 \\
\midrule
\multicolumn{4}{l}{\emph{Additional references and trajectory diagnostics}}\\
Fresh initial model, temperature one & 48.73 & 58.65 & 53.69 \\
Hot source, other shortlisted model & 50.41 & 58.49 & 54.45 \\
Hot source, prescribed endpoint & 19.20 & 33.93 & 26.56 \\
\bottomrule
\end{tabular}
\end{table}

\paragraph{Prospective confirmation.}
We fix the selected duration of 16 refreshes (32 updates) before repeating
training from the initial model and a fresh optimizer, using independent
response-generation randomness and the same prompt order.
The sampling temperature and learning-rate schedules are unchanged, so the source ends at
temperature 1.1766 rather than compressing its anneal to one.
Only this endpoint is evaluated, with fresh five-response panels for the trained
model and the initial model at temperatures 1 and 1.2; no additional selection
is performed.
CodeScore gains $1.19$ points $[-0.32,2.72]$ over the fresh Coding Baseline.
Pass@1 and Pass@5 differences are $+0.30$ $[-1.03,1.64]$ and
$+2.07$ $[-0.07,4.27]$ points.
Against the initial model at the same temperature-one decoder, CodeScore gains
are $1.56$ $[0.07,3.07]$ in the selected run and $1.80$ $[0.33,3.30]$ in the
confirmation. These paired-question intervals describe the evaluated models,
not variability across training runs.

The confirmation's LiveCodeBench and MBPP+ CodeScores are $37.14\%$ and
$73.84\%$, versus fresh Baseline values $35.83\%$ and $72.78\%$.
Their paired differences are $+1.31$ $[-1.09,3.83]$ and
$+1.06$ $[-0.74,2.88]$ points, respectively.
Both benchmarks contribute positive point estimates, and the repeat reproduces the direction and approximate magnitude of the initial gain. The longer trajectory shows why early adaptation and sustained training should be evaluated separately.

\paragraph{Computational budget.}
The initial hot-source run and repeat generate 10,240 training responses (57.40 million tokens) across 160 updates. Eight-GPU training windows total 58.03 GPU-hours, including loading, synchronization and checkpoint export. Adding the pilot, fixed-input probe, monitoring and evaluations gives 28,878 responses and 97.98 million tokens, excluding reused panels.

\end{document}